\documentclass[a4paper,twoside]{article}

\usepackage{amsfonts}
\usepackage{amssymb,bm}
\usepackage{amstext}
\newif\ifwrapfigavailable
\IfFileExists{wrapfig.sty}{\wrapfigavailabletrue}{\wrapfigavailablefalse}
\ifwrapfigavailable
  \usepackage{wrapfig}
\else
  \newenvironment{wrapfigure}[2]{%
    \begin{figure}[htbp]\centering
  }{%
    \end{figure}
  }
\fi
\IfFileExists{booktabs.sty}{%
  \usepackage{booktabs}%
}{%

}
\usepackage{amsmath}
\usepackage{xspace}
\usepackage{graphicx}
\graphicspath{{figures/}{images/}}
\usepackage{url}
\usepackage[authoryear,square]{natbib}
\IfFileExists{enumitem.sty}{\usepackage{enumitem}}{}
\IfFileExists{subfigure.sty}{%
  \usepackage{subfigure}%
}{%
  \IfFileExists{subcaption.sty}{\usepackage{subcaption}}{}%
}
\usepackage{graphics}
\IfFileExists{colordvi.sty}{\usepackage{colordvi}}{}
\IfFileExists{float.sty}{\usepackage{float}}{}
\IfFileExists{cite.sty}{\usepackage{cite}}{}
\newif\ifxcoloravailable
\IfFileExists{xcolor.sty}{\xcoloravailabletrue}{\xcoloravailablefalse}
\ifxcoloravailable
  \usepackage{xcolor}
\else
  \IfFileExists{color.sty}{\usepackage{color}}{}
\fi
\definecolor{DeepBlue}{rgb}{0.0,0.0,1.0}
\usepackage{framed}
\usepackage{setspace}
\ifxcoloravailable
  \colorlet{shadecolor}{orange!15}
\else
  \definecolor{shadecolor}{rgb}{1.0,0.94,0.86}
\fi
\newif\ifDantzigHyperrefAvailable
\IfFileExists{etoolbox.sty}{%
  \IfFileExists{hyperref.sty}{\DantzigHyperrefAvailabletrue}{\DantzigHyperrefAvailablefalse}%
}{%
  \DantzigHyperrefAvailablefalse
}
\ifDantzigHyperrefAvailable
  \usepackage{hyperref}%
  \hypersetup{%
    colorlinks=true,
    linkcolor=DeepBlue,
    anchorcolor=DeepBlue,
    citecolor=DeepBlue,
    urlcolor=DeepBlue
  }%
  \IfFileExists{xr-hyper.sty}{\usepackage{xr-hyper}}{}%
\else
  \providecommand{\href}[2]{{\color{DeepBlue}#2}}%
  \makeatletter
  \let\Dantzig@NAT@citex\NAT@citex
  \def\NAT@citex[#1][#2]#3{\color{DeepBlue}\Dantzig@NAT@citex[#1][#2]{#3}}%
  \let\Dantzig@NAT@citexnum\NAT@citexnum
  \def\NAT@citexnum[#1][#2]#3{\color{DeepBlue}\Dantzig@NAT@citexnum[#1][#2]{#3}}%
  \makeatother
\fi
\providecommand{\BLUE}[1]{{\color{blue}#1}}
\providecommand{\RED}[1]{{\color{red}#1}}

\IfFileExists{comment.sty}{\usepackage{comment}}{}
\IfFileExists{soul.sty}{\usepackage{soul}}{}

\newif\ifcaptionavailable
\IfFileExists{caption.sty}{\captionavailabletrue}{\captionavailablefalse}
\ifcaptionavailable
  \usepackage{caption}
\else
  \newcommand{\captionof}[2]{%
    \refstepcounter{#1}%
    \def\captiontype{#1}%
    \def\algorithmtype{algorithm}%
    \par\smallskip\noindent\small
    \ifx\captiontype\algorithmtype
      \textbf{Algorithm~\csname the#1\endcsname.} #2%
    \else
      \textbf{\csname #1name\endcsname~\csname the#1\endcsname.} #2%
    \fi
    \par
  }
\fi

\newif\ifalgorithmicavailable
\IfFileExists{algorithmic.sty}{\algorithmicavailabletrue}{\algorithmicavailablefalse}
\ifalgorithmicavailable
  \usepackage{algorithmic}
\else
  \newcounter{algorithmicline}
  \newcommand{\AlgorithmicItem}{%
    \refstepcounter{algorithmicline}%
    \item[{\footnotesize\thealgorithmicline:}]%
  }
  \newenvironment{algorithmic}[1][]{%
    \setcounter{algorithmicline}{0}%
    \begin{list}{}{\leftmargin=2.7em\labelwidth=2.1em
      \labelsep=.5em\itemsep=2pt\parsep=0pt}
  }{%
    \end{list}
  }
  \newcommand{\STATE}{\AlgorithmicItem}
  \newcommand{\FOR}[1]{\AlgorithmicItem \textbf{for} #1 \textbf{do}}
  \newcommand{\ENDFOR}{\AlgorithmicItem \textbf{end for}}

\fi

\providecommand{\bbR}{\mathbb{R}}
\providecommand{\Ex}{\mathbb{E}}
\providecommand{\Exk}{\mathbb{E}_k}

\providecommand{\AdamMini}{Adam-mini}
\providecommand{\AdamW}{AdamW}
\providecommand{\betabeta}{(\beta_1,\beta_2)}
\providecommand{\functionclass}{\mathcal{F}_{L, D_0, D_1}^{n,f^*}(\mathbb{R}^d)}
\providecommand{\yushunrevise}[1]{\color{black}{#1}}
\providecommand{\subfigure}[2][]{#2}
\renewcommand{\l}{\left}
\renewcommand{\r}{\right}
\providecommand{\rw}{\rightarrow}
\newcommand{\be}{\begin{equation}}
\newcommand{\ee}{\end{equation}}
\newcommand{\bea}{\begin{equation}\begin{aligned}}
\newcommand{\eea}{\end{aligned}\end{equation}}
\newcommand{\wt}{\widetilde}
\newcommand{\R}{\mathbb{R}}  
\newcommand{\mb}{\mathbb}

\newcommand{\E}{\mathbb{E}}

\newcommand\ml{\mathcal}

\renewcommand{\l}{\left} 
\renewcommand{\r}{\right} 
\newenvironment{Snugshade}[1][gray!90]{
    \colorlet{shadecolor}{#1}
    \begin{snugshade}
}{
    \end{snugshade}
}

\makeatletter
\long\def\@makecaption#1#2{%
  \vskip\abovecaptionskip
  \small
  \sbox\@tempboxa{#1 #2}%
  \ifdim \wd\@tempboxa >\hsize
    #1 #2\par
  \else
    \global\@minipagefalse
    \hb@xt@\hsize{\hfil\box\@tempboxa\hfil}%
  \fi
  \vskip\belowcaptionskip}

\@addtoreset{table}{section}
\renewcommand\thetable{\thesection.\arabic{table}}
\def\fnum@table{\tablename\nobreakspace\thetable:}

\@addtoreset{figure}{section}
\renewcommand\thefigure{\thesection.\arabic{figure}}
\renewcommand\figurename{Fig.}
\def\fnum@figure{\figurename\nobreakspace\thefigure.}

\def\@seccntformat#1{\csname the#1\endcsname. }
\renewcommand\section{\@startsection {section}{1}{\z@}%
                                     {3.5ex \@plus 1ex \@minus .2ex}%
                                     {2.3ex \@plus.2ex}%
                                     {\normalfont\large\bfseries\centering}}
\renewcommand\subsection{\@startsection {subsection}{2}{\z@}%
                                        {3.25ex\@plus 1ex \@minus .2ex}%
                                        {1.5ex \@plus .2ex}%
                                        {\normalfont\normalsize\bfseries}}

\renewcommand\@openbib@code{\parsep \z@ \itemsep \z@ \parskip \z@ \small}
\@addtoreset{equation}{section}

\def\@begintheorem#1#2{\trivlist
   \item[\hskip \labelsep{\bfseries #1\ #2.}]\itshape}
\newtheorem{theorem}{Theorem}[section]
\newtheorem{lemma}{Lemma}[section]

\newcounter{algorithm}[section]
\renewcommand\thealgorithm{\thesection.\arabic{algorithm}}
\providecommand{\algorithmname}{Algorithm}
\providecommand{\fnum@algorithm}{\algorithmname\nobreakspace\thealgorithm}
\newenvironment{algorithm}[1][]{%
  \par\smallskip\noindent
  \refstepcounter{algorithm}%
  \begingroup
  \begin{minipage}{\linewidth}
  \hrule height .8pt
  \def\caption##1{\vskip 3pt\noindent\small
    \textbf{\algorithmname\nobreakspace\thealgorithm.} ##1\par
    \vskip 3pt\hrule height .4pt\vskip 4pt}%
}{%
  \par\vskip 3pt\hrule height .8pt
  \end{minipage}
  \endgroup\smallskip
}
\newtheorem{assumption}{Assumption}[section]

\newcommand\ps@firstpage{%
  \def\@oddhead{\small 2026 INFORMS George B. Dantzig Dissertation Competition\hfil}%
  \let\@evenhead\@oddhead
  \def\@oddfoot{\hfil\thepage\hfil}%
  \let\@evenfoot\@oddfoot}

\@addtoreset{footnote}{page}
\renewcommand\thefootnote{\arabic{footnote})}
\renewcommand\maketitle{\par
  \begingroup
    \c@footnote\m@ne
    \renewcommand\thefootnote{%
      \ifcase\c@footnote
        *%
      \or
        \ensuremath{\dagger}%
      \else
        \@arabic\c@footnote)%
      \fi}%
    \def\@makefnmark{\hbox{\@textsuperscript{\normalfont\@thefnmark}}}%
    \long\def\@makefntext##1{\parindent 1em\noindent
            \hb@xt@1.8em{%
                \hss\@textsuperscript{\normalfont\@thefnmark} }##1}%
    \newpage
    \global\@topnum\z@
    \@maketitle
    \thispagestyle{firstpage}%
    \@thanks
  \endgroup
  \global\let\thanks\relax
  \global\let\maketitle\relax
  \global\let\@maketitle\relax
  \global\let\@thanks\@empty
  \global\let\@author\@empty
  \global\let\@date\@empty
  \global\let\@title\@empty
  \global\let\title\relax
  \global\let\author\relax
  \global\let\date\relax
  \global\let\and\relax
}

\def\@maketitle{%
  \newpage
  \null
  \vskip 2em%
  {\centering
    {\large\textbf{\@title}}\par
    \vskip 1.5em%
    {\small
     \def\and{\ifvmode \else {and }\fi}%
     \@author
    }%
  \par}}

\renewcommand\makeatother

\begin{document}

\setcounter{page}{1}

\markboth{Y.~ZHANG}{On the Principles Behind Neural Network Optimizers}

\title{On the Principles Behind Neural Network Optimizers}

\author{Yushun Zhang
\thanks{This is a 24-page summary of Y. Zhang's  Ph.D. thesis under the same title. This version is prepared for the Dantzig Dissertation Competition, which imposes a strict page limit. Accordingly, we only present the main result. Proofs are omitted, and the review of related work is substantially reduced and is not intended to be comprehensive in this version.}
\thanks{This Ph.D. dissertation is supervised by the following committee members:
Zhi-Quan Luo (Advisor), Ruoyu Sun (Chair), Yinyu Ye, Jim Dai,
Anthony Man-cho So, Madeleine Udell. The degree is conferred on March 31, 2026.}\\
School of Data Science, The Chinese University of Hong Kong, Shenzhen, China\\
\texttt{yushunzhang@link.cuhk.edu.cn}}

\maketitle

\begin{abstract}
Reliable optimization is central to neural network (NN) training, yet Adam, the default optimizer for modern LLMs, rests on a fragile foundation. This thesis develops a principled grounding for Adam and motivates new designs. First, we revisit Adam's divergence--convergence debate and show the existence of a problem-dependent phase transition: with properly chosen, batch-size-dependent hyperparameters, Adam converges, whereas under small-$\beta_2$ regimes it can diverge. Second, we investigate why Adam substantially outperforms SGD on Transformers through Hessian structure. We find that the Hessian evolves toward a near-block-diagonal form along training, accompanied by strong block heterogeneity. We prove that this structure makes Adam's diagonal preconditioner effective. We further show that this special Hessian structure originates from consecutive multiplications of large matrix variables, and we provide a rigorous analysis based on random matrix theory. Finally, these insights motivate Adam-mini, a new optimizer that reduces Adam's memory footprint by 50\% while preserving its performance. Our results also have broader implications beyond Adam: they reveal new local structures in matrix-based nonconvex problems, and also help understand and improve recent NN optimizers, such as Muon. 
\end{abstract}

\section{Introduction}
\label{sec_intro}

Neural networks (NNs),  evolving from classical multi-layer perceptrons (MLPs) to contemporary large language models (LLMs),  are the core engine of modern artificial intelligence (AI) (e.g., \citep{achiam2023gpt}).  
The effectiveness of these models hinges on their ability to learn complex patterns from data, a process fundamentally framed as an {\it optimization} problem. In this thesis, we focus on the following empirical risk minimization problem (ERM), which is the standard optimization formulation for training modern NNs:
{\small
\begin{equation} \label{finite_sum}
  \operatorname*{minimize}_{\theta \in \mathbb{R}^{d}} \quad \ell(\theta):=\frac{1}{n}\sum_{i=0}^{n-1} \ell_{i}(\theta).
\end{equation}
}
Here, $\theta \in \mathbb{R}^{d}$ denotes the trainable NN parameters, $n \in \mathbb{N}$ is the number of mini-batches that partition the dataset, and $\ell_i(\theta)$ denotes the training loss on the $i$-th mini-batch data.  

To solve \eqref{finite_sum} at scale, a stable and computationally efficient optimizer is needed.
Yet, despite their widespread deployment in industry, current NN optimizers are built on a surprisingly fragile foundation. 
The most prominent example is Adam \citep{kingma2014adam}. On the one hand, Adam has become the most widely used optimizer for modern NNs, including LLMs (e.g., Llama and DeepSeek series \citep{touvron2023llama,liu2024deepseek}).  On the other hand, Adam has long been criticized for having ``theoretical flaws''.
In particular, an influential work \citep{reddi2019convergence} (the winner of {\it ICLR 2018 Best Paper Award} \citep{ICLR2018BestPaper})  suggested that Adam can {\it diverge even on simple examples}. This finding raises serious concerns for Adam's deployment in AI model training,  where the divergence can raise alerts of unpredictable training failures.

Further, when deployed regardless of the risk,  Adam's empirical behavior usually falls outside the scope of existing theory and thus remains difficult to predict.  For example, it underperforms GD even on simple random Gaussian quadratic objectives, yet it consistently outperforms SGD on more complex Transformer training (see later in Section \ref{sec_hessian}), and these contrasting regimes are not well explained by existing theory. All these create substantial uncertainty and concern for deploying optimizers in real AI systems. The concern is dramatically magnified by the {\it multi-million-dollar} scale of modern training, where even minor risks caused by unknown factors become financially unaffordable.

This thesis aims to develop a theoretical grounding and principled foundation behind current NN optimizers.  We focus primarily on Adam, while we also offer new theoretical insights into the special Hessian structure of $\ell (\theta)$ in  \eqref{finite_sum}  that emerges in certain classes of NNs. Our findings have broader implications beyond Adam. For instance, they reveal new local structures in matrix-based nonconvex problems. They also help improve recent NN optimizers, such as Muon \citep{jordan2024muon}.

\begin{wrapfigure}{r}{0.37\textwidth}
    \centering
    \includegraphics[width=1\linewidth]{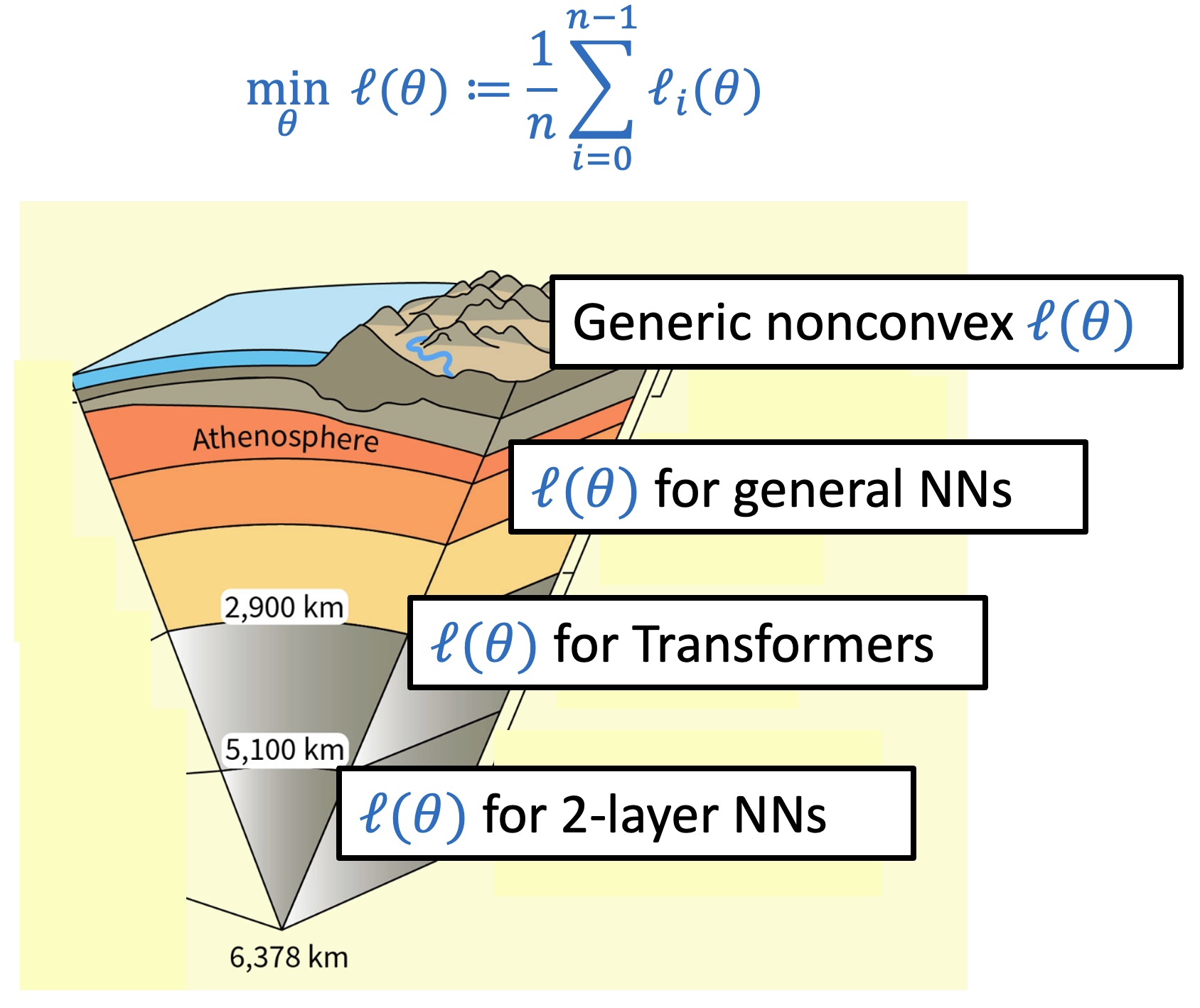}
    \caption{\small Overview of the thesis. We begin with general nonconvex objectives and then narrow down to the special structures found in NNs.}
    \label{fig:intro-roadmap-placeholder}
\end{wrapfigure}
This thesis proceeds in four stages: beginning with a revisit of the basic convergence guarantees of Adam on generic nonconvex $\ell (\theta)$ in \eqref{finite_sum}; moving to understand empirical behaviors of Adam by connecting it to the special Hessian structure of \eqref{finite_sum}  that emerges in certain classes of NNs; digging deeper to find out when and why the special structure occurs on 1-hidden-layer NNs; and ultimately, translating these insights into new optimizer designs. 
 Concretely, this thesis focuses on four questions:

(i) \textbf{Safety and guarantees:}  Does vanilla Adam diverge or converge? If it can converge, how should we choose its hyperparameters to avoid divergence risk?

(ii) \textbf{Mysterious behavior:} SGD (despite its success on CNNs) largely underperforms Adam on Transformers. What special problem structure drives this gap?

(iii) \textbf{Problem structure:} When and why does the special problem structure of NNs occur? What are the root causes?

(iv) \textbf{Design principles:} Can we exploit NN-specific problem structure to build better optimizers?

By investigating these questions both empirically and theoretically, this thesis enables a safer deployment, more reliable tuning, and a deeper understanding of current optimizers and the problem structure of NNs. These results establish a theoretical grounding and principled foundation
behind current NN optimizers. We also collect these insights and propose a new optimizer with principled design. One takeaway lesson from this thesis is that optimizer behavior is usually tied to some special intrinsic structure of NNs---a factor not fully captured by classical theory.

On the technical side, we develop new concentration tools for the convergence result of Adam, and new decoupling methods in random matrix theory for the Hessian-structure analysis, which may be of independent interest.

\subsection{Preliminaries}
\vspace{-0.5cm}
\begin{wrapfigure}{r}{0.45\textwidth}
\vspace{-0.25cm}
  \centering
  \begin{minipage}{\linewidth}
    \footnotesize
    \begin{algorithm}[H]
      \caption{Adam}
      \label{algorithm_wr}
      \begin{algorithmic}[1]
        \STATE Initialize $m_{0}$, $v_{0}$, $\theta_{1}$.
        \FOR{$k=1\to\infty$}
          \STATE Sample $\tau_k$ uniformly from $\{0,1,\dots,n-1\}$
          \STATE $m_k = \beta_1 m_{k-1} + (1-\beta_1)\nabla \ell_{\tau_k}(\theta_k)$
          \STATE $v_k = \beta_2 v_{k-1} + (1-\beta_2)\nabla \ell_{\tau_k}(\theta_k)\circ\nabla \ell_{\tau_k}(\theta_k)$
          \STATE $\theta_{k+1} = \theta_k - \frac{\eta_k}{\sqrt{v_k}+\epsilon}\circ m_k$
        \ENDFOR
      \end{algorithmic}
    \end{algorithm}
  \end{minipage}
\end{wrapfigure}
\paragraph{Notations.} 
Throughout the thesis, we consider problem \eqref{finite_sum}.
We use $\theta$  to denote the optimization variable, which refers to NN parameters in deep learning scenarios. 
We denote $\nabla \ell_{i}$ as the gradient of $\ell_{i}$. The update rule of Adam is presented in  Algorithm  \ref{algorithm_wr}. Here, $m$ and $v$ denote the 1st-order and 2nd-order momentum, controlled by hyperparameters $\beta_1, \beta_2 \in [0,1]$, respectively. The product $\circ$, division, and square-root operators are component-wise. We denote $\theta_{k}, m_{k}, v_{k} \in \mathbb{R}^{d}$ as the value of $\theta, m, v$ at the $k$-th iteration, respectively.   In practice, $\epsilon$ is adopted for numerical stability and it is often chosen to be $10^{-8}$. In our theory, we allow $\epsilon$ to be an arbitrary non-negative constant including $0$.

\paragraph{Popularity of Adam.}
In deep learning, Adam \citep{kingma2014adam} is one of the most popular algorithms for solving \eqref{finite_sum}. 
It has been applied to various machine learning domains such as natural language processing (NLP) and computer vision (CV) (e.g., \citep{vaswani2017attention,dosovitskiy2021an}).
Its impact is also evidenced by over 250,000 citations, a number that continues to grow rapidly \citep{adamcitation}. Moreover, with the rise of ChatGPT \citep{openai_chatgpt_2022}, Adam has become the crucial algorithm for training large language models (LLMs). Adam is reported to be used to train mainstream LLMs, including Llama series \citep{touvron2023llama},  Qwen series \citep{bai2023qwen}, and DeepSeek series \citep{liu2024deepseek}, etc. Adam is clearly serving as a major horsepower behind the advancement of AI.  Its influence was recently recognized when it received the {\it ICLR 2025 Test-of-Time Award} \citep{ICLR2025TestOfTime}.

 However, despite its ubiquity in industry, Adam is built on a surprisingly fragile foundation, and its practical behavior remains poorly understood. We will review the major theoretical criticisms and present new results and insights in the following chapters.

\paragraph{Organization of the thesis.}  Section \ref{sec_adam} revisits the divergence--convergence debate of Adam and presents a convergence guarantee of Adam under problem-dependent hyperparameters. Section \ref{sec_hessian} studies when Adam is effective, connecting the Adam--SGD gap on Transformers to Hessian block structure and block heterogeneity, and then explaining where this structure comes from in simplified neural networks. Section \ref{sec_adam_mini} turns these structural insights into Adam-mini, a new optimizer that keeps Adam-like performance while reducing Adam's memory footprint, and the final section concludes.

\section{Divergence or Convergence? A Story of Adam }
\label{sec_adam}

\begin{Snugshade}[gray!10]
{\small This section is mainly based on:
\begin{center}
\begin{itemize}
    \item \BLUE{\bf Adam Can Converge Without Any Modification On Update Rules.}  \\
    {\bf Yushun Zhang}, Congliang Chen, Naichen Shi, Ruoyu Sun,  Zhi-Quan Luo. 
    
    Advances in Neural Information Processing Systems (NeurIPS), 2022 \RED{\bf (Spotlight)}.
    \item  \BLUE{\bf Adam Converges Without Any Modification On Update Rules.}  \\
    {\bf Yushun Zhang}, Bingran Li, Congliang Chen, Zhi-Quan Luo, Ruoyu Sun. 
    
   (This is the journal extended version.)

   Under Review.
\end{itemize}
\end{center}
}
\end{Snugshade}

One central critique of Adam comes from an influential paper
\citep{reddi2019convergence} (the winner of {\it ICLR 2018 Best Paper Award} \citep{ICLR2018BestPaper}), where the authors provide an example that Adam diverges with a wide range of hyperparameters. A main result in  \citep{reddi2019convergence} states that: 
\begin{snugshade}
\begin{center}
   {\it \citep{reddi2019convergence}: For any $\beta_1, \beta_2$ s.t. $0 \leq \beta_1 < \sqrt{\beta_2} <1$, there exists a problem such that Adam diverges. } 
\end{center}
\end{snugshade}

\begin{figure*}[t]
  \vspace{-1cm}
    \centering
    \subfigure[Divergent region claimed by \citep{reddi2019convergence}]{
    \begin{minipage}[t]{0.24\linewidth}
    \centering
    \includegraphics[width=\linewidth]{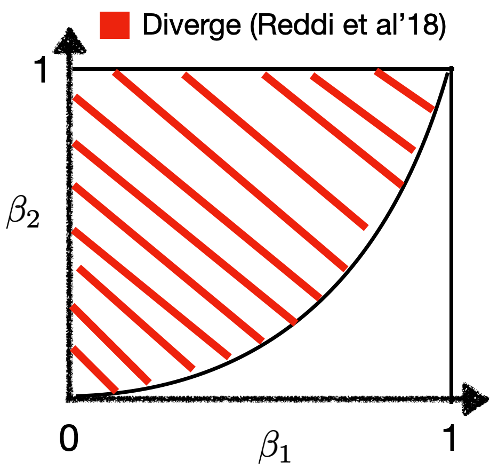}
    \end{minipage}%
    }%
    \subfigure[Our contribution]{
      \begin{minipage}[t]{0.24\linewidth}
      \centering
   \includegraphics[width=\linewidth]{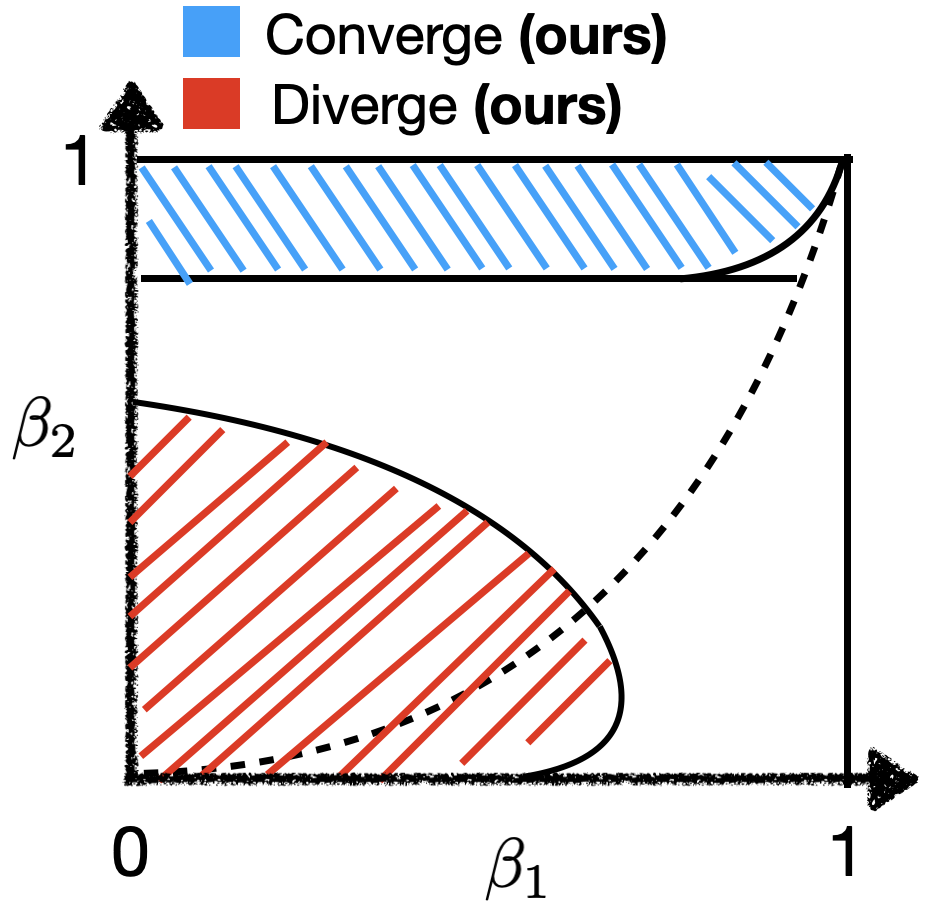}
      \end{minipage}%
      }%
    \subfigure[MNIST]{
      \begin{minipage}[t]{0.24\linewidth}
      \centering
    \includegraphics[width=\linewidth]{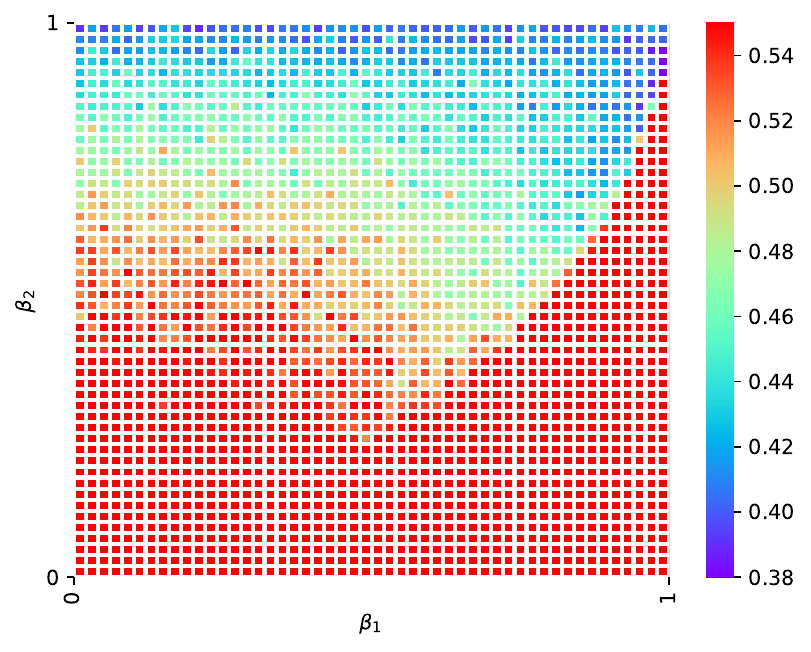}
      \end{minipage}%
      }%
    \subfigure[CIFAR-10]{
      \begin{minipage}[t]{0.24\linewidth}
      \centering
    \includegraphics[width=\linewidth]{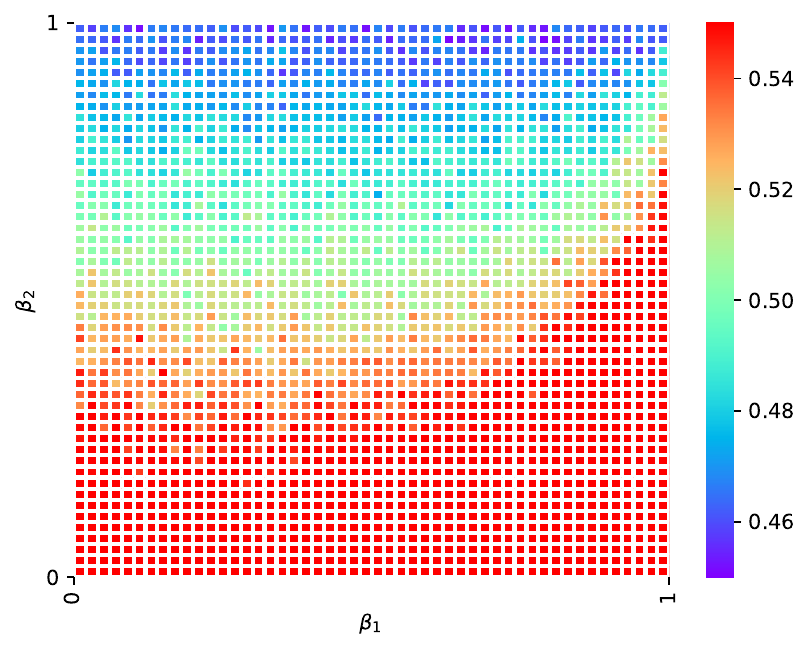}
      \end{minipage}%
      }%
    \centering
    \caption{ {\small {\bf (a)}: The divergent region of Adam claimed by \citep{reddi2019convergence}. They fix $\betabeta$ first and then pick a problem to construct the divergence example.   {\bf (b)}: An illustration of our contribution in $\betabeta$ phase diagram. We fix the problem before picking $\betabeta$.  Note that this is a different setting from {\bf (a)}, so there is no contradiction. {\bf Both boundaries of the red and blue regions depend on batch size (shown later).} The shape of the region follows the solution to our analytic conditions.  The dotted curve satisfies $\beta_1 =\sqrt{\beta_2}$. {\bf (c), (d)}:  The training loss on MNIST and CIFAR-10 over an extensive sweep of $\beta_1$ and  $\beta_2$. The performance of Adam reconciles with our theoretical characterization in {\bf (b)}.   }}%
    \label{fig:intro_paper}
\vspace{-1.5mm}
\end{figure*}

The divergence region is visualized in Figure \ref{fig:intro_paper} (a). This finding raises serious concerns for Adam's deployment in AI model training,  where the divergence can raise alerts of unpredictable training failures. Since then, many new variants have been designed. For instance, 
AMSGrad \citep{reddi2019convergence} enforces $v_k$ (defined later in Algorithm \ref{algorithm_wr}) to be non-decreasing; AdaBound \citep{luo2018adaptive} imposes a constraint $v_k \in [C_l, C_u]$ to ensure the boundedness of the effective stepsize.

On the other hand, counter-intuitively, vanilla Adam remains exceptionally popular.  Without any modification to its update rules, Adam works well in practice.  This is rather surprising due to the existence of divergence theory.
 Even more mysteriously, we find that the commonly reported hyperparameters actually satisfy the divergence condition stated earlier. 
 For instance, 
\citet{kingma2014adam} claim that  $\betabeta=(0.9,0.999)$ is a ``good choice for the tested machine learning problems'' and it is indeed the default setting in deep learning libraries. 
For LLMs such as GPT-3,  Llama series, and DeepSeek series \citep{brown2020language, touvron2023llama, liu2024deepseek}, $\betabeta$ is chosen to be $(0.9,0.95)$.
All these hyperparameters live in the divergence region $\beta_1 < \sqrt{\beta_2}$.
Surprisingly, instead of observing the divergence issue, these hyperparameters achieve good performance and they actually show signs of convergence.

 Why does Adam work well despite its theoretical divergence issue? Is there any mismatch between deep learning problems and the divergence example? 
We take a closer look into the divergence example and find out the mismatch {\it does} exist. 
In particular, \citep{reddi2019convergence} picks   $\betabeta$ {\it before} picking the problem (or precisely, the \# mini-batches $n$).
 That is, to construct the divergence example, they change $n$ for different $\betabeta$. On the other hand, in practical applications of Adam listed above, 
practitioners tune hyperparameters $\betabeta$ \textit{after} the problem (or $n$) is fixed.

Given the empirical success of Adam, we conjecture that Adam can converge when the problem is fixed. 
 To verify this conjecture, we run Adam on classification tasks on MNIST and CIFAR-10 as shown in Figure \ref{fig:intro_paper} (c) and (d). 
 While Adam's performance is unstable in the red region, we find that {\it it always performs well in the top blue region in Figure \ref{fig:intro_paper}.} This seems to suggest that: when the problem is fixed, Adam can converge without modification after proper tuning of $\beta_1$ and $\beta_2$.

In the remainder of this section, we theoretically quantify the different behaviors of Adam under different $\betabeta$. We show that vanilla Adam, without algorithmic modification, exhibits two qualitatively different regimes: 
in a \textit{safe} region of $(\beta_1,\beta_2)$ it provably converges to a neighborhood of critical points, 
whereas in a \textit{danger} region it can diverge to infinity. 
Together, these results reveal a divergence--convergence phase transition in the $(\beta_1,\beta_2)$ plane. 

We further point out that the critical boundary $(\beta_1^*, \beta_2^*)$ is problem-dependent, and particularly, dependent on {\it batch size}. This provides suggestions on how to tune $\beta_1$ and $\beta_2$: when Adam does not work well, we suggest tuning up $\beta_2$ inversely with batch size to surpass the threshold $\beta_2^*$, and then trying $\beta_1< \sqrt{\beta_2}$. Our suggestions are supported by reports from several empirical studies, which observe improved LLM training performance when applying them.

\paragraph{Problem Setup.} In our analysis, we make the following mild assumptions on $\ell_i(\theta) $ and $\ell(\theta)$  in the ERM problem \eqref{finite_sum}.
\begin{assumption} \label{assum1}
     {\bf (Lipschitz smoothness)} We assume $\|\nabla \ell_i(\theta_1) - \nabla \ell_i(\theta_2)\|_2 \leq L \|\theta_1 -\theta_2\|_2, \forall \theta_1, \theta_2 \in  \mathbb{R}^d$; $\ell(\theta)$ is lower bounded by a finite constant $\ell^*$. 
 \end{assumption}
\begin{assumption}\label{assum2}
 {\bf (Affine variance condition)} We assume $ \sum_{i=0}^{n-1}\left\|\nabla \ell_{i}(\theta)\right\|_{2}^{2} \leq D_{1}\|\nabla \ell(\theta)\|_{2}^{2}+D_{0}, \forall \theta\in \mathbb{R}^d$, where $D_1, D_0 \geq 0$ and are not both zero.
  \end{assumption}
  When $n,  L, D_0,$ and $D_1$ are fixed a priori, we define the corresponding problem class
  {\small
\begin{equation}
\functionclass 
\;:=\;
\Bigr\{
\ell(\theta)  \Big|\ 
\ \ell(\theta)=\tfrac1n\sum_{i=0}^{n-1} \ell_i(\theta),\ \theta \in \mathbb{R}^d \
\text{and Assumptions~\ref{assum1}--\ref{assum2} hold with }(L,D_0,D_1)
\Bigr\}.
\end{equation}
}
    
    The Lipschitz condition in Assumption \ref{assum1} is standard.  
    The ``affine variance'' condition in Assumption \ref{assum2} is quite general.  It is originally proposed by \citep{bertsekas2000gradient} and is later popularized by \citep{bottou2018optimization} in an equivalent form below:
    {\small 
        \begin{equation}
        \label{eq_a2_affine_variance}
       \frac{1}{n} \sum_{i=0}^{n-1}\left\|\nabla \ell_{i}(\theta) - \nabla \ell(\theta) \right\|_{2}^{2} = \frac{1}{n}\left( \sum_{i=0}^{n-1}\left\|\nabla \ell_{i}(\theta)\right\|_{2}^2 \right)  -  \|\nabla \ell(\theta)\|_{2}^2  \leq   \frac{(D_{1} -n)}{n}\|\nabla \ell(\theta)\|_{2}^{2} + \frac{D_0}{n}, \forall \theta\in \mathbb{R}^d.
    \end{equation}
    }
     Assumption \ref{assum2} is weaker than commonly-used ``bounded variance condition'', which requires $D_1 = n$ and the variance is not allowed to grow with the gradient $\|\nabla \ell(\theta)\|_{2}^{2}$. Please note that  $D_1 = n$ is rather restricted. {\bf First},  it does not even hold on convex quadratic minimization, and we provide a proof in Appendix A.2 in \citep{zhang2026adam}.  {\bf Second}, as we will elaborate later, it excludes some divergence counterexamples of SignSGD (which is equivalent to Adam with $\beta_1 = \beta_2 = 0$). As such, analysis under $D_1 = n$ may not reveal the full picture of Adam.  
     
     Here, we allow generic $D_1 \geq 0$. Assumption \ref{assum2} is also recommended in the influential survey \citep{bottou2018optimization} as it is {\it ``relatively minor''} and {\it ``variance is allowed to grow quadratically in any direction.''} We provide more details on the validity of Assum.~\ref{assum2}  in Sec. 2.1 in \citep{zhang2026adam}.

Finally, we emphasize that we do {\it not} add bounded gradient assumption {\small $\|\nabla \ell(\theta)\|\leq G$}, which is commonly used in the literature of stochastic gradient methods. 
This is crucial for revealing the phase transition: with gradients bounded a priori, Adam cannot diverge, while we prove that it can happen under certain $\betabeta$.

\subsection[A Brief Review of the Counterexample]{A Brief Review of  \citep{reddi2019convergence}}
\label{sec_reddi}

Before stating our theoretical results, we first restate the counterexample by \citep{reddi2019convergence}. 
They consider the following one-dimensional convex problem:  $\operatorname*{minimize} \frac{1}{n} \sum_{i=0}^{n-1} \ell_i(\theta)$ where $\theta \in [-1,1]$, $n \geq 3$:
	{\small
	\begin{equation}\label{counterexample_reddi}
			\ell_{i}(\theta)=\left\{\begin{array}{ll}n \theta, & \text { for } i =0 \\ -\theta, & \text { otherwise, }\end{array}\right.
	\end{equation}
	}

Note that \eqref{counterexample_reddi} satisfies both Assumption \ref{assum1} and \ref{assum2}  (with $D_1= n^4 +n^3 -n^2$ and $D_0 =0$), so our assumptions do not rule out this counterexample a priori.
This is a constrained problem with feasible set $\theta \in [-1,1]$; the optimal solution is $\theta^*=-1$.   We restate their main claim as follows.

\begin{theorem}[Theorem 2 in \citep{reddi2019convergence}]
For any fixed $\betabeta$ satisfying $\beta_1 < \sqrt{\beta_2}$, there exists a sufficiently large $n$ such that Adam applied to the function \eqref{counterexample_reddi} (under cyclic sampling) converges to the sub-optimal point $\theta=1$.
\end{theorem}
We briefly discuss the divergence condition for this Theorem.  As stated in Eq. (7), Appendix B in \citep{reddi2019convergence},  for every fixed $\betabeta$, they need a ``constant $n$ that depends on $\beta_1$ and $\beta_2$''. 
As such, they require different $n$ to cause divergence on different $\betabeta$, which means the considered problem class is changing with  $\betabeta$.
\begin{figure}[h!]
\begin{center}
\subfigure[Cyclic update order]{
    \begin{minipage}[t]{0.3\linewidth}
    \centering
    \includegraphics[width=\textwidth]{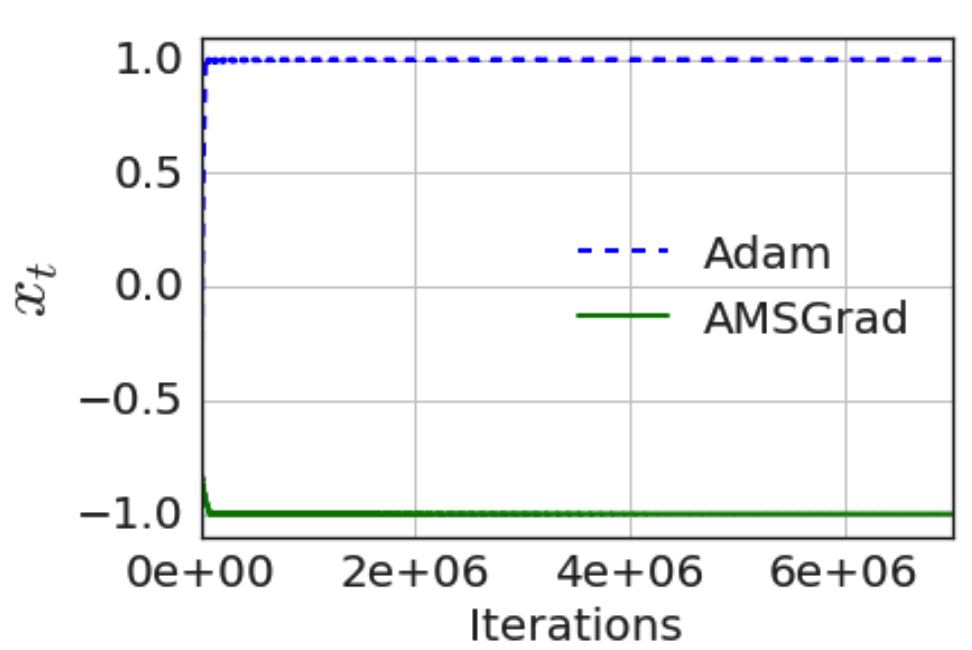}
    \end{minipage} }%
    \subfigure[Randomized update order]{
    \begin{minipage}[t]{0.3\linewidth}
    \centering
    \includegraphics[width=\linewidth]{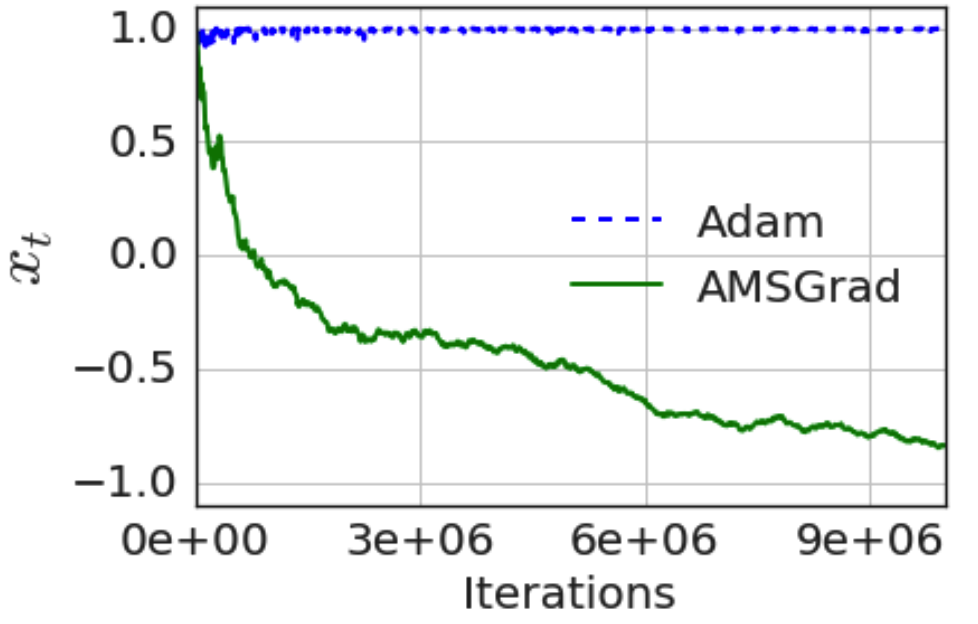}
    \end{minipage}%
    }%
    \subfigure[Diminishing stepsize]{
    \begin{minipage}[t]{0.28\linewidth}
    \centering
    \includegraphics[width=\linewidth]{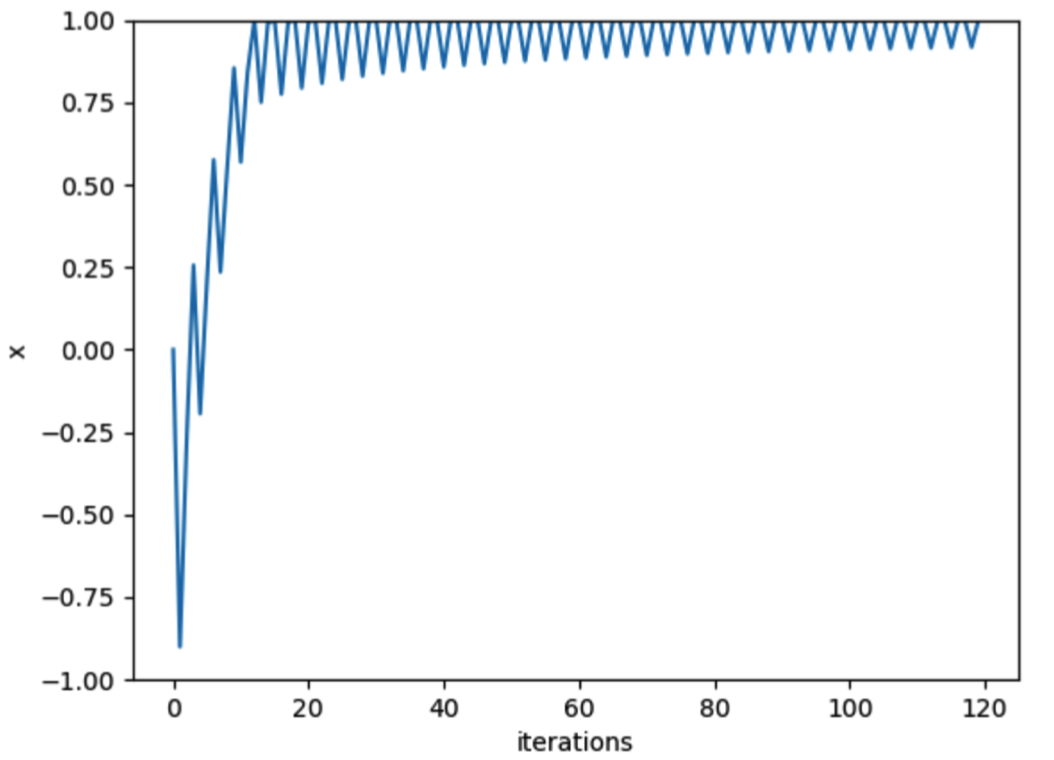}
    \end{minipage}%
    }%
\caption{{\small 
{\bf (a, b):} We restate Figure 1 from \citep{reddi2019convergence}. The divergence of Adam happens under both cyclic and random update orders, so randomization cannot prevent divergence.  Since they consider constrained problems, the term ``divergence'' here means getting stuck at the sub-optimal solution $\theta=1$.  In the figures, AMSGrad is a different method, and it is not our focus.  {\bf (c):} Diminishing stepsize $\eta_k = \frac{1}{\sqrt{k}}$ does not prevent divergence.}
}
\label{fig:reddi_figures}
\vspace{-2mm}
\end{center}
\end{figure}

We note that these divergence results also hold when randomized update orders are used instead of cyclic orders, as proved in \citep[Theorem 3]{reddi2019convergence}. Consequently, randomization does not prevent divergence. This is different from the classical di-convergence debate of multi-block ADMM, where in that case randomization can serve as an effective remedy to prevent divergence \citep{chen2016direct,sun2020efficiency}. Further, the proof of \citep{reddi2019convergence} uses $\eta_k \propto \frac{1}{\sqrt{k}}$, so diminishing stepsize does not prevent divergence, either. These claims are supported by numerical evidence in Figure \ref{fig:reddi_figures}.

To prevent divergence, \citet{reddi2019convergence}  suggested {\it ``  using...a different $\epsilon$, $\beta_1$, and $\beta_2$ for each dimension. However, this defeats the purpose of adaptive methods since it requires tuning a large set of parameters.''}
In the following, we show that in our setting such divergence can be avoided without changing the algorithm or introducing new hyperparameters.
The key is to choose $\betabeta$ in a problem-dependent manner, with a particular dependence on the number of mini-batches $n$ (equivalently, the batch size).
Importantly, this does \textit{not} require per-coordinate tuning of $\betabeta$---a direction mentioned in the discussion of \citet{reddi2019convergence}.

\subsection{Main Results}
\label{sec_conv_wr}
First, we prove that Adam converges with proper problem-dependent hyperparameters.

\begin{theorem}
\label{thm_wr}
{\bf (Convergence result under large $\beta_2$)}
Assume Algorithm \ref{algorithm_wr} satisfies:
{\small $0\leq\beta_1 < \sqrt{\beta_2} <1$};  $\beta_2$ satisfies {\small$ \beta_2 \in [0,1), \beta_2 \geq   {\gamma_1(n): = 1- \mathcal{O}\left(\frac{1-\beta_1^n}{n^5}\right)}$}; and {\small $\eta_k = \frac{\eta_0}{\sqrt{k}}$}. For any {\small $\ell(\theta) \in \functionclass$}, when   $T \in  \mathbb{N}$ is sufficiently large, we have
\begin{equation}
\label{eq_thm_wr}
  \min_{k \in [1, T]} \Ex\left[\min  \left\{ \frac{\|\nabla \ell(\theta_{k})\|_2^2 }{\sqrt{D_{0}} }  , \frac{\|\nabla \ell(\theta_k)\|_2}{2\sqrt{d D_1 }}\right\} \right]
   =\mathcal{O}\left(\frac{\log T }{(1-\beta_2)\sqrt{T}} \right)  + \mathcal{O}((1-\beta_2)\sqrt{D_0}).
\end{equation}
\end{theorem}

\begin{figure*}[h]
    \centering
    \subfigure[batch size vs. $\beta_2$]{
    \begin{minipage}[t]{0.3\linewidth}
        \centering
        \includegraphics[width=\linewidth]{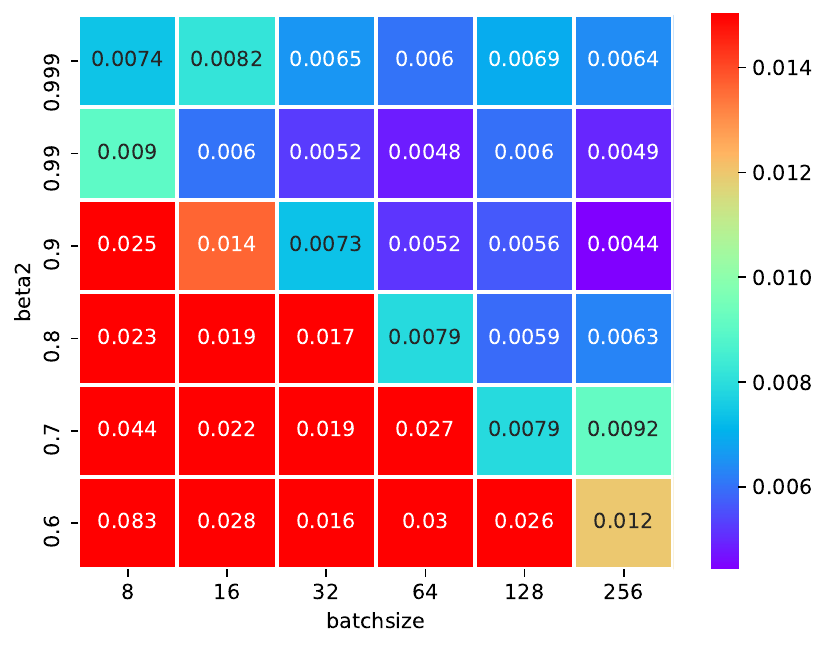}
    \end{minipage}    
    }%
    \subfigure[Large $\beta_2$ and $D_0>0$]{
      \begin{minipage}[t]{0.3\linewidth}
      \centering
    \includegraphics[width=\linewidth]{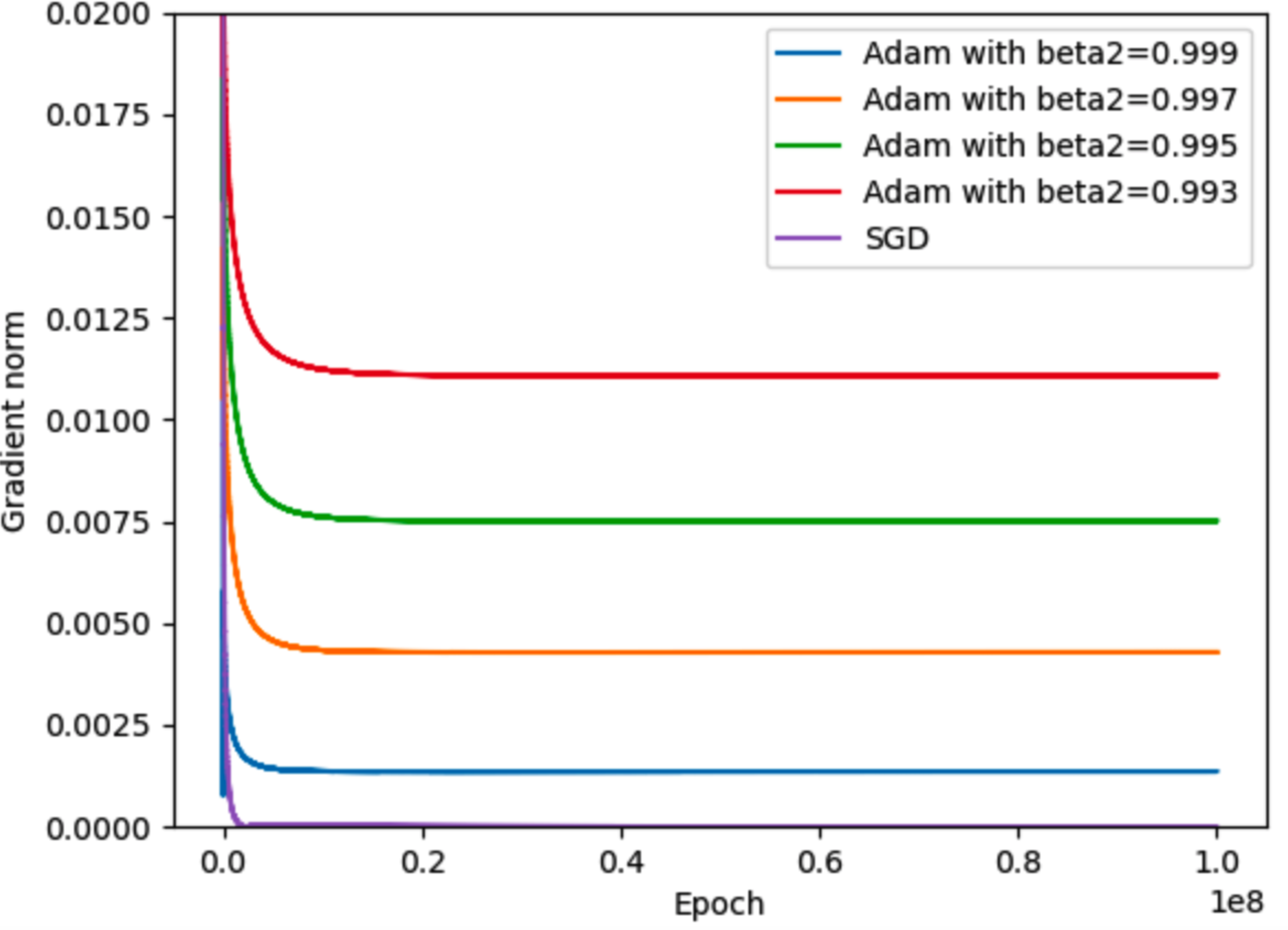}
      \end{minipage}%
      }%
    \subfigure[Large $\beta_2$ and $D_0= 0$]{
    \begin{minipage}[t]{0.3\linewidth}
    \centering
\includegraphics[width=\linewidth]{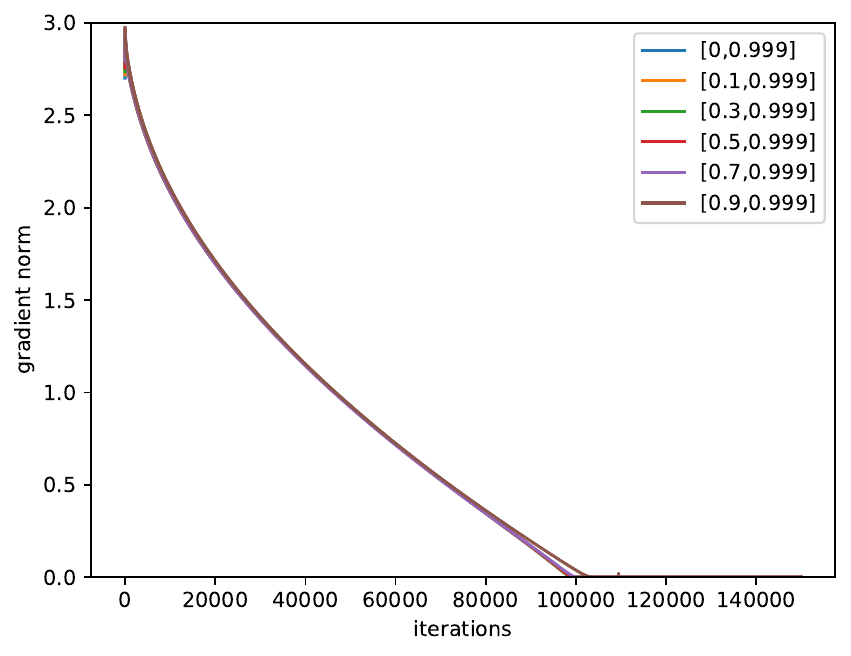}
    \end{minipage}%
    }%
    \centering
    \caption{{\small {\bf (a):} The training
loss of Adam on MNIST under different batch size and $\beta_2$. The trend aligns with our theory: we need a larger $\beta_2$ when batch size is small. Here, we used the default $\beta_1=0.9$. 
    {\bf (b) (c):} 
  large-$\beta_2$ Adam converges to a neighborhood of critical points when $D_0>0$ and converges to exact critical points when $D_0 = 0$. We use diminishing stepsize $\eta_k = 0.1 /\sqrt{k}$ as in our theory. The labels in (c) stand for $[\beta_1,\beta_2]$.
    }}
    \label{fig:cifar_nlp_mnist}
\end{figure*}

Thereom \ref{thm_wr} provides a positive result that Adam is ``safe'' under large and problem-dependent $\beta_2$.
Second, we prove a negative result that small-$\beta_2$ Adam can diverge, and the divergence region is also problem-dependent.  
The divergence of small-$\beta_2$ Adam suggests that ``large $\beta_2$'' is necessary for Theorem \ref{thm_wr}.    It is worth mentioning that Sign-SGD is a special case of Adam with $\beta_1 =\beta_2 =0$, so the following divergence result applies to Sign-SGD as well.

\begin{theorem}\label{thm_diverge} 
{\bf (Divergence result under small $\beta_2$)}
For any  problem class $\functionclass$ with $n \geq 3$, $D_1 \geq 2n^2$, and $D_0 \geq 0$, there exists a $\ell(\theta) \in \functionclass$, s.t. when $\betabeta$ satisfies our analytic divergence conditions, we have 
{\small \begin{equation}
    \|x_{k,i}\|, \|\nabla f(x_{k,i})\|, |f(x_{k,i})|  \to+\infty.
\end{equation}}
The size of the region depends on $n$ and it expands as $n$ grows to infinity. 
\end{theorem}
\begin{figure}[h]
\begin{center}
\subfigure[Divergence region]{
    \begin{minipage}[t]{0.24\linewidth}
    \centering
    \includegraphics[width=\textwidth]{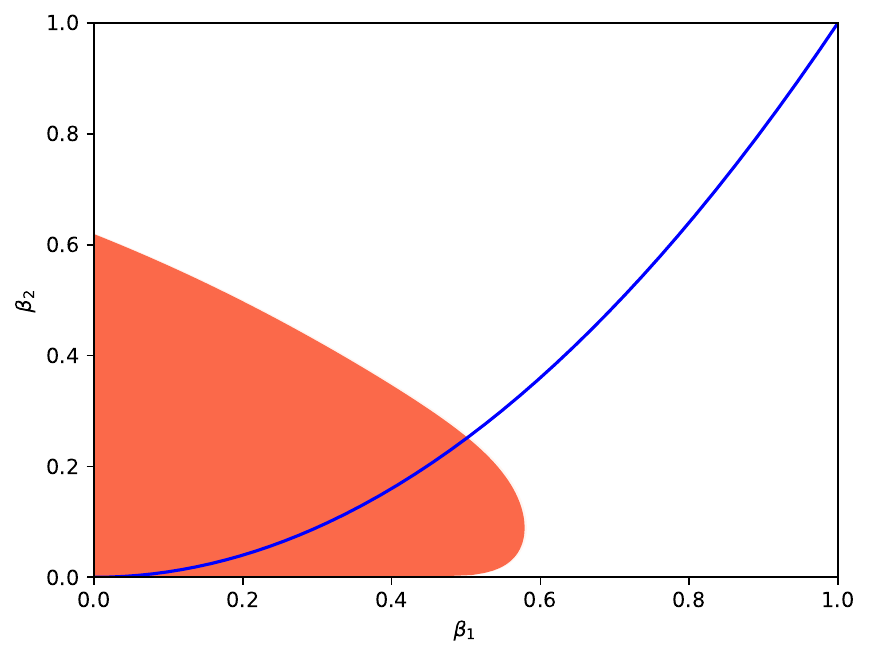}
    \end{minipage} }%
    \subfigure[The effect of small $\beta_2$]{
    \begin{minipage}[t]{0.24\linewidth}
    \centering
    \includegraphics[width=\linewidth]{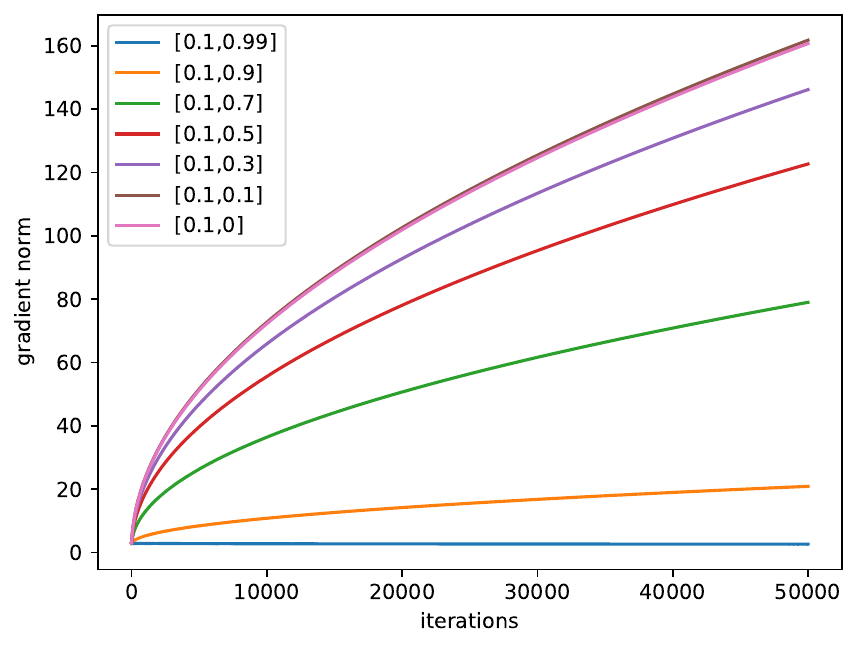}
    \end{minipage}%
    }%
    \subfigure[$n=10$]{
      \begin{minipage}[t]{0.24\linewidth}
      \centering
    \includegraphics[width=\linewidth]{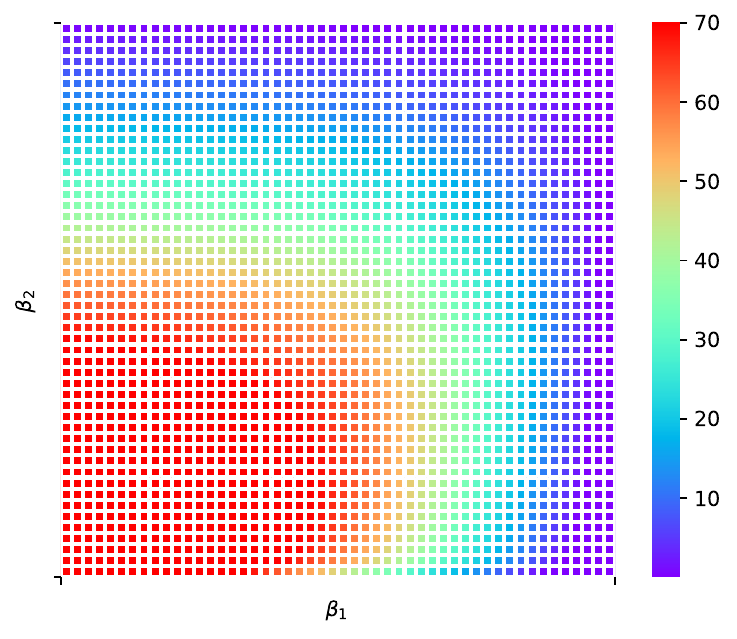}
      \end{minipage}%
      }%
    \subfigure[$n=50$]{
      \begin{minipage}[t]{0.24\linewidth}
  \includegraphics[width=\linewidth]{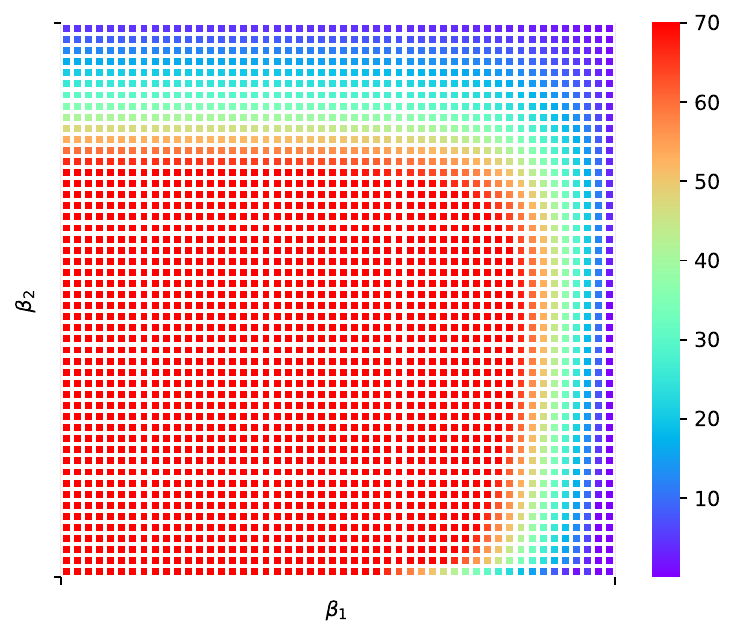}
      \end{minipage}%
      }%
\caption{{\small 
{\bf (a):} On our counterexample with  $n=20$, Adam diverges to infinity in the colored region. The region is plotted by solving our analytic divergence conditions in \texttt{NumPy}.
The blue curve satisfies  $\beta_1 = \sqrt{\beta_2}$.  {\bf (b):} When $\beta_2$  is small, Adam diverges to infinity. The labels in (b) stand for $[\beta_1,\beta_2]$.  {\bf (c, d)}  The optimality gap $|\theta-\theta^*|
        $ after running 50k iterations of Adam on our counterexample under $n = 10$ and $50$.}
}
\label{fig:counter_boundary_paper}
\vspace{-2mm}
\end{center}
\end{figure}

Our counterexample in Theorem \ref{thm_diverge} is constructed based on a 1D quadratic objective, which we present below.  \yushunrevise{ Consider
{\small $f(x)= \frac{1}{n}\sum_{i=0}^{n-1} f_i(x)$} for $n \geq 3$ and
$x \in \mathbb{R}$, and define $f_i(x)$ as:
{\small \begin{equation}
f_{i}(x)=
\left\{
\begin{array}{ll}
\frac{1+(n-1)a}{2}x^2,
& i=0, \\[2mm]
-\frac{a}{2}x^2,
& i=1,\ldots,n-1,
\end{array}
\right.
\label{counterexample1}
\end{equation}}
where $a>0$ is arbitrary finite positive constant. Summing up all the $f_i(x)$, one can see that
{\small \begin{equation}
  f(x)=\frac{x^2}{2n}.
\end{equation}}
Thus, for any finite positive $a>0$, $f(x)$ belongs to $\functionclass$.
For instance, when $a=1/(n-1)^2$, we have $D_1 =
\frac{19}{8}n^2$, $D_0=0$, and $f(x)$ is lower bounded with optimal
$x^*=0$. Further, it satisfies Assumption \ref{assum2}, but it does {\it not}
satisfy the bounded variance  which
restricts $D_1$ to be $n$.
}

Function \eqref{counterexample1} is modified based on \citep{reddi2019convergence}, but we allow iterates, gradients, and function values to diverge to infinity.  We note that Sign-SGD is a special case of Adam with $\beta_1 =\beta_2 =0$, so the divergence result applies to Sign-SGD as well.

\begin{figure}[t]
\begin{center}
\vspace{-0.5cm}
      \subfigure[$\beta_1 =0, \beta_2 = 0$]{
        \begin{minipage}[t]{0.45\linewidth}
        \centering
        \includegraphics[width=\linewidth]{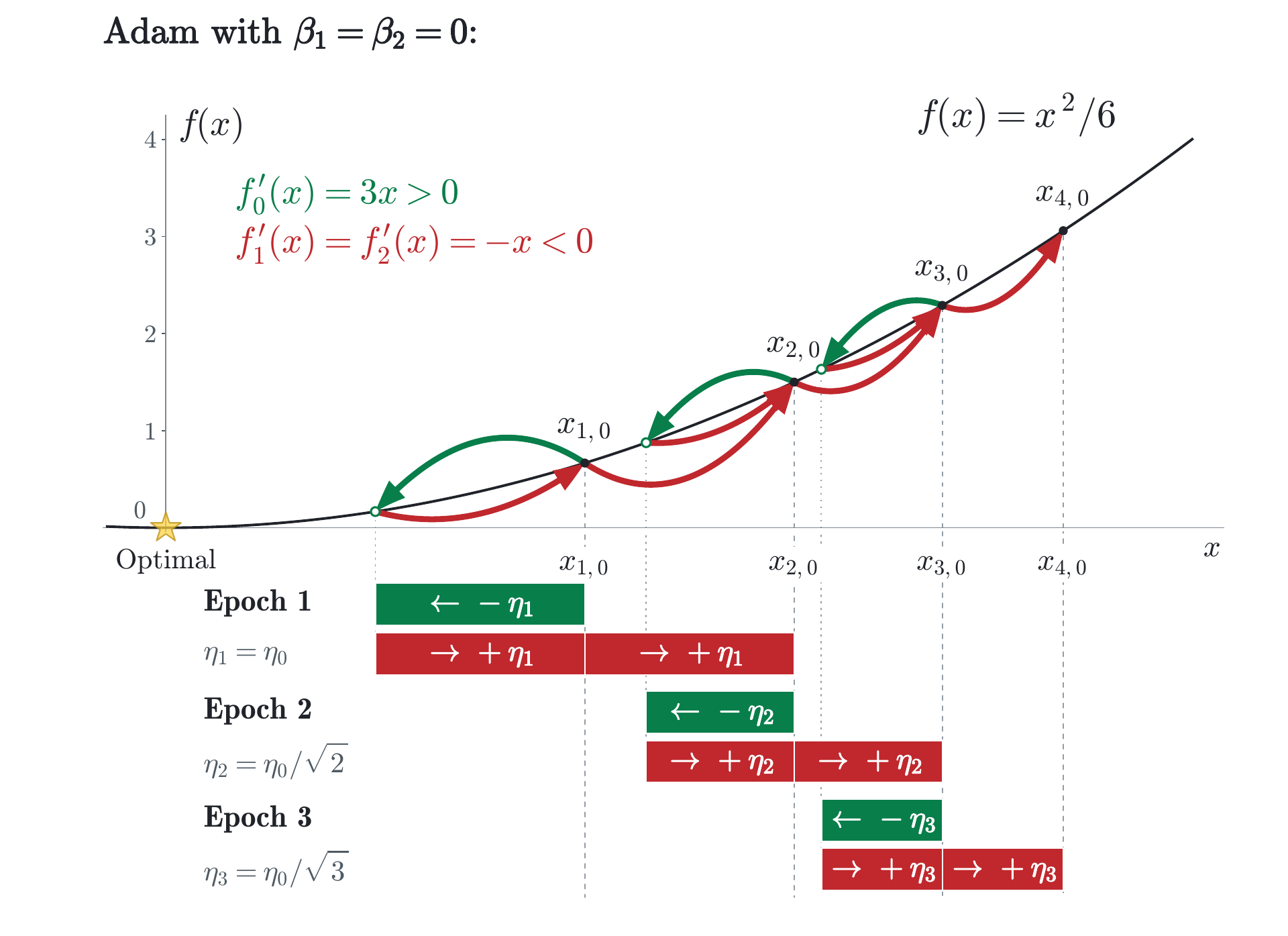}
        \end{minipage}%
        }%
        \subfigure[$\beta_1 =0, \beta_2 = 0.999$]{
        \begin{minipage}[t]{0.45\linewidth}
        \centering
        \includegraphics[width=\linewidth]{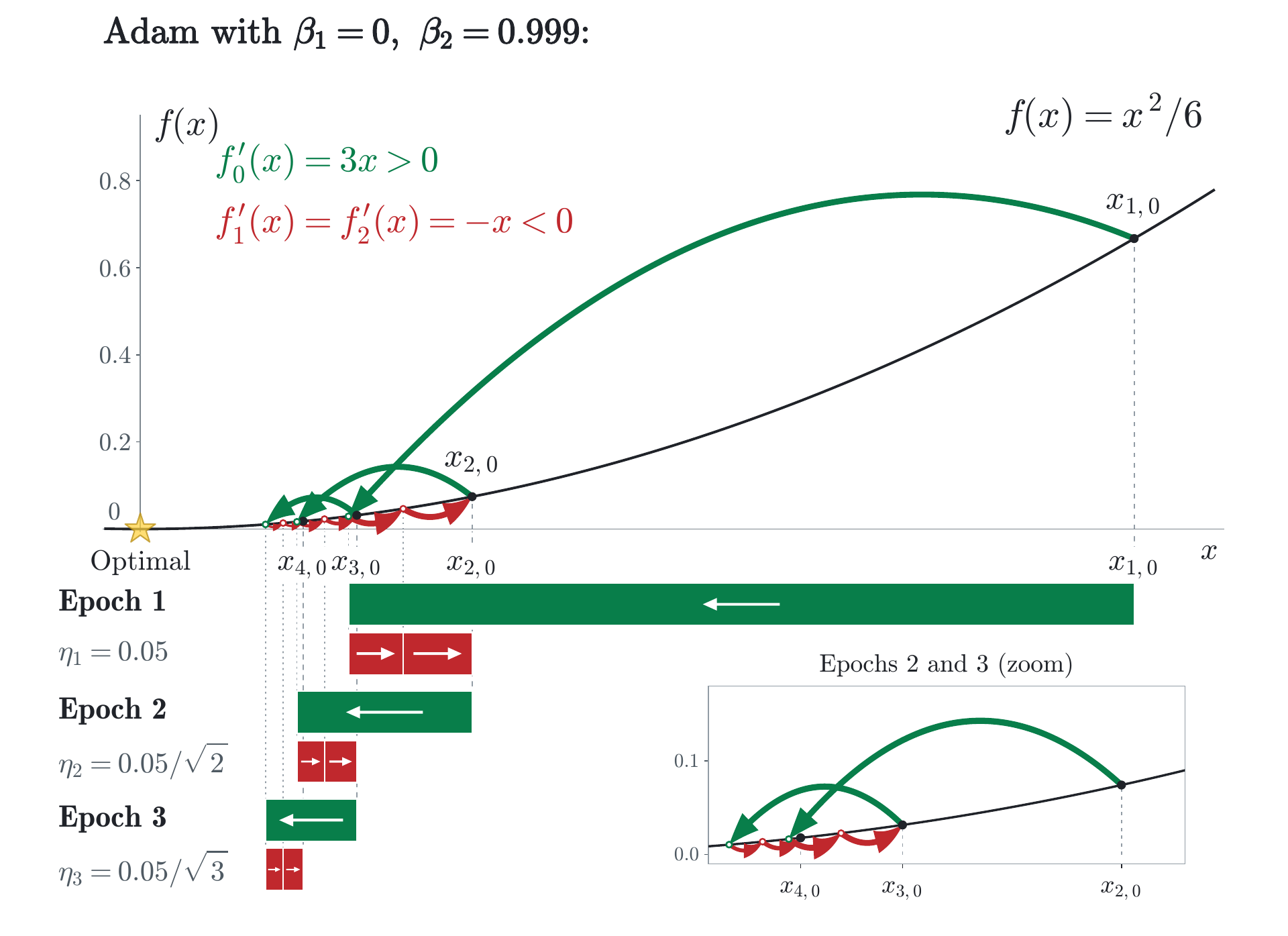}
        \end{minipage}%
        }%
        \\
        \subfigure[$\beta_1 =0.5, \beta_2 = 0.999$]{
        \begin{minipage}[t]{0.45\linewidth}
        \centering
        \includegraphics[width=\linewidth]{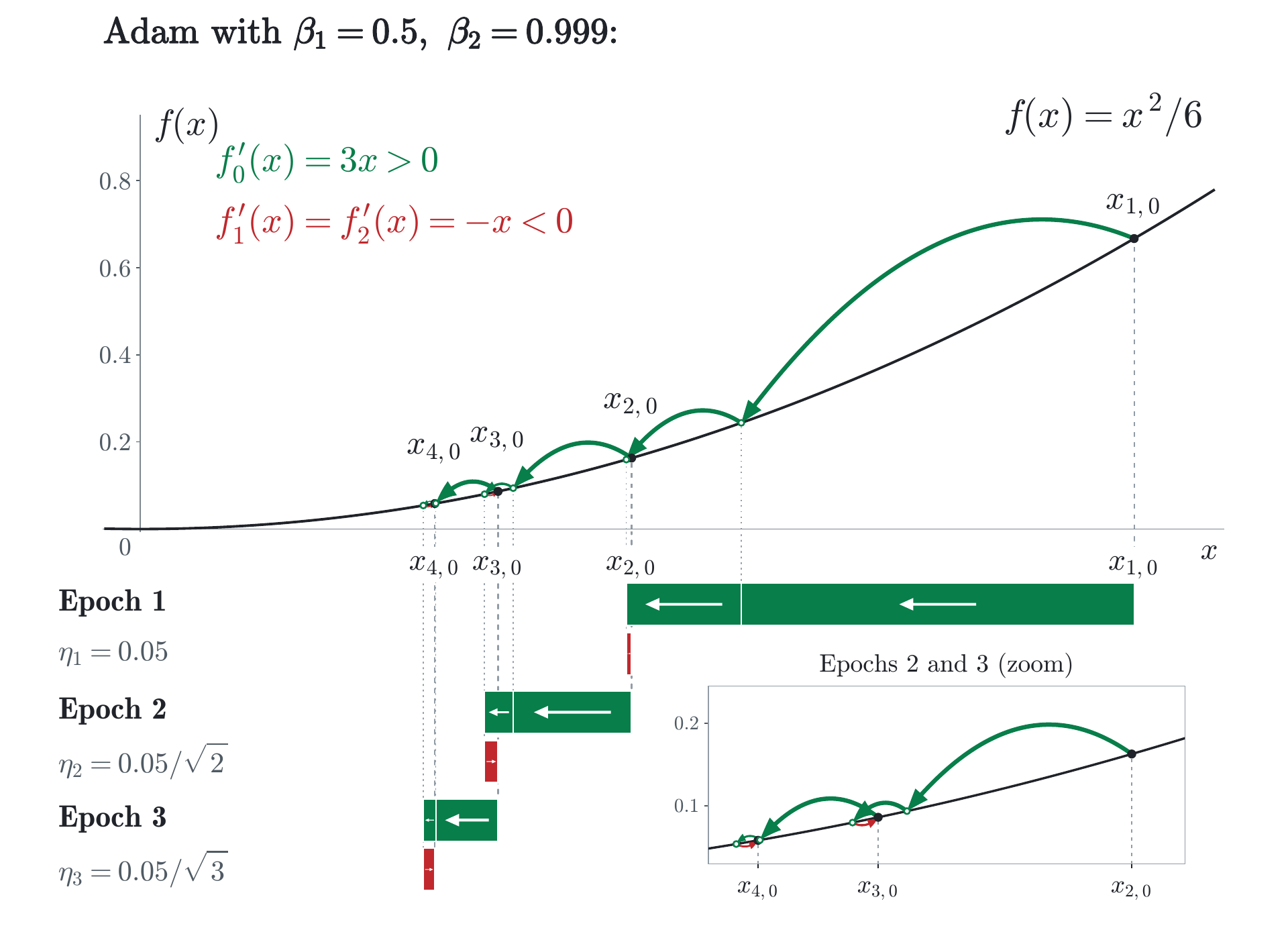}
        \end{minipage}%
        }%
        \subfigure[$\beta_1 =0.9, \beta_2 = 0.999$]{
        \begin{minipage}[t]{0.45\linewidth}
        \centering
        \includegraphics[width=\linewidth]{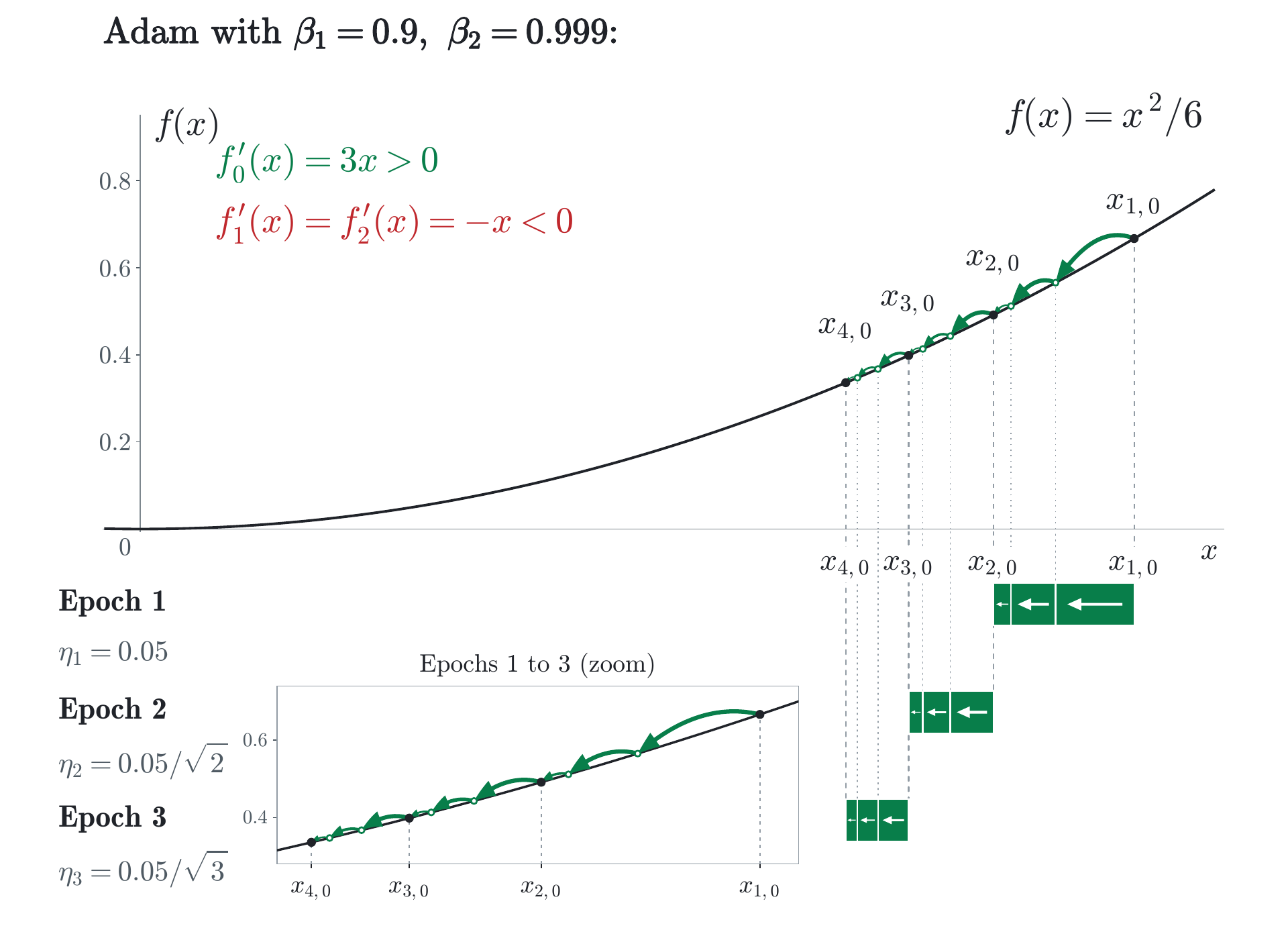}
        \end{minipage}%
        }%
\caption{
{\small 
{\bf (a,b,c,d):} The dynamics of Adam on counterexample \eqref{counterexample1} with $a = 1$ and $n =3$ under different $\beta_1$ and $\beta_2$. The optimal $x^* =0$.  {\bf (a)}: Adam diverges. {\bf (b, c, d)}: Adam converges. The divergence happens when the red forces outweigh the green forces, which happens when $\beta_2$ is small or equals 0 in {\bf (a)}.  The green forces become dominant when $\beta_2$ is large in {\bf (b)}.  Compared with {\bf (b)},    $\beta_1 = 0.5$  in {\bf (c, d)} further helps shrink the red forces.
}
}
\label{fig:divergence_figure_paper}
\end{center}
\end{figure}

\yushunrevise{To flush out the insight behind the counterexample,   we present the dynamics of Adam on counterexample \eqref{counterexample1} in Figure \ref{fig:divergence_figure_paper}. Here, the divergence happens when the red forces (towards right) outweigh the green forces (towards left), and the convergence happens if otherwise. When both $\beta_1$ and $\beta_2$ are small as in (a), the fast movement of $v_k$ provides up-to-date normalization, which weakens the strength of the green forces, causing divergence. The divergence can be avoided by using a larger $\beta_2$ as in (b), where the changes of $v_k$ is slowed down, and green forces dominate the red forces. In (c, d), a positive $\beta_1$ such as  $0.5$ or $0.9$ further helps shrink the red forces.}

Combining the positive result (Theorem \ref{thm_wr}) and negative result (Theorem \ref{thm_diverge}) together,  we establish a phase transition for Adam from divergence to convergence when changing the $(\beta_1, \beta_2)$ combination. To our knowledge, this is the first convergence guarantee for unmodified Adam, and the first phase transition in the $(\beta_1,\beta_2)$ 2D plane reported in the literature, providing rigorous theoretical guarantees for Adam.

  \paragraph{Remark 1: convergence to a neighborhood of critical points.}
  When {\small $D_0>0$}, Adam converges to a neighborhood of critical points, in lieu of the exact critical points. We emphasize that this is {\it not} due to the limitation of the analysis, and this phenomenon is also observed in practice: even for simple convex quadratic function with  {\small $D_0>0$},  Adam with diminishing stepsize {\it cannot} reach exactly zero gradient (see Figure \ref{fig:cifar_nlp_mnist} (b)). 
  In the non-realizable case ({\small $D_0>0$}),  converging to the neighborhood of critical points is common for stochastic methods, including constant-stepsize SGD  \citep{luo1991convergence,bertsekas1997nonlinear, yan2018unified,yu2019linear,liu2020improved} and diminishing-stepsize RMSProp \citep{Manzil2018adaptive,shi2020rmsprop}.

  In the realizable case (i.e. $D_0=0$), Theorem \ref{thm_wr}  states that Adam can converge to critical points.  This matches our numerical results in Figure \ref{fig:cifar_nlp_mnist} (c).   The convergence rate in Theorem \ref{thm_wr} is comparable to that of SGD under the same condition.

  \paragraph{Remark 2: $\beta_2$ and batch size.}  The condition of $\beta_2$: {\small $\beta_2 \geq \gamma_1(n) = 1- \mathcal{O}\left(\frac{1-\beta_1^n}{n^5}\right)$} is problem-dependent and it increases with \# mini-batches $n$. This suggests that we need larger $\beta_2$ when $n$ is large, or equivalently, we need larger $\beta_2$ when the batch size, which equals $\frac{\mathcal{D}}{n}$, is small. ($\mathcal{D}$ denotes the total sample size.) 
  This aligns with our experiments in Fig. \ref{fig:cifar_nlp_mnist} (a):  on MNIST, smaller batch size $\frac{\mathcal{D}}{n}$ (which is equivalent to larger $n$) requires a larger $\beta_2$ to reach small loss.     Note that our threshold of $\beta_2$ is not claimed tight.  Tightening the threshold for $\beta_2$ will be an interesting future direction.

\paragraph{Remark 3: guidance to LLM pre-training.} Our divergence-convergence theory indicates that larger $\beta_2$ is  necessary when $n$ is large. This equivalently indicates that a larger $\beta_2$ is required when the batch size, which equals $\frac{\mathcal{D}}{n}$, is small. This message can provide guidance for hyperparameter tuning in LLM pre-training, as confirmed by various literature.  We list some numerical evidence as follows.
\begin{itemize}
    \item \citet{sreckovic2025your}: {\it ``\citet{zhang2022adam} shows that larger $\beta_2$  substantially improves small-batch training, and  \citet{marek2025small} highlights the importance of scaling $\beta_2$ in this regime.''} 
    \item  \citet{zhang2024does} emphasize that  {\it ``larger $\beta_2$ (increase from 0.95 to 0.99 or 0.999) substantially improves small batch size training ... aligning with findings in \citep{zhang2022adam}''}.  
    \item \citet{porian2024resolving} report that {\it ``enlarging $\beta_2$ (from 0.95 to 0.999) is essential at lower batch sizes ... This matches the theoretical work \citep{zhang2022adam}.''}
\end{itemize}

\paragraph{Remark 4: Is larger $\beta_2$ always better?}
To ensure convergence, Theorem \ref{thm_wr} states that $\beta_2$ needs to be larger than a batch-size-dependent threshold $\beta_2^*$. But what is the optimal $\beta_2$  within the range of $[\beta_2^*, 1)$? We provide an initial discussion based on our current results. 

In Theorem \ref{thm_wr}, the rate includes the constant  $\frac{1}{1-\beta_2}$, which is magnified as $\beta_2 \rightarrow 1$.  We note that such dependency is somewhat unavoidable as it is embedded in the design of Adam: when $\beta_2 = 1$ and initial $v_0= 0$, $v_k$ always equals $0$ and Adam is not well-defined. Consequently, our theoretical bound must naturally capture this worst-case singularity. Nevertheless, the bound in \eqref{eq_thm_wr} remains strictly bounded and valid provided that $\beta_2 \in [0,1)$. In practical scenarios such as $\beta_2 =0.95$ or $ 0.999$, this corresponds to a finite multiplier of 20 or 1000. 

Theorem \ref{thm_wr} suggests that: within the convergence range $[\beta_2^*, 1)$, a larger $\beta_2$ is not necessarily better.  This aligns with the empirical observation in trend in LLM where: when Adam converges under $\beta_2 = 0.95$, a larger $\beta_2$ such as $0.999$ can yield worse performance \citep{wortsman2023small,bai2025adaptive}. 

Theoretically, finding the optimal $\beta_2$ requires substantially more effort, and we leave it as a future direction.
\yushunrevise{Empirically, \citep{marek2025small} suggests a heuristic strategy, which can be helpful for practitioners: starting from the popular choice, e.g., $\beta_2 = 0.95$, and if we were to scale the batch size from $B_1$ to $B_2$,  change $\beta_2$ to $\beta_2^{B_2/B_1}$. This strategy aligns with our theory. This tuning strategy is recently adopted in Marin -- an open LLM pretraining project \citep{held2026delphi}.}

\paragraph{Remark 5: implication to SignSGD and Muon.} Theorem \ref{thm_diverge} directly implies that SignSGD (Adam with $\beta_1 = \beta_2 = 0$) can diverge when $n \geq 3$. Further, since SignSGD is Muon \citep{jordan2024muon} with $\beta= 0$ and matrix size equals 1, the same divergence also applies to Muon under this configuration. One can also easily extend the counterexample \eqref{counterexample1} to the matrix variables using diagonal matrices. As such, we conjecture there is also a divergence-convergence phase transition of Muon under problem-dependent $\beta$.

We note that the bounded variance assumption  (restrict $D_1 =n$)  excludes our counterexample \eqref{counterexample1} from the considered function class,  and SignSGD will converge to the neighborhood of critical points with $\eta_k = \eta_0 /\sqrt{k}$ in that case. However, the analysis under bounded variance does not reveal the true divergent behavior of SignSGD. Our counterexample, which satisfies Assumption \ref{assum2} (with $D_1 \geq  2n^2$) but not bounded variance, provides additional motivation to relax it. An intriguing open question is whether there exist more counterexamples with $D_1 \in (n, 2n^2)$. We leave it for future investigation.
 
\paragraph{Acknowledgment from the authors of Adam.} 
Finally, our result is mentioned by the authors of Adam in the \href{https://iclr.cc/virtual/2025/test-of-time/37599}{\BLUE{[ICLR 2025 Test of Time speech]}}, for \href{https://zyushun.github.io/images/speech_adam_can_converge.jpeg}{\BLUE{[grounding Adam theoretically]}}.

\clearpage
\subsection{Key Lemmas for the Convergence Result in  Theorem \ref{thm_wr}}
\label{sec_proofidea_wr}

Here, we summarize the key challenges behind Theorem \ref{thm_wr}.

\begin{figure}[h!]
\begin{center}
\centerline{\includegraphics[width=0.45\textwidth]{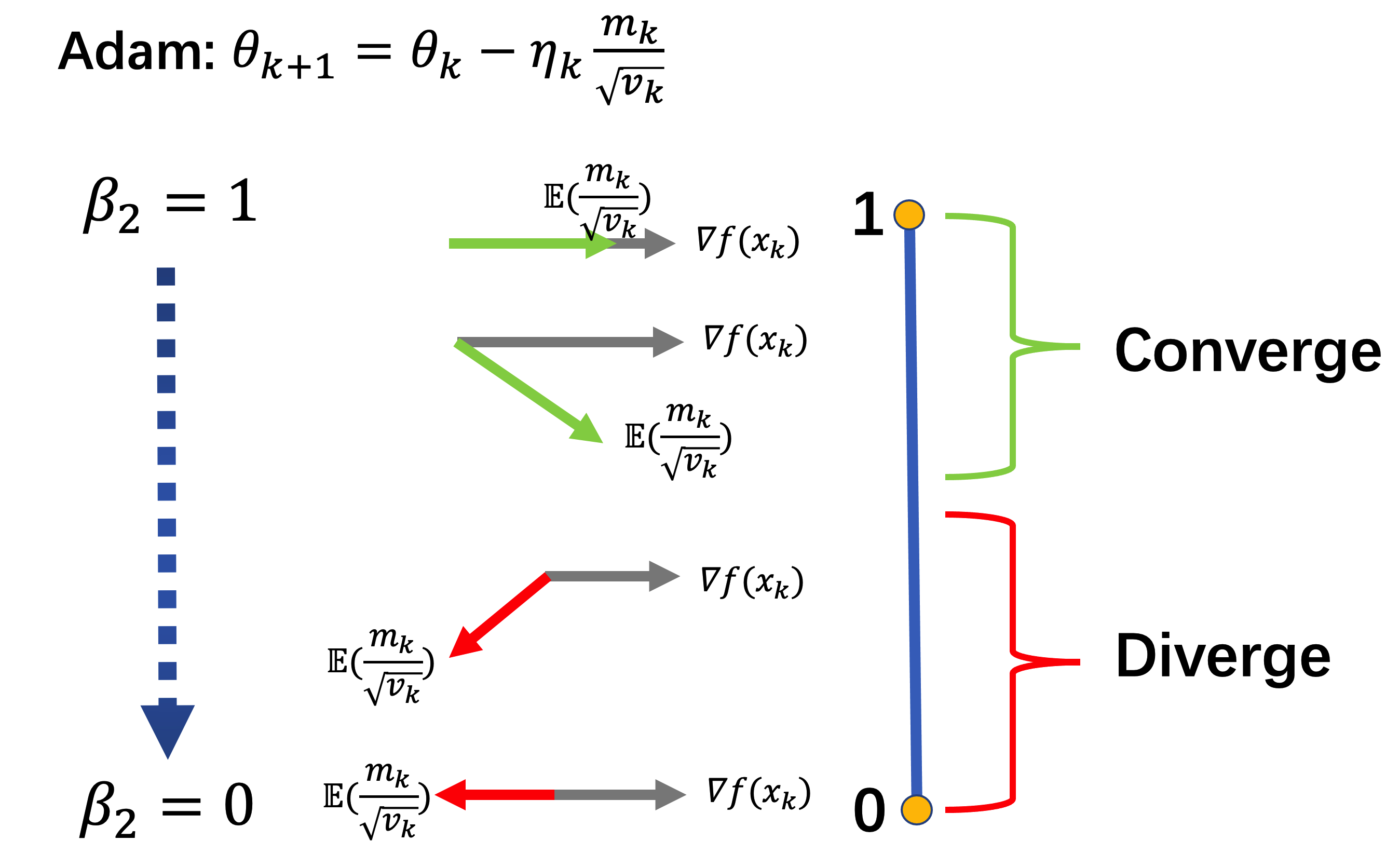}}
\caption{{\small An illustration of the changes of Adam's update direction when $\beta_2$ changes. } }
\label{fig:update_direction}
\end{center}
\vspace{-1cm}
\end{figure}

Our key insights are illustrated in Figure \ref{fig:update_direction}: we find that Adam's update direction  $\Ex \left(\frac{m_k}{\sqrt{v_k}}\right)$ is close to $\nabla \ell(\theta_k)$ when $\beta_2$ is large (leading to convergence) and starts to deviate to the opposite direction when $ \beta_2$ is small (causing divergence).

\paragraph{Concentration effects of $\frac{1}{\sqrt{v_k}}$ when $\beta_2$ is large.} On the technical side, we find that $\frac{1}{\sqrt{v_k}}$ concentrates around $\frac{1}{\sqrt{\Exk (v_k)}}$ when $\beta_2$ is large. In particular, we prove that:
{\small
\begin{equation}
\label{eq_concentration_informal}
    \frac{1}{\sqrt{v_k}} \approx  \frac{1}{\sqrt{\Exk (v_k)}},  \text{ w.h.p., when $\beta_2$ is large.}
\end{equation}
}
 As a result, $-\frac{\nabla \ell_{\tau_k}(\theta_k)}{\sqrt{v_k}}$ becomes a descent direction in this case, i.e.,
$$\Exk\left(\frac{\nabla \ell_{\tau_k}(\theta_k)}{\sqrt{v_k}}\right) \approx \frac{\Exk \left(\nabla \ell_{\tau_k}(\theta_k)\right)}{\sqrt{\Exk (v_k)}} = \frac{\nabla \ell(\theta_k)}{\sqrt{\Exk (v_k)}}, $$ 
which lies in the dual cone of the gradient direction $\nabla \ell(\theta_k)$. {\bf Why does large $\beta_2$ help?} Intuitively, this is because large $\beta_2$ slows down the changes of $v_k$, and its behavior will become largely predictable.  
It requires additional effort to handle the momentum $m_k$. We handle it via a new potential function, and we omit the details here for brevity.

The major challenge in the analysis is that  $v_k$ is a random variable and it appears in the denominator.  This makes the entire system a stochastic non-linear dynamic system, which is difficult to analyze in general.  Further, $v_k$ can potentially hit 0, which imposes extra difficulties.

We prove \eqref{eq_concentration_informal} in Lemma \ref{lemma_concentrate_v}, and the result holds {\it without} any boundedness condition on stochastic gradients. We prove this result by utilizing two special properties of  Adam: (1) stochastic gradients have a ``geometric sum'' structure in  $v_k$.  (2) the indices of stochastic gradients are uniformly sampled from a finite index set. With these two special properties, we find the behavior of $\frac{1}{\sqrt{v_k}}$ is largely predictable when $\beta_2$ is large.

\begin{lemma} {\bf (Concentration of Adam's preconditioner, simplified).} 
\label{lemma_concentrate_v}
Consider the same condition as Theorem \ref{thm_wr}. For any $0< \delta\leq \frac{1}{4n}$, there exists a sufficiently large $k$ such that  with probability at least 
$$
1-n \exp\left(-\frac{\delta^2}{(1-\beta_2)(\frac{28}{3n}+\frac{8}{3}\delta)}\right),
$$
we have the following concentration bound:
{\small
\begin{equation}
    \frac{C_{\mathrm{lower}}}{\sqrt{\mathbb{E}_k(v_{l,k})}} \le \frac{1}{\sqrt{v_{l,k}}} \le \frac{C_{\mathrm{upper}}}{\sqrt{\mathbb{E}_k(v_{l,k})}},
\end{equation}
}
where $v_{l,k}$ denotes the $l$-th component of $v_k$, $\Exk(\cdot)=\Ex(\cdot\mid\mathcal{F}_k)$ is the conditional expectation, and $C_{\text{lower}}, C_{\text{upper}}$ are defined as below, and they both approach 1 as $\beta_2 \rightarrow 1$.
$$
C_{\text{lower}} := 1- (1-\beta_2) \frac{4n}{(1-2n\delta)\beta_2^n}, \quad 
C_{\text{upper}} := \left(1-(1-\beta_2)\frac{8n}{(1-2n\delta)\beta_2^n}\right)^{-1/2}.
$$
\end{lemma}
\paragraph{Proof idea.} Technically, the concentration result is established via two steps:  (i) we map the dynamics of  $1/\sqrt{v_k}$ to the dynamics of a sequence of i.i.d. Bernoulli r.v.s, which is inherently bounded and is much easier to analyze; (ii) we map the dynamics back by several decoupling steps.   Step (i) and (ii) allow us to track the dynamics of possibly unbounded r.v. sequence using bounded Bernoulli proxies.  The concentration result is shown in Lemma \ref{lemma_concentrate_v}.  Lemma \ref{lemma_concentrate_v} can be used as a generic tool for analyzing Adam-type algorithms.

\section[Why Transformers Need Adam]{Why Transformers Need Adam: An Investigation into Hessian}
\label{sec_hessian}

\begin{Snugshade}[gray!10]
{\small This section is mainly based on:
\begin{center}
\begin{itemize}
    \item  \BLUE{\bf Why Transformers Need Adam: A Hessian Perspective.}  \\
    {\bf Yushun Zhang}, Congliang Chen, Tian Ding, Ziniu Li, Ruoyu Sun, Zhi-Quan Luo. 
   
     Advances in Neural Information Processing Systems (NeurIPS), 2024.
     \item \BLUE{\bf Towards Quantifying the Hessian Structure of Neural Networks.}

     Zhaorui Dong$^*$, {\bf Yushun Zhang$^*$}, Jianfeng Yao, Ruoyu Sun
     
    (*: Equal contribution. Alphabetically ordered.)

    IEEE Transactions on Information Theory, 2026.

     \item \BLUE{\bf Adam-mini: Use Fewer Learning Rates To Gain More.}
     
     {\bf Yushun Zhang$^*$}, Congliang Chen$^*$, Ziniu Li, Tian Ding, Chenwei Wu, Diederik P. Kingma, Yinyu Ye, Zhi-Quan Luo, Ruoyu Sun.

    (*: Equal contribution.)
    
     International Conference on Learning Representations (ICLR), 2025.
\end{itemize}
\end{center}
}
\end{Snugshade}

\begin{figure}[h]
\centering
    \subfigure[Adam on Gaussian quadratic objective]{\includegraphics[width=0.4\linewidth]{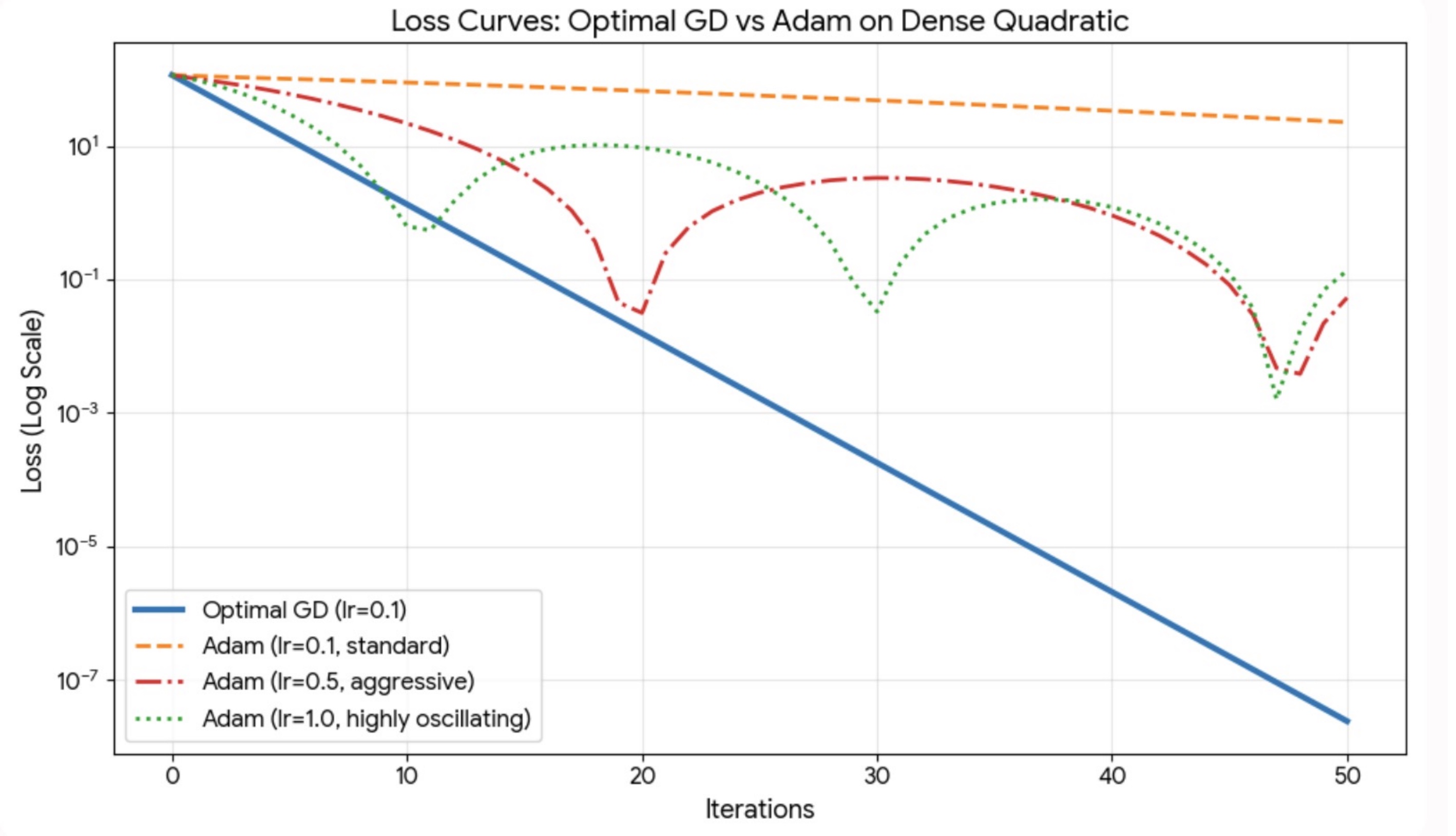}}
    \subfigure[Adam on Transformers]{\includegraphics[width=0.4\linewidth]{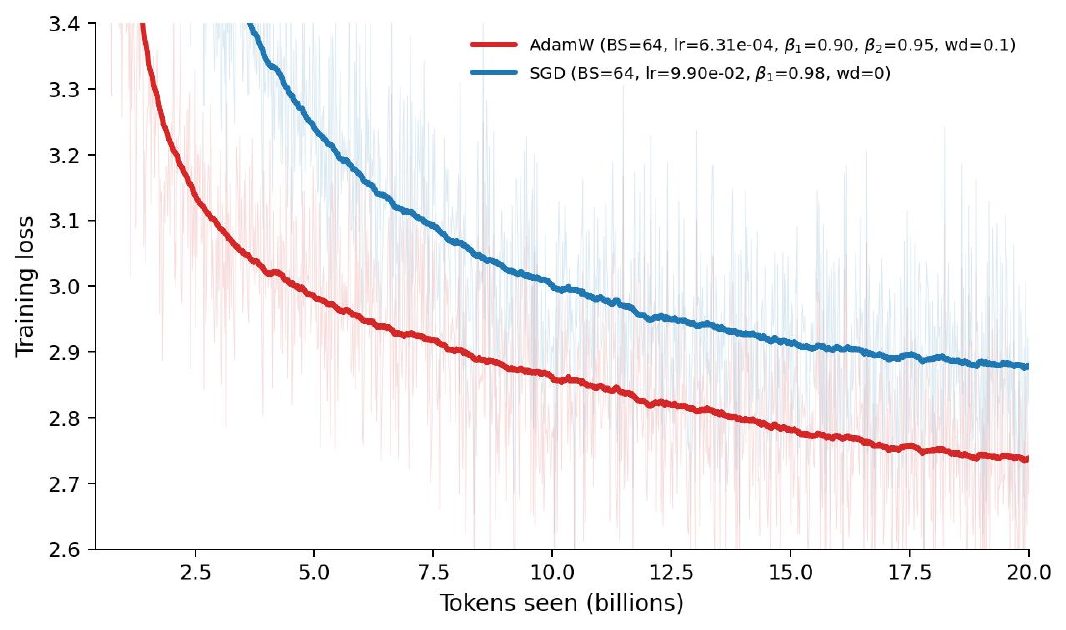}}
\caption{\small  Adam's practical advantage is not universal. {\bf (a)} Adam performs worse than GD on simple random Gaussian quadratic problems. {\bf (b)} Adam performs substantially better than SGD on a Transformer with 200M parameters. All methods were well-tuned through an extensive hyperparameter search.}
\label{fig_adam_performance}
\end{figure}

The convergence result explains that Adam can be safe under problem-dependent tuning, but it does not tell us when Adam is superior to SGD. Indeed, Adam's practical advantage is not universal: Adam performs substantially better than SGD on complex modern NNs like Transformers; performs on par with SGD on CNNs; and meanwhile, performs worse than GD on simple random Gaussian quadratic problems (see Figure \ref{fig_adam_performance}). These phenomena beg the question: what special structures in Transformers make Adam useful?

\paragraph{Adam as diagonal preconditioner.} In this section, we interpret the Adam optimizer in Algorithm \ref{algorithm_wr} as a diagonal preconditioning method:
{\small
\begin{equation}
    \theta_{k+1} = \theta_{k} - \eta_k \operatorname{Diag} (v_k)^{-\frac{1}{2}} m_k := \theta_{k} - \eta_k D_{\operatorname{Adam}}^{-\frac{1}{2}} m_k,
\end{equation}
}
where the preconditioner  $D_{\operatorname{Adam}} \in \mathbb{R}^{d\times d}$ is a diagonal matrix.
In numerical linear algebra, the efficacy of a diagonal preconditioner is known to depend on {\bf the structure of the Hessian}—whether it is diagonal, tridiagonal, diagonally dominant, or otherwise \citep{forsythe1955best,sun2021worst,qu2022optimal,das2024towards}.  Diagonal preconditioners typically perform poorly on dense Hessians,  because they cannot capture off‑diagonal couplings that dominate the curvature. The random Gaussian quadratic problem in Figure \ref{fig_adam_performance} (a) has a dense Hessian by construction, and we conjecture that this denseness is a primary reason for Adam’s poor performance in that regime.

\begin{figure}[h]
\centering
    \subfigure[$r$ vs. dimension $d$]{\includegraphics[width=0.4\textwidth]{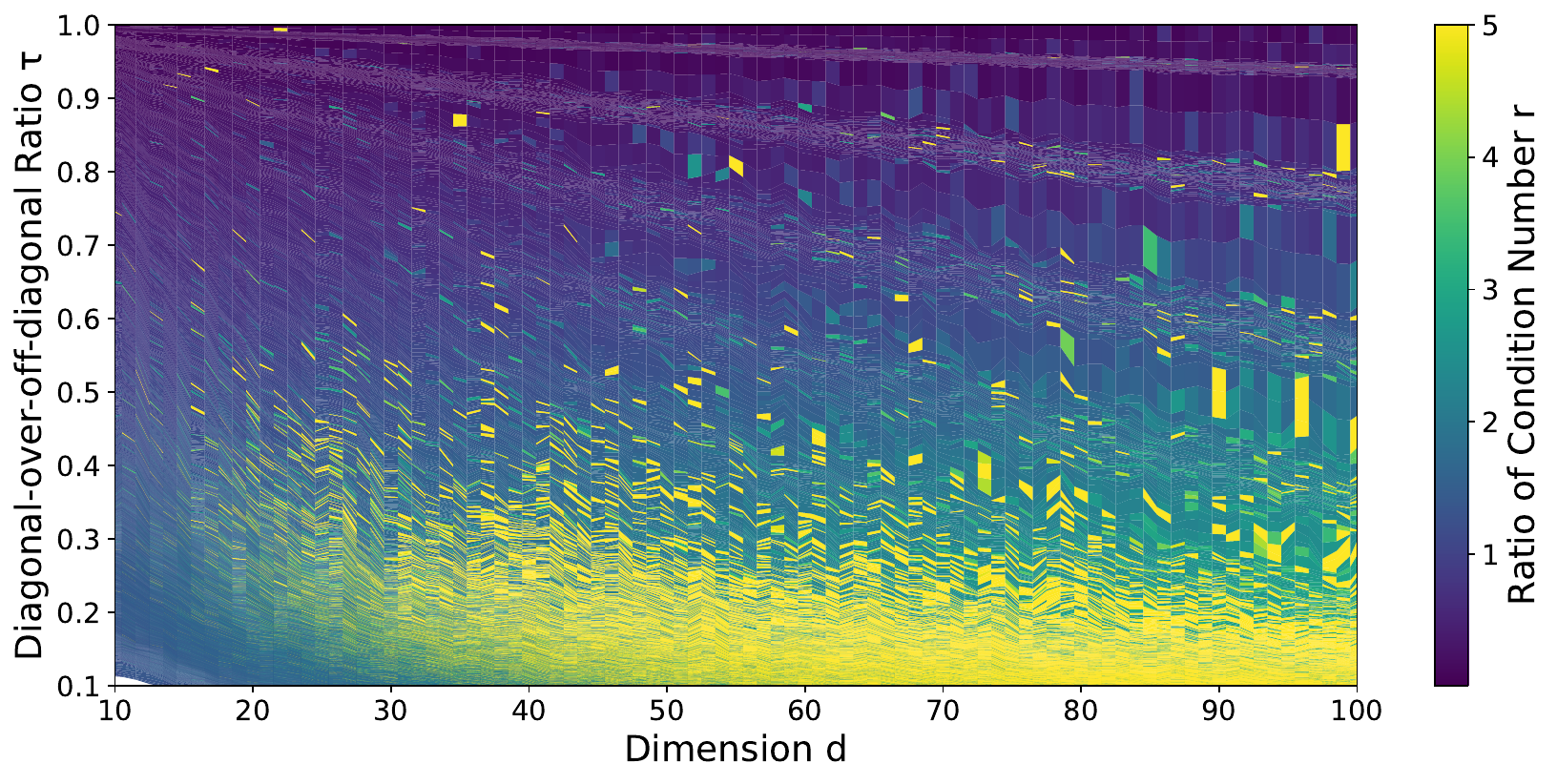}}
    \subfigure[$r$ vs. condition number $\kappa$]{\includegraphics[width=0.4\textwidth]{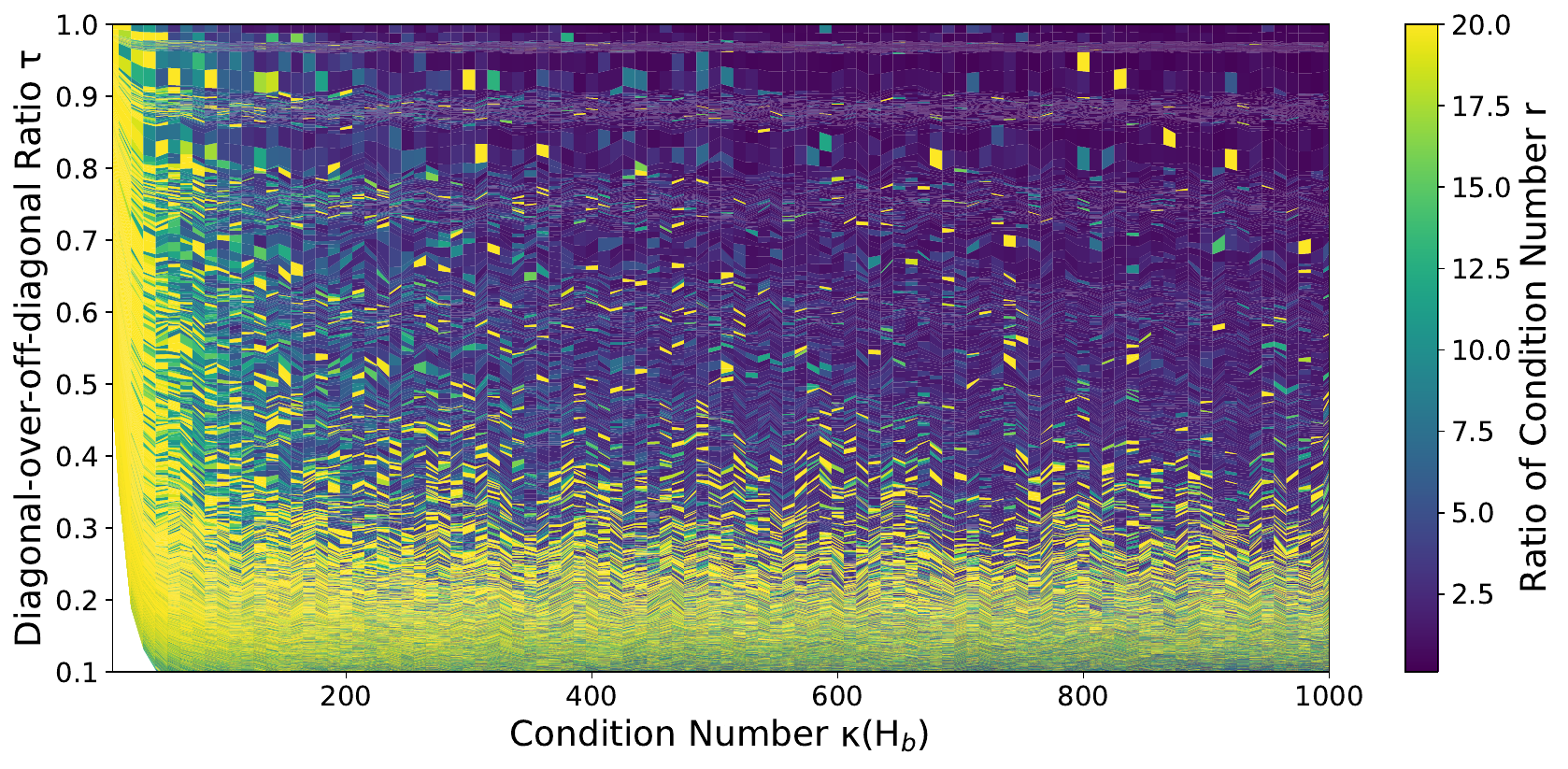}}
\caption{\small The effectiveness of Adam's preconditioner $D_{\text{Adam}}$ on different structures of a Hessian. The $y$-axis, denoted by $\tau \in [0,1]$, serves as a measure of Hessian diagonality; the smaller the value of $\tau$, the closer the Hessian is to a diagonal matrix. In the heat map, yellow corresponds to cases where  $D_{\text{Adam}}$  effectively reduces the Hessian's condition number, whereas blue denotes cases where it does not. Across different sizes and condition numbers of the Hessian, we find that $D_{\text{Adam}}$ is less effective when Hessian is dense.}
\label{fig:adam-preconditioner-block}
\end{figure}

To test this hypothesis, we numerically investigate the effectiveness of $D_{\operatorname{Adam}}$ on different Gaussian quadratic setups by varying the Hessian structure, its size, and its condition number. Our results in Figure \ref{fig:adam-preconditioner-block}  reveal a clear trend: as the Hessian becomes closer to diagonal, Adam’s performance consistently improves. This finding further supports the view that the interplay between the diagonal preconditioner and the underlying Hessian structure is a key determinant of Adam’s performance.

So what structure in Transformers makes Adam useful? In the following, we will show that Hessians of NNs and Transformers have special structure: they approach {\it near-block-diagonal} structure as training proceeds. Further, Transformer Hessians are not only block-structured but also strongly block-heterogeneous. We will verify both theoretically and empirically that these two facts together make Adam superior to SGD.

\paragraph{The Hessian of 1-hidden-layer NNs.}
We find that Hessians of NNs are far from generic dense matrices. Let us start with a 1-hidden-layer network:
{\small
\begin{equation}
\label{eq_1_hidden_layer_nn}
     f(W,V;x)=V\sigma(Wx)\in\bbR^C,
\end{equation}
}
where $W\in\mathbb{R}^{m\times d}$ is the hidden weight matrix, $V\in\mathbb{R}^{C\times m}$ is the output weight matrix, and $\sigma$ is ReLU. Here, $d, m, C$ denote the input dimension, the hidden dimension (i.e., number of neurons), and the output dimension (i.e., number of classes), respectively. The output is then fed into a Cross-Entropy (CE) loss, which we denote by $\ell(\theta)$, where \(\theta = (\operatorname{vec}(W^\top)^\top, \operatorname{vec}(V^\top)^\top)^\top \in \mathbb{R}^{md+mC}\), i.e., the row-wise flattening of the two matrices.

We now examine the Hessian $\nabla^2\ell(\theta)$ on this simple 1-hidden-layer network with $m = 8, d= C = 500$. As shown in Figure \ref{fig:closer_look_ce}, the Hessian has a composite structure:  
\begin{figure}[t]
 \vspace{-2cm}
    \centering
    \subfigure[Hessian at initialization]{\includegraphics[width=0.32\textwidth]{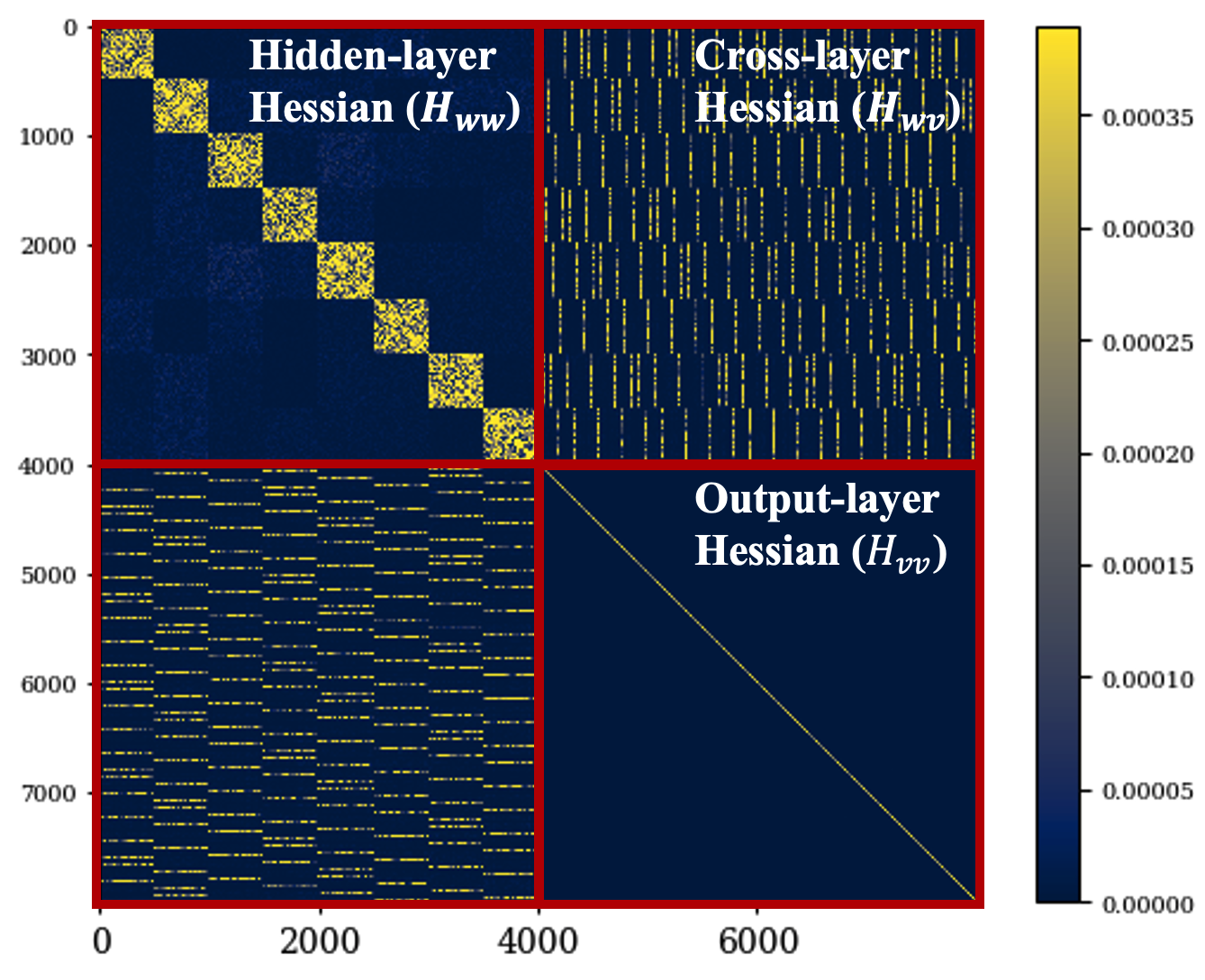}}
    \subfigure[Hessian at 50\% steps]{\includegraphics[width=0.32\textwidth]{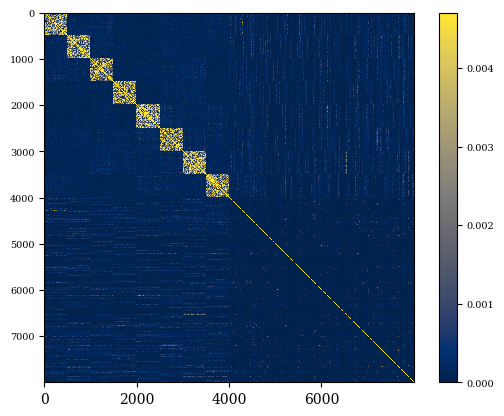}}
    \subfigure[Hessian at 100\% steps]{\includegraphics[width=0.32\textwidth]{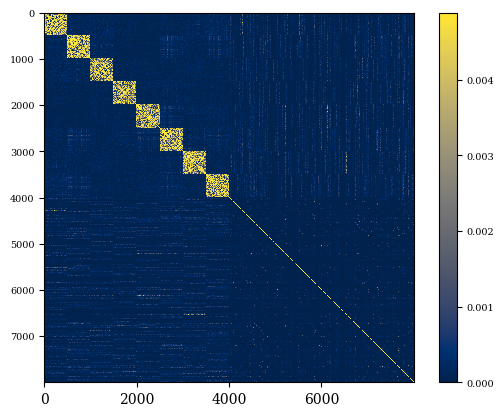}}
    \subfigure[Hessian at initialization]{\includegraphics[width=0.32\textwidth]{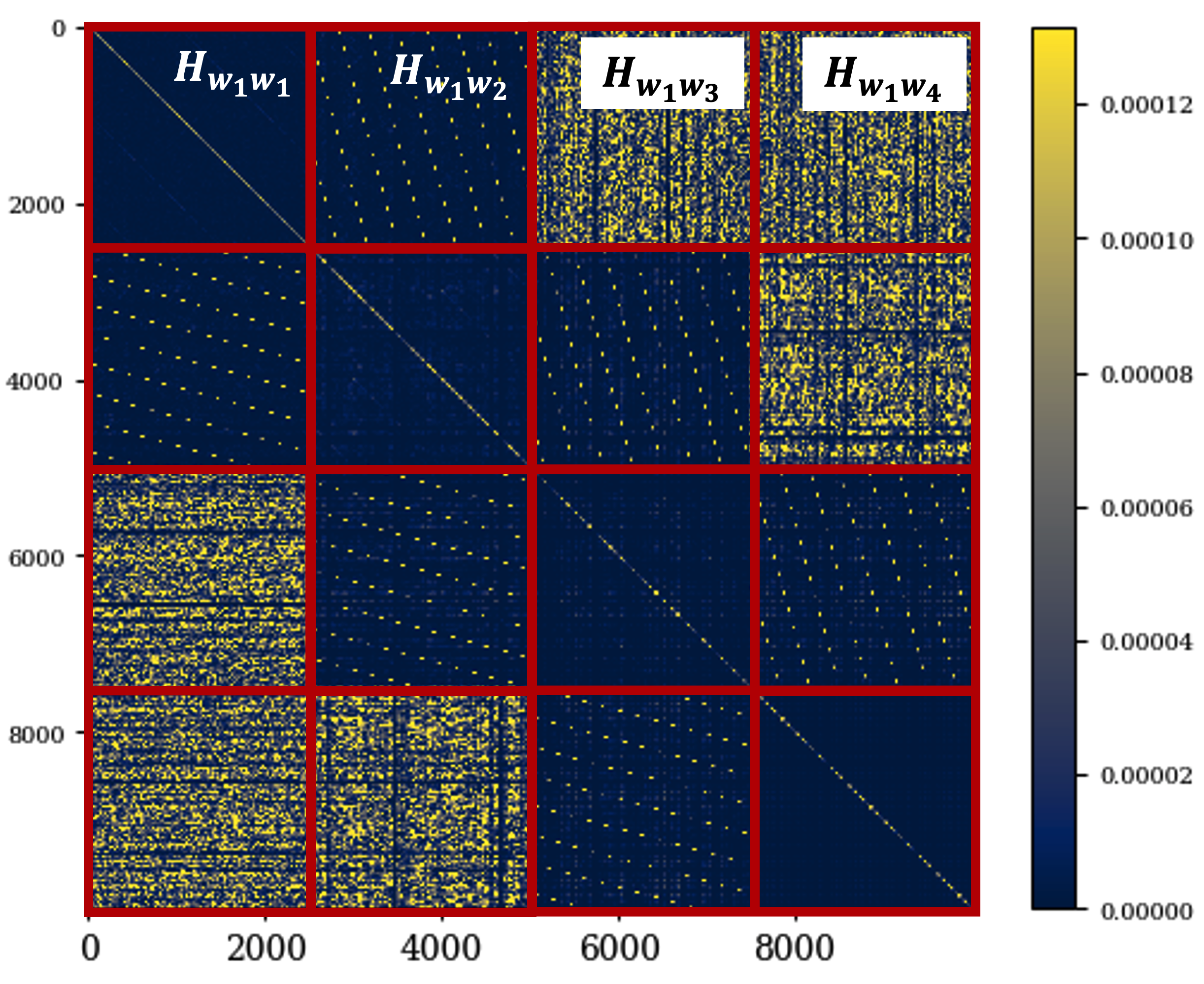}}
    \subfigure[Hessian at 50\% steps]{\includegraphics[width=0.32\textwidth]{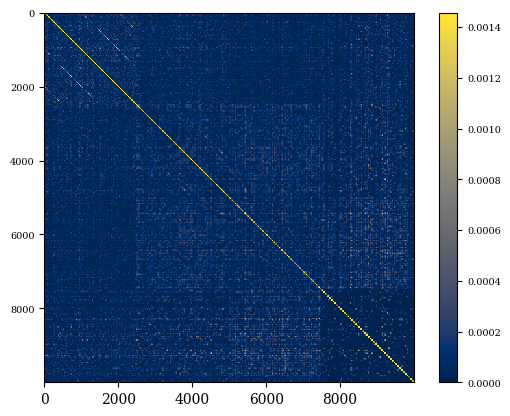}}
    \subfigure[Hessian at 100\% steps]{\includegraphics[width=0.32\textwidth]{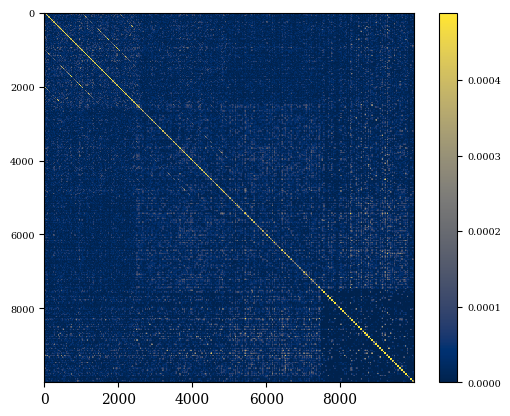}}
    \caption{\small  {\bf(a-c):}  The Hessian of a 1-hidden-layer network on Gaussian synthetic data under CE loss. At initialization, we observe the “block-circulant” pattern in $H_{wv}$, and the near-block-diagonal structure in $H_{ww}$ and $H_{vv}$ (with $m + C = 508$ blocks in total). We refer to it as {\it “block-circulant-block-diagonal matrix”}.  We notice that the “block-circulant” pattern in $H_{wv}$ vanishes along training, while the near-block-diagonal patterns in  $H_{ww}$ and $H_{vv}$ are preserved.  {\bf(d-f):}   The Hessian of a 4-layer network. We observe additional dense blocks in distant inter‑layer Hessian. Other structures are similar to those in 1-hidden-layer networks.  }
  \label{fig:closer_look_ce}
\end{figure}
\begin{itemize}
    \item {\bf Structure 1: block-diagonal.} The within-layer Hessian sub-matrices exhibit near-block-diagonal patterns, where one block corresponds to the parameters linked to one output neuron, i.e., one row in $W\in\mathbb{R}^{m\times d}$ or $V\in\mathbb{R}^{C\times m}$. The structure persists along training. In this case, $m= 8$ and $C = 500$, so we observe 8 and 500 blocks in the hidden and output layer, respectively. 
    \item {\bf Structure 2: block-circulant.} At the random initialization, the intra-layer Hessian sub-matrix exhibits a ``near-block-circulant'' pattern with periodic stripes. The structure vanishes along training.
\end{itemize}

\paragraph{The Hessian of deep NNs.} A similar structure also exists in deeper NNs, with a richer compound structure at initialization. We now extend the above \eqref{eq_1_hidden_layer_nn}  to a 4-layer network:
{\small
\begin{equation}
\label{eq_deep_nn}
        f(W,V;x)=W_4\sigma(W_3 \sigma (W_2  \sigma (W_1 x)) )\in\bbR^C,
\end{equation}
}
where $W_1\in\bbR^{m\times d}, W_2, W_3 \in \bbR^{m\times m}, W_4 \in \bbR^{C\times m}$ are the weight matrices. When $d = m = C = 500$, we observe the following phenomena in Figure \ref{fig:closer_look_ce}:

\begin{itemize}
    \item {\bf Structure 1: block-diagonal.} The within-layer Hessian sub-matrices exhibit near-block-diagonal patterns, where one block corresponds to the parameters associated with one output neuron, i.e., one row in $W_i, i = 1,\cdots, 4$. The structure persists along training. In this case, $m= C = 500$, so we observe 500 blocks in each layer.
    \item {\bf Structure 2: block-circulant.} At the random initialization, the adjacent intra-layer Hessian sub-matrix exhibits a ``near-block-circulant'' pattern with periodic stripes. The structure vanishes along training.
    \item {\bf Structure 3: dense.} At the random initialization, the distant inter‑layer Hessian sub-matrix exhibits a dense pattern. The structure vanishes along training.
\end{itemize}
We summarize the key observations as follows.
\begin{snugshade}
   {\bf  Observation 1.} The Hessians of shallow and deep NNs appear to be ``{\it block-circulant-block-diagonal}'' and ``{\it dense-block-circulant-block-diagonal}'' matrices at the initialization. 
   
   {\bf  Observation 2.} Training tends to simplify these matrices towards  ``near-block-diagonal'' ones, where one block corresponds to one row in the weight matrix.
\end{snugshade}

\begin{figure}[h]
    \centering
    \includegraphics[width=0.85\textwidth]{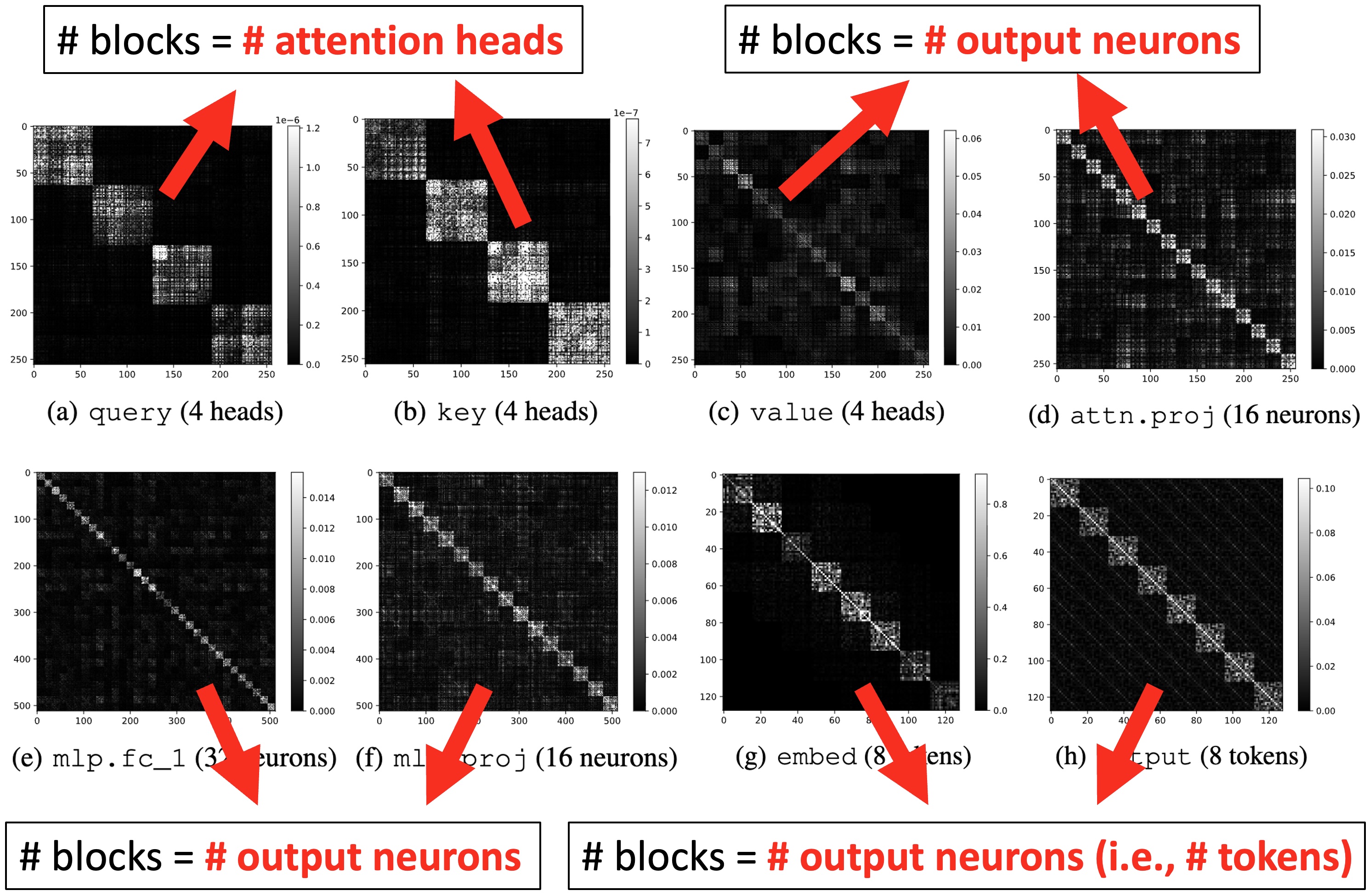}
  \label{fig:transformer-hessian-placeholder}
      \caption{\small  {\bf(a-f):} The Hessian of different modules in a Transformer.}
    \label{fig_transformer_hessian}
\end{figure}

\paragraph{The Hessian of Transformers.} In Figure \ref{fig_transformer_hessian}, we observe a similar phenomenon in Transformers: for most modules, including MLPs, each Hessian block corresponds to a row of the weight matrix; for Query and Key modules, each block corresponds to a head. This head‑wise structure is a direct consequence of the inherent parallel computation across heads.

\paragraph{Intuition on Linear Nets under MSE loss.} Here, we provide simple intuition on why the block-structure appears in the Hessian of NNs, and why one block corresponds to one row in the weight matrix. We reveal that the source stems from {\it consecutive multiplication of large matrix variables.}

We start with one layer in a linear NN without ReLU activation $y = Wx = [w_1^\top x; \cdots; w_m^\top x]$. One linear algebra fact is that: the change of $i$-th neuron $y_i$ only depends on the $i$-th row of W, i.e., $w_i^\top$, but not other rows. This forms the foundation of the block-diagonal pattern: one block corresponds to one row of $W$.

\begin{figure}[t]
\vspace{-1.5cm}
    \centering
    \includegraphics[width=0.85\textwidth]{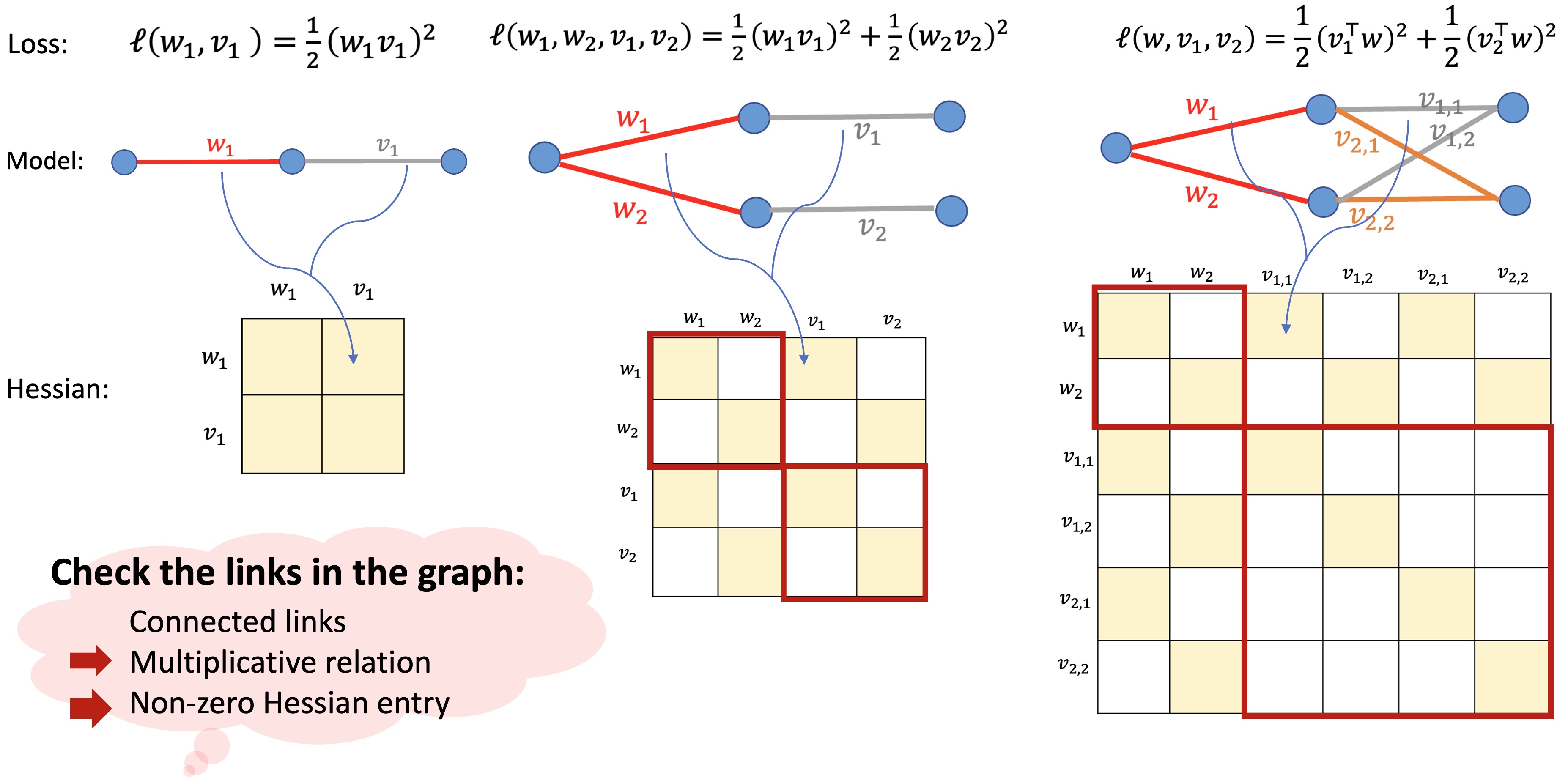}
      \caption{\small The Hessian structure of  NNs can be understood through their computation graphs: any pair of connected links yields a multiplicative relation in the loss, which further gives a non-zero Hessian entry; while the non-connected pair gives a vanishing or exactly zero Hessian entry. }
    \label{fig_computation_graph}
\end{figure}

The above linear algebra fact also helps explain the  ``block-circulant-block-diagonal'' pattern in the 1-hidden-layer linear NN case. 
 By the definition of matrix product, some links are connected, while some are not. As shown in Figure \ref{fig_computation_graph}, any pair of connected links yields a multiplicative relation in the loss, which further gives a non-zero Hessian entry; while the non-connected pair gives a vanishing or exactly zero Hessian entry.

We can further use this observation to predict the Hessian structure of deep linear networks. The question then becomes ``whether there exists a connected path between any two distant links in the NN computation graph.'' For deep linear networks, the answer is essentially always yes. This explains the dense pattern observed in {\bf Structure 3.}

We note that the above discussion is useful for intuition, but it is not a rigorous argument. Further, it does not fully explain why the intra-layer components vanish during training. A rigorous theory requires deriving the explicit Hessian expression and analyzing the asymptotic spectrum of each Hessian block as the network size grows.

The main technical challenges in the proof come from the ReLU nonlinearity and the non-separable Cross-Entropy loss. In this case, we empirically find that the matrix sizes need to be large to cancel out the off-diagonal signals (see Figure~\ref{fig:hessian_nn_mse_different_c} below). We summarize the key insights as follows.

\begin{snugshade}
\begin{center}
   {\bf Key insight.} The special Hessian structure of NNs stems from consecutive multiplication of \\ large matrix variables.
\end{center}
\end{snugshade}

We remark that the near-block-diagonal Hessian was first empirically reported in \citep{collobert2004large} on a simple NN, and the author attributed it to CE loss. Our findings challenge this conventional wisdom and suggest that CE loss is not crucial. On contrary, matrix product is the major source.

\subsection{Rigorous Theory}

We now provide rigorous theory on the Hessian structure of 1-hidden-layer NN in \eqref{eq_1_hidden_layer_nn}. The major technical challenge comes from ``non-linearity'', i.e., CE loss and ReLU activation. We introduce new methods in random matrix theory to handle them.  We reveal that the reported Hessian structure comes from a mixture of two forces: a ``static force'' rooted in the definition of matrix product, and a ``dynamic force'' arisen from training.

\begin{figure}[t]
\vspace{-1cm}
    \centering
    \subfigure[\small $C = 10$]{\includegraphics[width=0.30\textwidth]{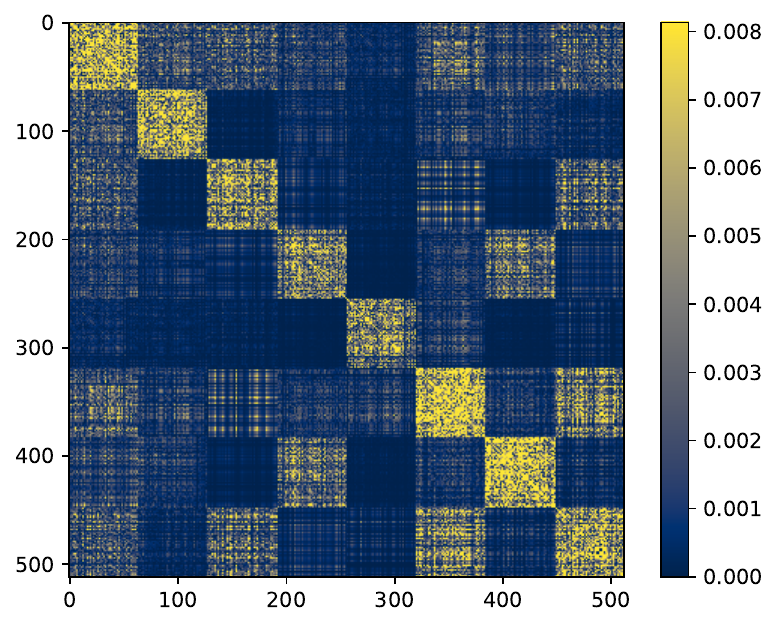}}
    \subfigure[\small $C = 100$]{\includegraphics[width=0.30\textwidth]{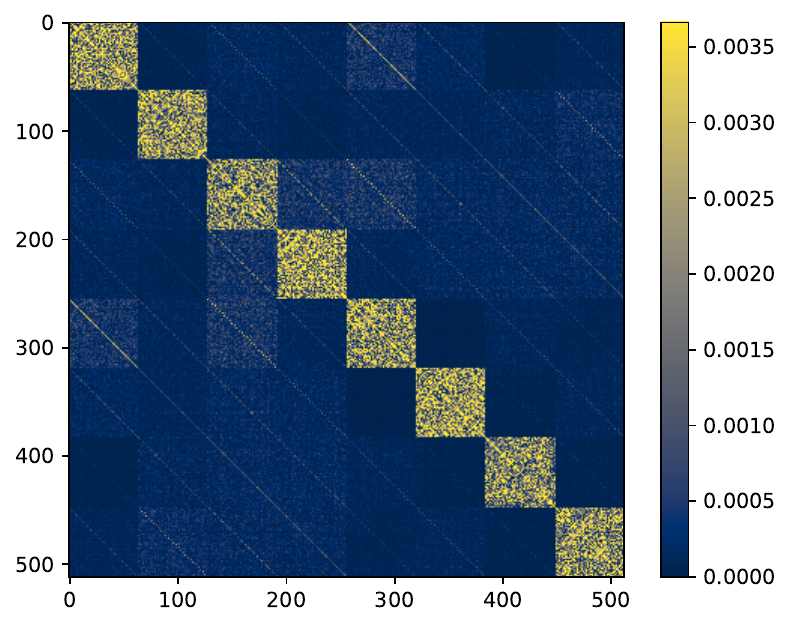}}
    \subfigure[\small $C = 1000$]{\includegraphics[width=0.30\textwidth]{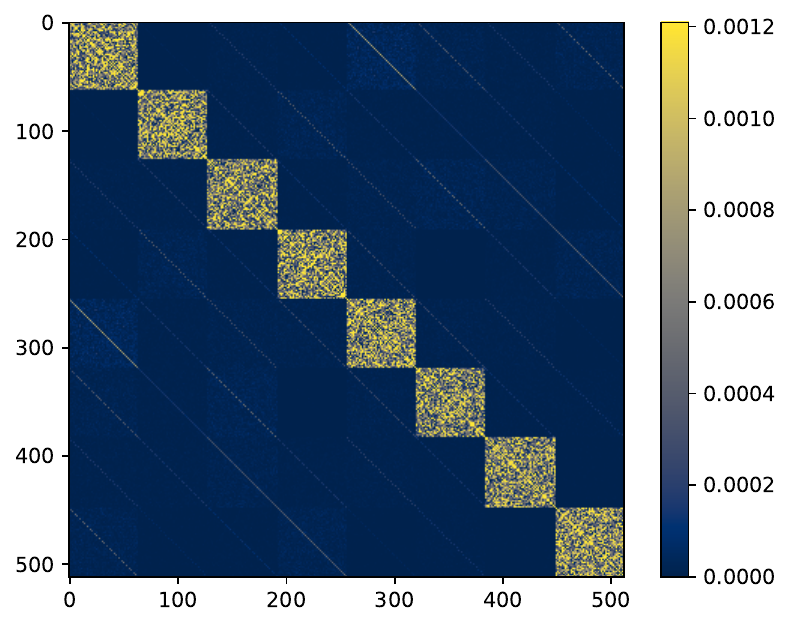}}
    \caption{\small   The hidden-layer Hessian $H_{ww}$ of a 1-hidden-layer network. The network has 8 hidden neurons.  We observe that the block-diagonal Hessian structure in $H_{ww}$ becomes clearer as $C$ increases. }
  \label{fig:hessian_nn_mse_different_c}
\end{figure}

\paragraph{``Static force'' shapes the within-layer structure.} As suggested above, the static force originates from matrix multiplication, and the matrix size needs to be large in non-linear NNs. We now state the rigorous statement at random initialization. We find that the degree of block-diagonality grows with output dimension $C$. This is also observed numerically in Figure~\ref{fig:hessian_nn_mse_different_c}, where the background noise in Hessian vanishes as $C$ grows. This suggests that a large $C$ is necessary for the structure.

\begin{theorem}[One-hidden-layer networks, simplified]
\label{thm_1_hidden_layer_nn}
Consider the Hessian of \eqref{eq_1_hidden_layer_nn}. Assume each entry of the input data matrix ${X}=({x}_1,\cdots, {x}_N)\in\R^{d\times N}$  follows i.i.d. $\mathcal{N}(0,1)$. Assume the model weights in $W$ and $V$ are initialized by LeCun initialization.  Then for any fixed $m\geq 3$, suppose $d,N\rightarrow \infty, \frac
    {d}{N}\rightarrow \gamma\in (0,+\infty)$, it holds that
{\small \begin{equation}
\label{eq_thm_nn_ratio}
    \lim_{d,N\rw\infty} \frac{\Ex\l[\bigg\|\frac{\partial^2\ell(\theta)}{\partial w_i  \partial w_j^\top}\bigg\|_{\operatorname{F}}^2\r]} {\Ex\l[\bigg\|\frac{\partial^2\ell(\theta)}{\partial w_i  \partial w_i^\top}\bigg\|_{\operatorname{F}}^2\r]} = \mathcal{O}(1/C),\quad
    \lim_{d,N\rw\infty} \frac{\Ex\l[\bigg\|\frac{\partial^2\ell(\theta)}{\partial v_i  \partial v_j^\top}\bigg\|_{\operatorname{F}}^2\r]} {\Ex\l[\bigg\|\frac{\partial^2\ell(\theta)}{\partial v_i  \partial v_i^\top}\bigg\|_{\operatorname{F}}^2\r]} = \mathcal{O}(1/C^2),
\end{equation}}
where $C$ denotes the output dimension, i.e., the number of classes.
\end{theorem}
\paragraph{``Dynamic force'' erases the intra-layer components.}  By direct calculation on the Hessian expression, one can see that the magnitude of the intra-layer ``block-circulant''  component is proportional to the optimality gap. That is:
{\small
\begin{equation}
    \label{eq_cross_layer}
    \frac{\partial^2\ell(\theta)}{\partial w_i  \partial v_j^\top} \approx     \left[\begin{array}{cccccc}
0 & \cdots & a_{1, i} & 0 & \cdots & 0 \\
\vdots & \ddots & \vdots & \vdots & \ddots & \vdots \\
0 & \cdots & a_{d, i} & 0 & \cdots & 0
\end{array}\right] \in \mathbb{R}^{d \times m}, \text{ and } a_{d^\prime, i} = \mathcal{O} \l(\ell(\theta) - \ell^* \r),d^\prime \in [d].
\end{equation}
}
Therefore, as training progresses, the intra-layer component is expected to shrink. The proof is straightforward and we omit the detailed calculation here.

\yushunrevise{The above calculation suggests that the Hessian of the cross-layer components is proportional to the optimality gap (i.e.,  $\mathcal{O} \l(\ell(\theta) - \ell^* \r)$), which requires proper training to vanish. Once it vanishes, it in turn accelerates the training process, thereby forming a positive feedback loop. Empirically, recent work \citep{abreu2025potential} compares the full Gauss–Newton (GN) method with the layer-wise GN approximated version for LLM pretraining, and they find that the layer-wise version incurs negligible degradation on final performance. This suggests that, when paired with an appropriate optimizer, the influence of the cross-layer Hessian can be neglected.}

\paragraph{Proof idea for Theorem \ref{thm_1_hidden_layer_nn}.}
We highlight some key technical contributions in our proof. The major challenge lies in characterizing limiting spectrum of product of {\it dependent} random matrices $X \Lambda X^\top$, where each entry in $X \in \mathbb{R}^{d \times N}$ follows i.i.d. Gaussian and $\Lambda \in \mathbb{R}^{N \times N}$ is a diagonal matrix that is {\it dependent} on $X$.  

Dependent matrix product is a non-standard problem in random matrix theory. For our matrix of interest, we find an extra structure: the dependency between $X$ and $\Lambda$ purely comes from ReLU activation and CE loss, and it diminishes as $d \rightarrow \infty$, a phenomenon we refer to as ``asymptotic independence''. With this observation, we manage to find the limiting spectrum of  $X \Lambda X^\top$ via two steps: first, we propose a decoupling strategy based on {\it Lindeberg interpolation principle} \citep{Lindeberg1922, pastur2020random} to address such dependency; second, after decoupling, we apply the generalized Marcenko-Pastur (GMP) law \citep{Pastur67} to characterize the spectrum.   

 \yushunrevise{The  Lindeberg principle is originally an elegant proof for the Central Limit Theorem (CLT) \citep{Lindeberg1922}, by replacing the random variables with Gaussian ones incrementally and proving the impact is negligible under certain conditions. The Lindeberg principle is also applicable for random matrices \citep{Chatterjee2006,gotze2015asymptotic,pastur2020random}. We find that such methods are useful for handling asymptotic independence in our case. }

We now illustrate our proof strategy. We consider the Hessian block of $i$-th row in output weight $V$ (denoted as ${H}_{ii}$) as an example. 
For ${H}_{ii}$, it has the  following expression under CE loss:
{\small
\be
\label{eq_hvv_ii}
{H}_{ii} = \frac{1}{N}\sum_{n= 1}^N  {p}_{n,i} (1-{p}_{n,i}) x_n x_n^\top,\quad {p}_{n,i}:=\frac{\exp({v_i^\top {x}_n})}{\sum_{c=1}^C\exp({v_c^\top {x}_n})},
\ee
}
The goal is to find the  limit of $\|{H}_{ii}\|_\text{F}^2$ as $N$ and $d$ grow proportionally.
\begin{itemize}
    \item {\bf Step 1.} We introduce the following decoupled matrix 
{\small
\be
\wt{H}_{ii}= \frac{1}{N}\sum_{n= 1}^N  \wt{p}_{n,i} (1-\wt{p}_{n,i}) x_n x_n^\top,\quad \wt{p}_{n,i}:=\frac{\exp({v_i^\top \wt{x}_n})}{\sum_{c=1}^C\exp({v_c^\top \wt{x}_n})},
\ee
}
where $\wt{X} = (\wt{x}_1,\cdots,\wt{x}_N)\in\R^{d\times N}$ is an independent copy of $X$. The goal is to prove that 
$$
\lim_{N\rw\infty} \l(s_{{H}_{ii}}(z)-s_{\wt{H}_{ii}}(z)\r)=0,\quad \text{a.s.}\quad \forall z\in \mb{C}^+,
$$
where $s_{{H}_{ii}}(z)$ denotes the Stieltjes transform \citep{tao2012topics} for the eigenvalue distribution of  $H_{ii}$. 
From standard measure concentration results, $s_{{H}_{ii}}(z),s_{\wt{H}_{ii}}(z)$ concentrate around their means as $N\rw\infty$. Therefore, it suffices to prove that
$\lim_{N\rw\infty} \l(\E[s_{{H}_{ii}}(z)]-\E[s_{\wt{H}_{ii}}(z)]\r)=0$.
\item {\bf Step 2.} Following  the Lindeberg principle, we 
define the matrix interpolation process
$$
X(t)=\sqrt{t} X + \sqrt{1-t}\wt{X},\quad t\in [0,1].
$$
Note that $X(1)=X$ and $X(0)=\wt{X}$. We then define
$$
{H}_{ii}(t) = \frac{1}{N}\sum_{n= 1}^N  {p}_{n,i}(t) (1-{p}_{n,i}(t)) x_n x_n^\top,\quad \text{ where }   {p}_{n,i}(t):=\frac{\exp({v_i^\top {x}_n(t)})}{\sum_{c=1}^C\exp({v_c^\top {x}_n(t)})}.
$$
Then we have
{\small
\be \label{eq_integrand}
\mathbb{E}\left[ s_{H_{ii}}(z) - s_{\widetilde{H}_{ii}}(z) \right]
= \int_0^1 \mathbb{E}\left[ \frac{d}{dt} s_{H_{ii}}(t) \right] dt.
\ee
}
\item {\bf Step 3.} Then we calculate the integrand in \eqref{eq_integrand} using Stein's  Lemma:   for  $Z\sim \ml{N}(0,1)$ and differentiable function $f:\mb{R}\rw\mb{C}$ with sub-exponential decay at infinity, we have: 
{\small
\bea\label{SteinLemma_proof_sketch}
    \E[Zf(Z)]= \E[f^\prime(Z)].
    \eea
}
    We then prove that the r.h.s. of  \eqref{SteinLemma_proof_sketch} decays to zero at rate  $\mathcal{O}(1/\sqrt{N})$, where $N$ is the sample size, and thus ${H}_{ii}$ shares the same limiting eigenvalue distribution as $\wt{H}_{ii}$.  This step uses the assumption of Gaussian data and initialization in Theorem \ref{thm_1_hidden_layer_nn}. 
    
    \yushunrevise{We note that Step 3 would fail without the ``asymptotic independence'' arising from the ReLU activation and CE loss. Therefore, we do not address the general dependent-random-matrix problem; instead, we resolve only a specific case of it.}
\item {\bf Step 4.} Apply GMP law \citep{Pastur67} to the decoupled matrix $\wt{H}_{ii} = X\wt{\Lambda}X^\top$ (where $\wt{\Lambda}$ is independent of $X$), and we get the limit eigenvalue distribution of $\wt{H}_{ii}$. Then we apply the Taylor expansion in the functional equation in the GMP law to get the limiting second moment of $\mu_{\wt{H}_{ii}}$, which is also the limit of $\|{H}_{ii}\|_\text{F}^2$. This concludes  the proof.
\end{itemize}

\subsection{Why Transformers Need Adam: Block-Heterogeneity}
Based on the findings above, we briefly discuss why Adam is effective on Transformers. We find it is a result of the joint effect of {\it structure} and {\it eigenvalues} in the Hessian.

\begin{figure}[h]
    \centering
     \subfigure[ResNet18]{\includegraphics[width=0.27\textwidth]{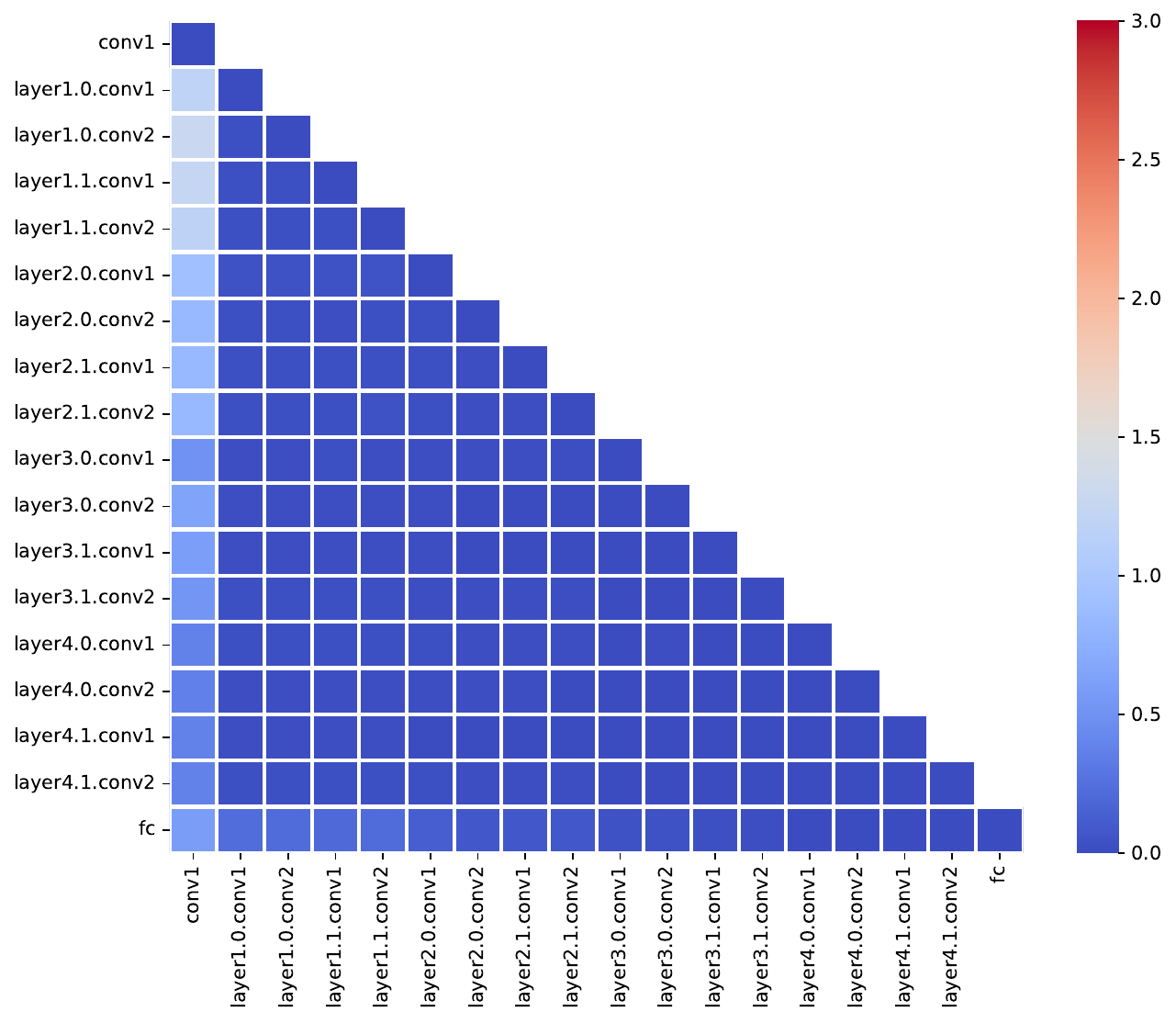}}
    \subfigure[BERT]{\includegraphics[width=0.27\textwidth]{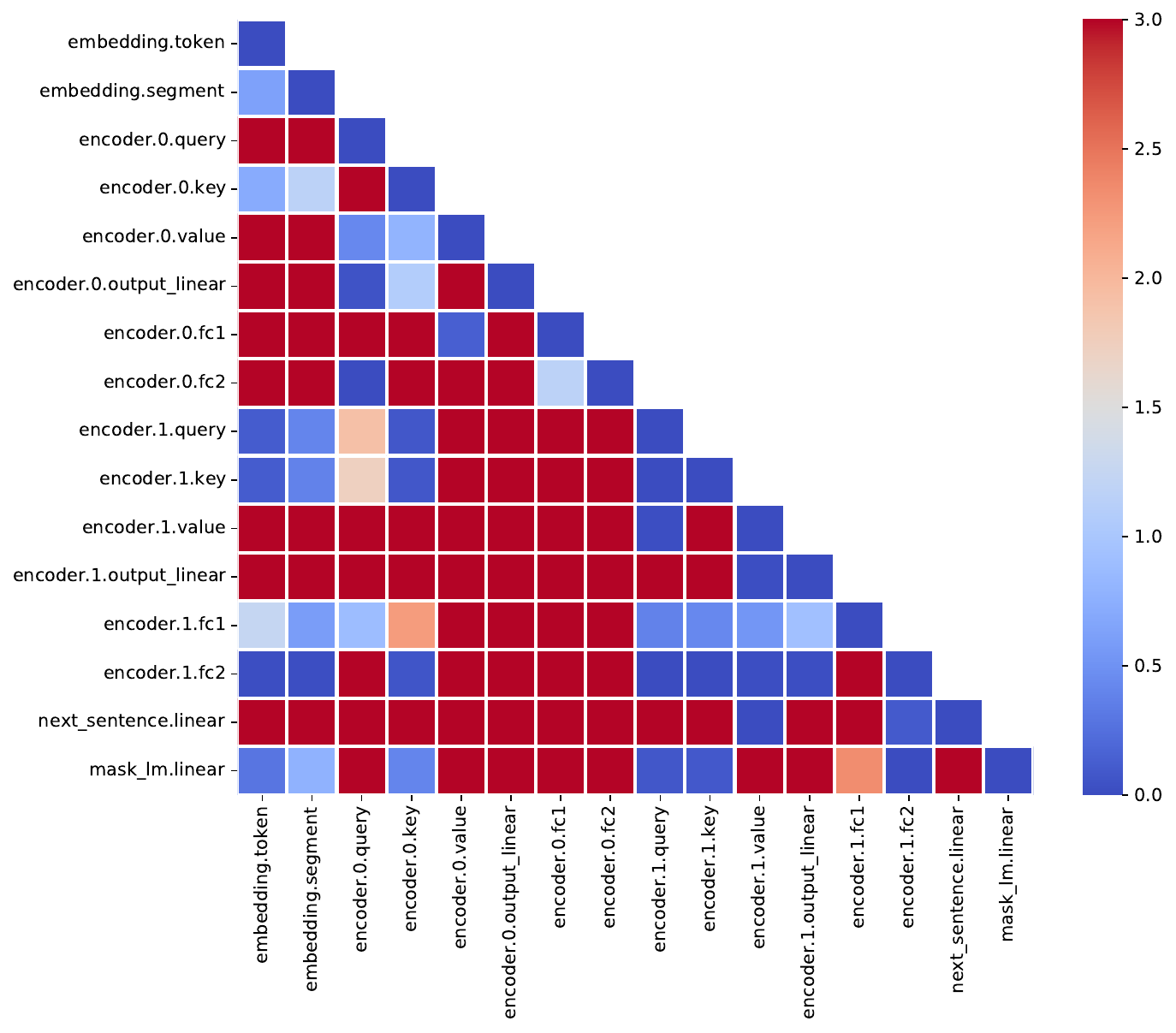}}
    \subfigure[GPT2-nano]{\includegraphics[width=0.27\textwidth]{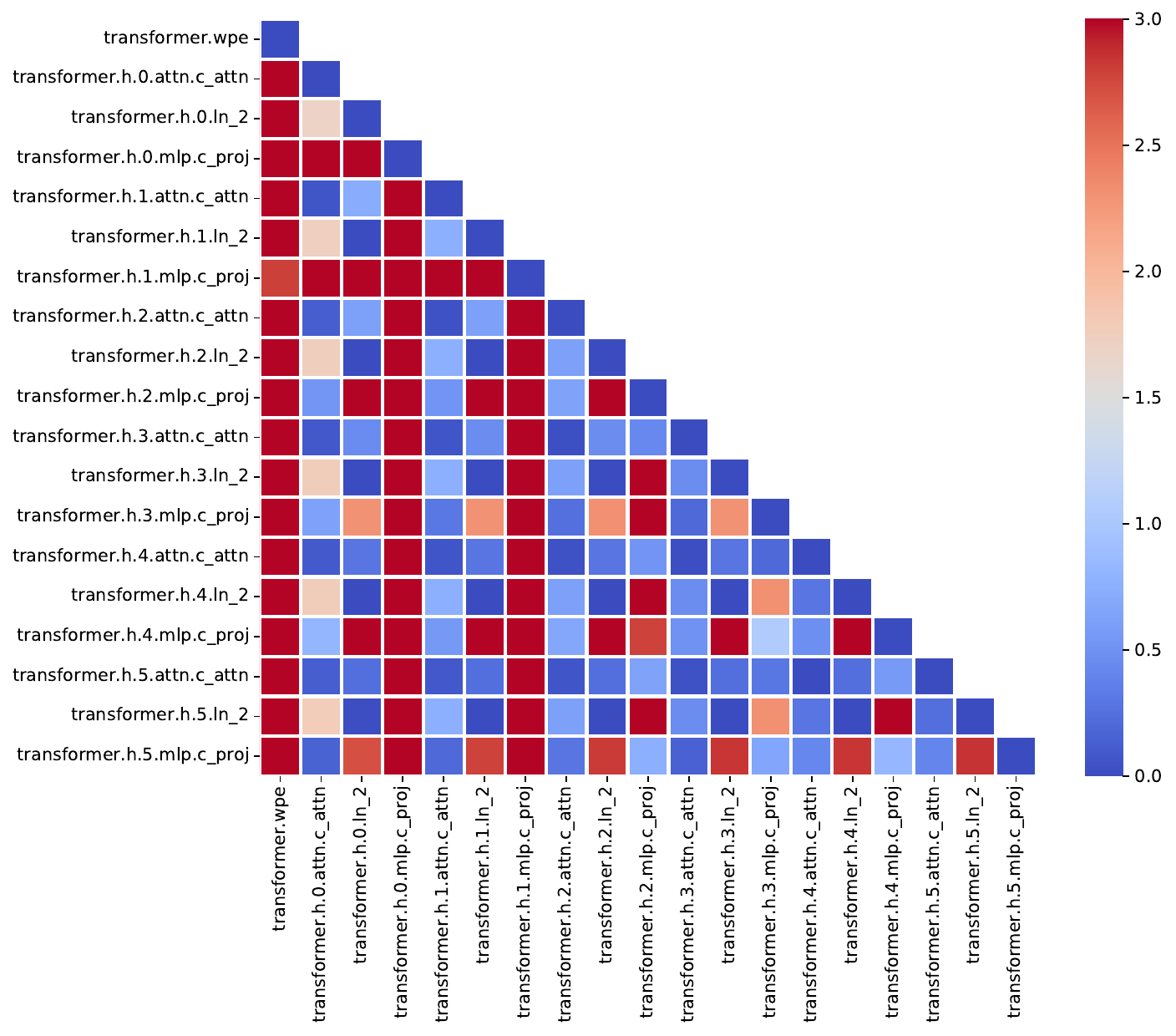}}
    \caption{\small  {\bf(a-c):} The Jensen-Shannon (JS) distance among blockwise Hessian spectra at initialization. We find that the JS distance of blockwise spectra in CNNs is significantly smaller than that in Transformers. }
    \label{fig_blockwise_heatmap}
\vspace{-0.3cm}
\end{figure}

\begin{itemize}
    \item First, as shown in Figure \ref{fig_transformer_hessian}, Transformers share a similar Hessian structure with MLPs.  This is because: (i) Transformers also consist of consecutive multiplication of large matrix variables; (ii) the head-wise parallel operations in attention insert an additional source of block-diagonality.
    \item Second, in Figure \ref{fig_blockwise_heatmap}, we report that different Hessian blocks in Transformers have very different eigenvalue distributions (i.e., spectra), a phenomenon we refer to as {\bf block-heterogeneity}. \yushunrevise{We will discuss the source of block-heterogeneity} later.
\end{itemize}

Through extensive experiments on Transformers, CNNs, MLPs, and quadratic problems, we find that SGD can perform on par with Adam on problems without block heterogeneity (e.g., on CNNs), but performs worse than Adam when the heterogeneity exists.   Intuitively, SGD is worse because it applies one single learning rate to all blocks, which cannot handle the heterogeneity among blocks. This limitation could be ameliorated if we use coordinate-wise learning rates, as designed in $D_{\operatorname{Adam}}$.  

\paragraph{About Adam on CNNs.} Figure~\ref{fig_blockwise_heatmap} shows that the block‑wise spectra in CNNs are relatively homogeneous, which may explain why Adam $\approx$ SGD on CNNs. To be rigorous, however, we have not verified whether CNNs exhibit a similar near‑block‑diagonal Hessian structure. In fact, for a given parameter budget, the matrix dimensions in CNNs are typically much smaller than those in Transformers, so observing a similar Hessian structure may require training CNNs at a much larger scale. The verification would be computationally prohibitive, and we have not yet conducted it. To this end, why Adam lacks advantage on CNNs remains largely open. Two plausible explanations are the homogeneous spectra (assuming the block structure does appear in large-scale CNNs) or a dense Hessian (if it does not)—both of which can undermine the coordinate‑wise learning rates in $D_{\operatorname{Adam}}$.

\paragraph{Initial theory on Adam vs. GD.} On quadratic objectives, we prove that Adam's diagonal preconditioner $D_{\operatorname{Adam}}$ is provably effective under block-heterogeneity. We summarize our key results in Theorem \ref{thm_gd_lower_bd}.   Adam has the rate $\tilde{O}\left(\max_{l\in [L]} \kappa_l  \right)$, which can be faster than the rate of GD $\Omega(\kappa)$, where $\kappa$ and $\kappa_{l}$ denote the condition number of full Hessian and the $l$-th block in Hessian, respectively.
\begin{theorem}[Adam vs. GD on quadratic objective, simplified]
\label{thm_gd_lower_bd}  Consider  $\min _\theta \ell(\theta) =\frac{1}{2} \theta^\top H \theta- h^\top \theta$ where $H \in \mathbb{R}^{d \times d}$ is positive definite and block-diagonal, and $h \in \mathbb{R}^{d} $. Let $\theta_{\operatorname{Adam}}^k$ and $\theta_{\operatorname{GD}}^k$  be the output of GD  and Adam after $k$ steps, respectively. For Adam, consider random initialization under any continuous distribution, $\beta_1 = 0$, $\beta_2 = 1$, $v_0$ equals the initial gradient, and constant step size $\eta_k = \eta$, we have the following result w.p.1.:
{\small
\begin{equation}
\label{eq_adam}
       \ell(\theta_{\operatorname{Adam}}^{k+1})-\ell^* \leq  \max_{l\in [L]} \left(1  - \frac{1}{  r   \kappa_{l} } \right)
\left(\ell(\theta_{\operatorname{Adam}}^k)  - \ell^*\right),
\end{equation}
}
where $r$ is an initialization-dependent constant and $\kappa_l$ is the condition number of $l$-th block in $H$.
For GD, there exists a block diagonal matrix $H$, $h$ and an initial point  $\theta^0$, s.t., for any $\eta$, we have: 
   {\small
\begin{equation}
\label{eq_gd}
       \ell(\theta_{\operatorname{GD}}^{k+1})-\ell^* \geq  \left(1  - \frac{2}{ \kappa + 1} \right)^2
\left(\ell(\theta_{\operatorname{GD}}^k)  - \ell^*\right),
\end{equation}
}
where $\kappa$ is the condition number of $H$.
\end{theorem}

In the proof of Theorem \ref{thm_gd_lower_bd}, we try to bound the error-yet-to-be-optimized (a.k.a., cost-to-go) and the per-step improvement for each block. The ratio of these two terms characterizes the rate of convergence of each block, which is determined jointly by the preconditioner and Hessian eigenvalues. Our core finding is that: the arrangement of the intermediate eigenvalues across blocks affects Adam's performance, but leaves GD's unchanged.  Detailed analysis can be seen in Appendix E in \citep{zhang2024transformers}.

We now compare the complexity of Adam and that of  GD.
By Theorem \ref{thm_gd_lower_bd}, Adam is faster than GD when $r \cdot\max_{l \in [L]}  \kappa_l \leq \kappa $. In the quadratic model with heterogeneous blocks, 
 our simulation over 1000 random trials shows that $r \leq 1000$ with probability $\geq \frac{2}{3}$ when using standard Gaussian random initialization. Since $\max_{l \in [L]}  \kappa_l \approx 1$, we have $r \cdot \max_{l \in [L]} \kappa_l \leq 1000$, w.h.p., and is about $5 \times$ smaller than $\kappa = 5000$. So Adam could be $5 \times$ faster than GD, w.h.p., under block-heterogeneity.

 In short,  the complexity of Adam is now independent of the ``global'' condition number $\kappa$, and such changes could lead to considerable improvement over GD.  Such improvement in complexity is attributed to the properties of the considered function class: the block diagonal structure in Hessian as well as its heterogeneous blockwise spectra. We believe 
  these results can bring new insights to modern NN optimization problems.

\paragraph{Limit cycle of constant-step-size Adam under  $\beta_2 <1$.} \yushunrevise{ Theorem \ref{thm_gd_lower_bd}  considers $\beta_2 = 1$. Thus, it uses $v_0$, which is the initial gradient, as the fixed preconditioner at arbitrary step $k >0$. We note that $\beta_2 =1$ is necessary for analyzing the rate for constant-step-size Adam, as any $\beta_2 <1$ will incur non-convergence with limit cycle on quadratic objectives \citep{da2020general,bai2026towards}. The non-convergence is due to constant step size, and it does not contradict our convergence results in Section \ref{sec_adam}, which considers diminishing step size.}

\paragraph{Source of block-heterogeneity.} \yushunrevise{The source of block-heterogeneity is later partially attributed to (i) the imbalanced data \citep{kunstner2024heavy}, which brings block-heterogeneity in the  Hessian of the output layer (Figure 8 and Proposition 2 in \citep{kunstner2024heavy}); and (ii) softmax operator in architecture \citep{ormaniec2024does}, which brings block-heterogeneity in the  Hessian of the Attention module.}

\begin{figure}[t]
 \vspace{-2cm}
    \centering
    \subfigure[Hessian of the embedding layer at initialization]{\includegraphics[width=0.32\textwidth]{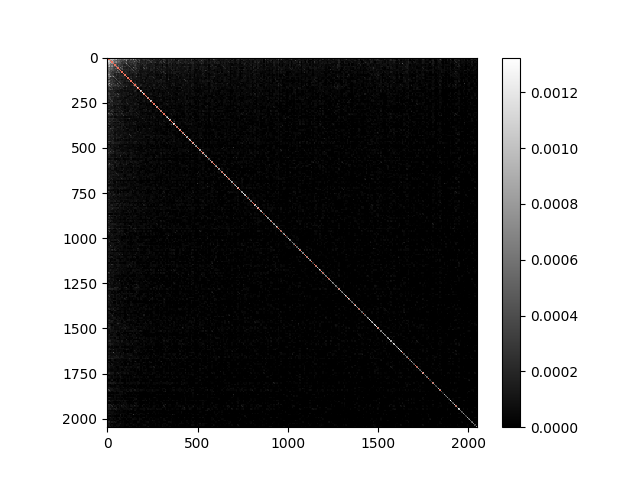}}
    \subfigure[Hessian the embedding layer at 50\% steps]{\includegraphics[width=0.32\textwidth]{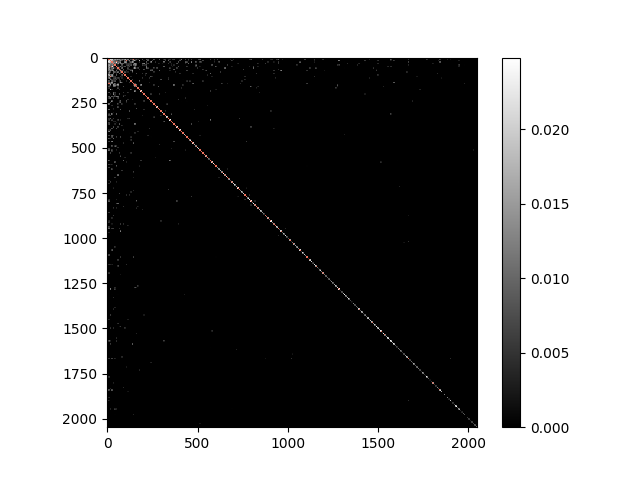}}
    \subfigure[Hessian the embedding layer at 100\% steps]{\includegraphics[width=0.32\textwidth]{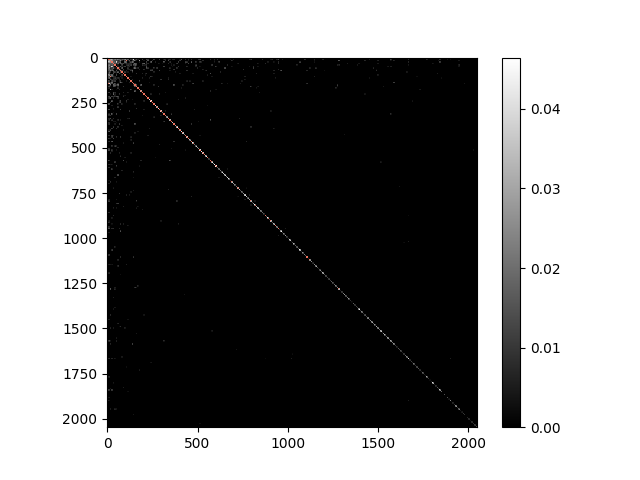}}
    \subfigure[Hessian of the output layer at initialization]{\includegraphics[width=0.32\textwidth]{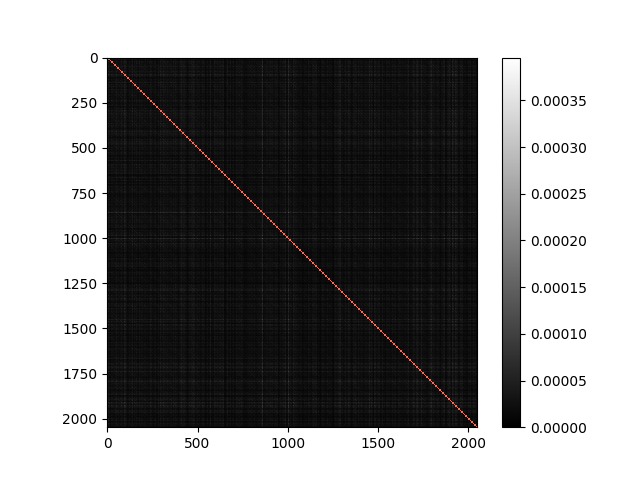}}
    \subfigure[Hessian of the output layer at 50\% steps]{\includegraphics[width=0.32\textwidth]{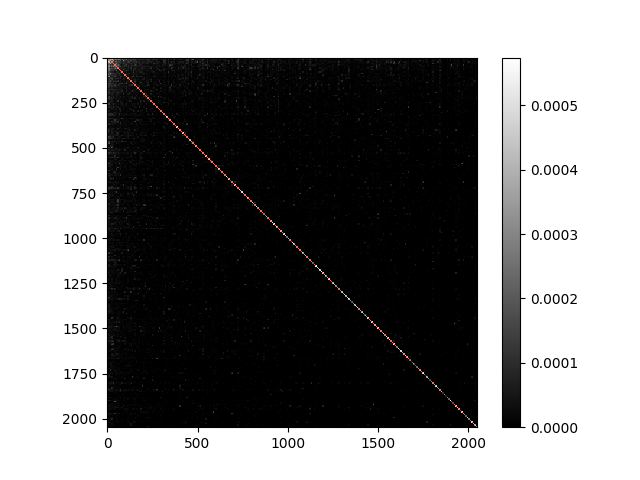}}
    \subfigure[Hessian of the output layer at 100\% steps]{\includegraphics[width=0.32\textwidth]{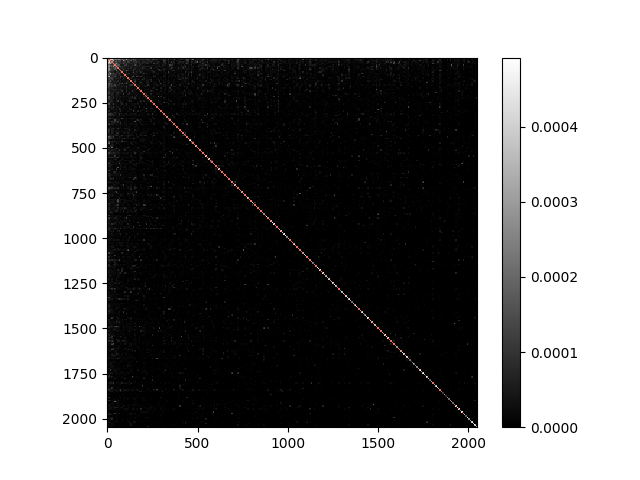}}
    \caption{\small  The Hessian of {\bf(a-c)} the embedding and {\bf(d-f)}  the output layer of a small Transformer on imbalanced-label data. Red color is used for pixels with values exceeding the color bar’s upper limit.  In each figure, one block corresponds to one token. The total number of blocks is 256, which equals the vocabulary size. In all figures except for {\bf(d)}, the magnitude of each block is related to the frequency of the corresponding token in the data set, which brings Hessian heterogeneity. }
  \label{fig:closer_hessian_embedding}
\end{figure}

In Figure \ref{fig:closer_hessian_embedding}, we provide more empirical evidence for the argument (i) above.  We track the Hessian of a small Transformer on imbalanced-label data following Zipf's law \citep{zipf1949human}. We set the vocabulary size to 256. We summarize our empirical findings below and leave more theoretical investigation as a future direction.
 \begin{itemize}
     \item  The Hessian of the embedding and the output layer are both near block diagonal, where one block corresponds to one token embedding. The number of blocks is 256, which equals the vocabulary size. 
     \item The Hessian of the embedding layer exhibits block heterogeneity for both initialization and throughout training. The magnitude of each block (i.e., token) is related to the frequency of this token in the data set.  Meanwhile, the Hessian of the output layer exhibits homogeneity at initialization, and the heterogeneity arises during training.
 \end{itemize}

\paragraph{Broader implications and Muon.} We believe our results have broad implications. 

{\bf First: matrix optimization.} Besides NNs, our insights can be applied to a broad class of non-convex optimization with consecutive matrix products, e.g., matrix factorization / completion. The connection is discussed in recent theoretical work on matrix factorization  \citep{ma2026preconditioning}. 

{\bf Second: why Muon.} our results also help understand why Muon \citep{jordan2024muon} performs well on NNs. Given a weight matrix $W \in \mathbb{R}^{m \times n}$ and its momentum matrix $M_k \in \mathbb{R}^{m \times n}$ at step $k$, the update rule of  Muon is as follows: 
{\small
\begin{equation}
\label{eq_muon}
    W_{k+1} = W_k - \eta_k \left( M_k M_k^\top \right)^{-\frac{1}{2}} M_k
         = W_k - \eta_k \, I_m \, M_k \left( M_k^\top M_k \right)^{-\frac{1}{2}}
\end{equation}
}
In the flattened vector form, \eqref{eq_muon} is equivalent to the following {\it block-diagonal preconditioner:}
{\small
\begin{eqnarray}
\label{eq_muon_flatten}
    \operatorname{vec}(W_{k+1})
 &\overset{(*)}{ = } & \operatorname{vec}(W_k)
  - \eta_k \left( I_m \otimes \left( M_k^\top M_k \right)^{-\frac{1}{2}} \right)_{mn \times mn}
    \operatorname{vec}(M_k), \nonumber \\
 & =&  \operatorname{vec}(W_k)
  - \eta_k  \begin{bmatrix}
M_k^\top M_k & \cdots     & 0 \\
\vdots     & \ddots & \vdots \\
0              & \cdots & M_k^\top M_k
\end{bmatrix}^{-\frac{1}{2}}_{mn \times mn} \operatorname{vec}(M_k)
\end{eqnarray}
}
where $(*)$ uses the property of Kronecker product $
(A \otimes B)  \operatorname{vec}(C) = \operatorname{vec}(A C B)$. As such, Muon can be viewed as a block-diagonal preconditioned GD, where one block corresponds to one row in $W$. The block partition is consistent with the Hessian structure of NNs.

{\bf Third: head-wise Muon.} our findings advocate ``head-wise Muon''. For Transformers, Figure~\ref{fig_transformer_hessian} suggests that we should partition the Query and Key weights by head before applying the Muon update. This provides a separate preconditioner for each head, which better handles the block heterogeneity across heads. 

{\bf Forth: about the embedding and output layer.} we explain why Muon is not the best option for the embedding and output layer: As shown in  \eqref{eq_muon_flatten}, Muon applies the same preconditioner to every token, which cannot handle the Hessian heterogeneity across tokens as in Figure \ref{fig:closer_hessian_embedding}.

\paragraph{Practical adoption.}  The design of ``head-wise Muon'' has been tested and then adopted by DeepSeek V4.1 Flash \citep{deepseekai2026v41flash}, where the authors cited \citep{zhang2024transformers} and an early version of this script as the references for this design.

\yushunrevise{Concurrently, the head-wise Muon design is also adopted in Kimi K3 model \citep{team2026kimi}, as referred as ``Per-head Muon''. As mentioned by the key contributor, the intuition of such a design is that ``Each head is inherently relatively independent and should not be coupled together \citep{kexuefm-11848}.''  This intuition aligns with our observation on head-wise block Hessian structure. }

\clearpage
\section{\AdamMini: A New Optimizer Motivated by the Hessian Structure}
\label{sec_adam_mini}
\begin{Snugshade}[gray!10]
{\small This section is mainly based on:
\begin{center}
\begin{itemize}
     \item \BLUE{\bf Adam-mini: Use Fewer Learning Rates To Gain More.}
     
     {\bf Yushun Zhang$^*$}, Congliang Chen$^*$, Ziniu Li, Tian Ding, Chenwei Wu, Diederik P. Kingma, Yinyu Ye, Zhi-Quan Luo, Ruoyu Sun.

    (*: Equal contribution.)
    
     International Conference on Learning Representations (ICLR), 2025.
\end{itemize}
\end{center}
}
\end{Snugshade}

By collecting the results accumulated in the previous sections, we now present  Adam-mini, a new optimizer that achieves on-par or better performance than Adam with  50\% less memory footprint.  This demonstrates that our preceding Hessian analysis is not only useful for understanding Adam, but can also directly motivate the design of new optimizers.

\paragraph{Motivation.} Adam requires the memory for its ``optimizer states'': the 1st-order and 2nd-order momentum $m$ and $v$. These in total take at least $2\times$ the memory of the model size.
This memory consumption  burdens LLM training: to train a 13B model, Adam alone requires about 104 GB for $m$ and $v$. This is expensive for advanced graphics cards (e.g., A100-80GB). To support training,
CPU-offload and optimizer state sharding \citep{rajbhandari2020zero} must be used in practice, which unfortunately increases the latency and slows down the training \citep{rajbhandari2021zero}. 
\begin{figure}[h]
  \centering
  \begin{minipage}{0.48\textwidth}
    \centering
    \begin{algorithm}[H]   %
      \caption{One step of Adam-mini}
      \label{algorithm_adam_mini}
      \begin{algorithmic}[1]
        \STATE{Partition params into \texttt{param\_blocks} by Hessian structure}
        \FOR{\texttt{param} in \texttt{param\_blocks}}
          \STATE{\texttt{g} = \texttt{param.grad}}
          \STATE{$\texttt{m} = (1-\beta_1) * \texttt{g} + \beta_1 * \texttt{m} $}
          \STATE{\BLUE{$\texttt{v}_{\operatorname{mean}} = (1-\beta_2)*\texttt{mean}(\texttt{g} \odot \texttt{g}) + \beta_2*\texttt{v}$}}
          \STATE{\texttt{param} = \texttt{param} - $\eta$ * $\frac{\texttt{m}}{\sqrt{\texttt{v}_{\operatorname{mean}}  } + \epsilon}$}
        \ENDFOR
      \end{algorithmic}
    \end{algorithm}
  \end{minipage}
  \hfill
  \begin{minipage}{0.48\textwidth}
    \centering
    \includegraphics[width=\linewidth]{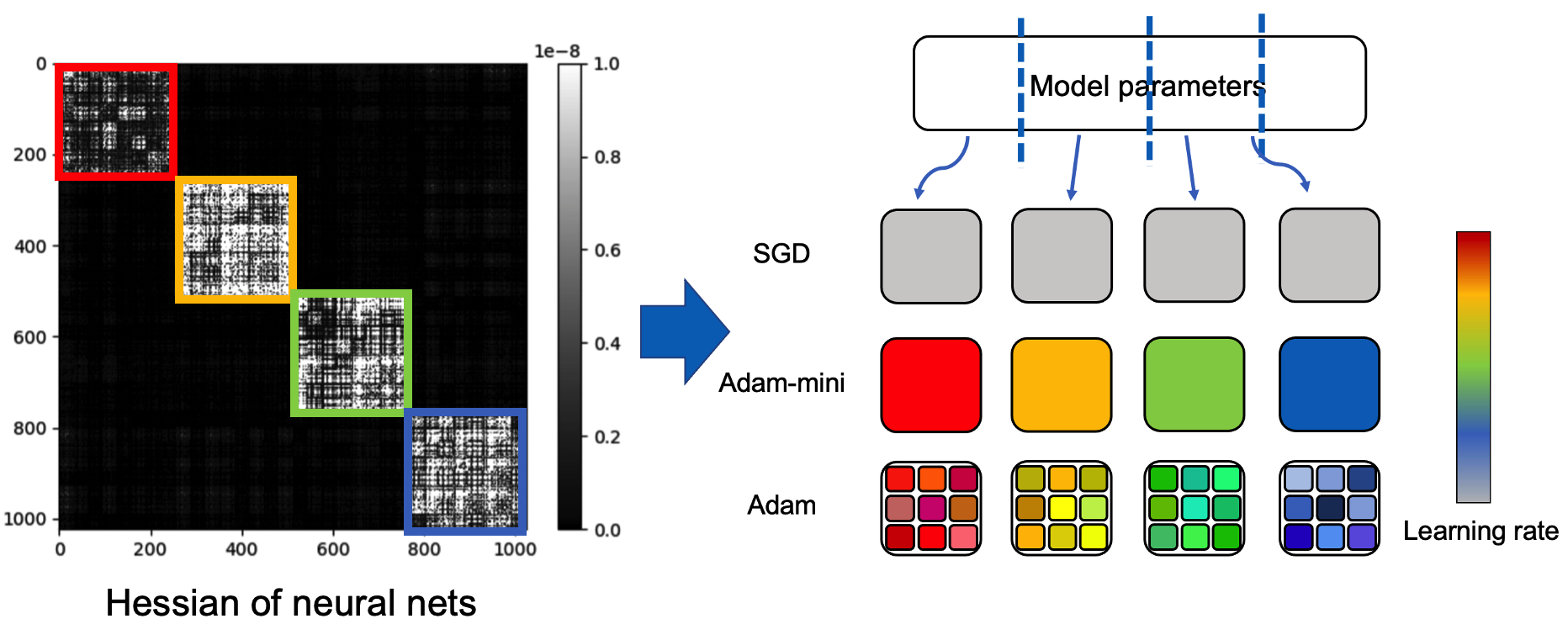}
    \captionof{figure}{An illustration of Adam-mini. Adam-mini assigns learning rates (lrs) by Hessian structure. It uses more lrs than SGD but fewer than Adam.}
    \label{fig_square_plot}
  \end{minipage}
  \vspace{-0.5cm}
\end{figure}
\paragraph{Design idea of Adam-mini.} 
We propose to cut down the learning rate resources in Adam, i.e., $1/\sqrt{v}$. This is motivated by two observations from Section \ref{sec_hessian}. 
\begin{itemize}
    \item First, the findings in Section \ref{sec_hessian} suggest that it is necessary to use a customized learning rate for {\it each block} to handle the block-heterogeneity. Nonetheless, Adam does much more than that: it assigns an individual learning rate not just for each block, but for {\it each parameter}. 
    \item Second,  as shown in Figure \ref{fig:adam-preconditioner-block}, Adam is not effective in dense Hessian blocks. In other words, we find that more learning rates do not necessarily bring extra gain. 
\end{itemize}

Our findings imply that it is possible to achieve good performance with much fewer learning rates than in Adam. 
The remaining question is how to find them efficiently.

We then propose a simple way to find good learning rates that are sufficient to perform on-par or better than Adam: we first partition the gradient vector into $B$ sub-vectors according to the dense Hessian sub-blocks, and call each $g_b$ for $b \in \{1, \cdots, B\}$. For each $g_b$, we calculate the quantity below.
\begin{small}\begin{align}
\label{eq_adam_mini_vmean}
\boxed{
v_{\operatorname{mean},b} = (1 - \beta_2)  * \texttt{mean} (g_b \odot g_b) + \beta_2 * v_b, \quad b = 1, \cdots, B.
}
\end{align}
\end{small}
We then use $\eta/\sqrt{v_{\operatorname{mean},b}}$ as the learning rate for the parameters in the block associated with $g_b$. We call the method Adam-mini.  Such design changes   $\geq 99.9\%$ of Adam's $v$ to a negligible amount of scalars and thus reduces the memory. Overall, it saves about $50\%$ of Adam's optimizer-state memory.

\paragraph{Partition principle.}  We partition the parameters by Hessian structure. According to Figure \ref{fig_transformer_hessian}, this corresponds to two cases in Transformers. 
\begin{itemize}
    \item {\bf Case 1:} we partition Query and Key weight matrices {\it by heads}.
    \item 
 {\bf Case 2:} we partition other weight matrices {\it by rows}, or equivalently, by output neurons. 
\end{itemize}

In matrix form, such partition strategy is simple and straightforward to implement: given a momentum matrix {\small $M_k  = \begin{bmatrix}
m_{k,1}^\top  \\
\vdots     \\
m_{k,m}^\top          
\end{bmatrix} \in \mathbb{R}^{m \times n}$}, for {\bf Case 2}, we simply normalize $M_k$ row‑wise by $v_{\operatorname{mean}} \in \mathbb{R}^m$:
{\small
\begin{eqnarray}
\label{eq_adam_mini_matrix_form}
    W_{k+1}
 & =&  W_k
  - \eta_k  \begin{bmatrix}
 \RED{v_{\operatorname{mean},1}} & \cdots     & 0 \\
\vdots     & \ddots & \vdots \\
0              & \cdots & {\color{green} v_{\operatorname{mean},m}}
\end{bmatrix}^{-\frac{1}{2}}_{m \times m}  \begin{bmatrix}
m_{k,1}^\top  \\
\vdots     \\
m_{k,m}^\top          
\end{bmatrix}_{m \times n} = \begin{bmatrix}
m_{k,1}^\top  /  \RED{\sqrt{v_{\operatorname{mean},1}}}  \\
\vdots     \\
m_{k,m}^\top  / {\color{green} \sqrt{v_{\operatorname{mean},m}}}         
\end{bmatrix}_{m \times n} .
\end{eqnarray}
}
Similarly, for {\bf Case 1}, we normalize $M_k$ head‑wise by $v_{\operatorname{mean}} \in \mathbb{R}^{h}$, and each group consists of $m /h$  consecutive rows corresponding to a single head, where $h$ denotes the number of heads.

We note that both {\bf Case 1} and {\bf Case 2} are naturally compatible with distributed setups. As the head and row are usually not sharded across distinct GPUs, computing $v_{\operatorname{mean}}$
  requires no communication overhead.

\begin{figure}[h]
\centering
\begin{minipage}{0.56\linewidth}
\centering
\includegraphics[width=\linewidth]{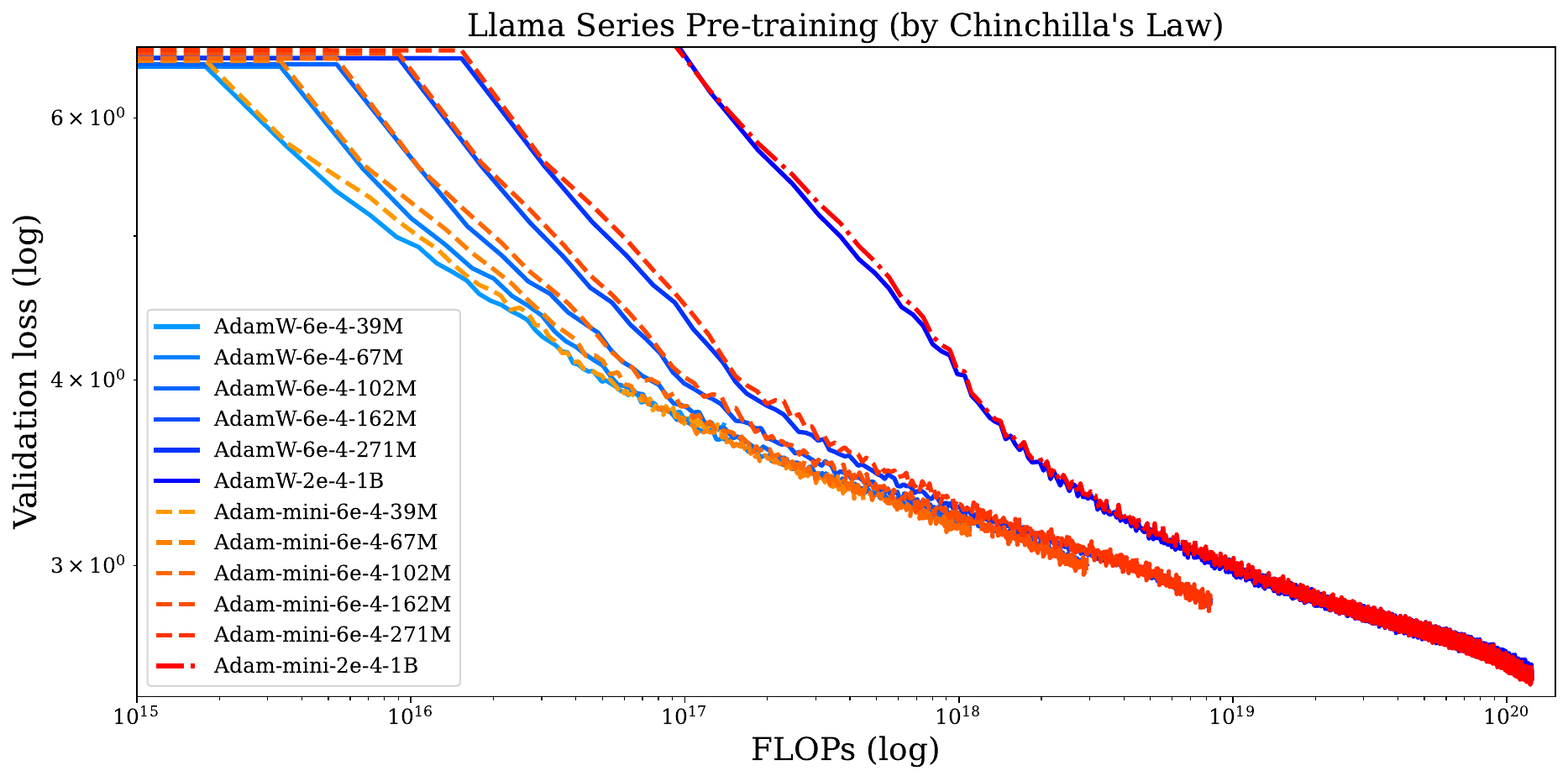}
\end{minipage}\hfill
\begin{minipage}{0.38\linewidth}
\centering
\includegraphics[width=\linewidth]{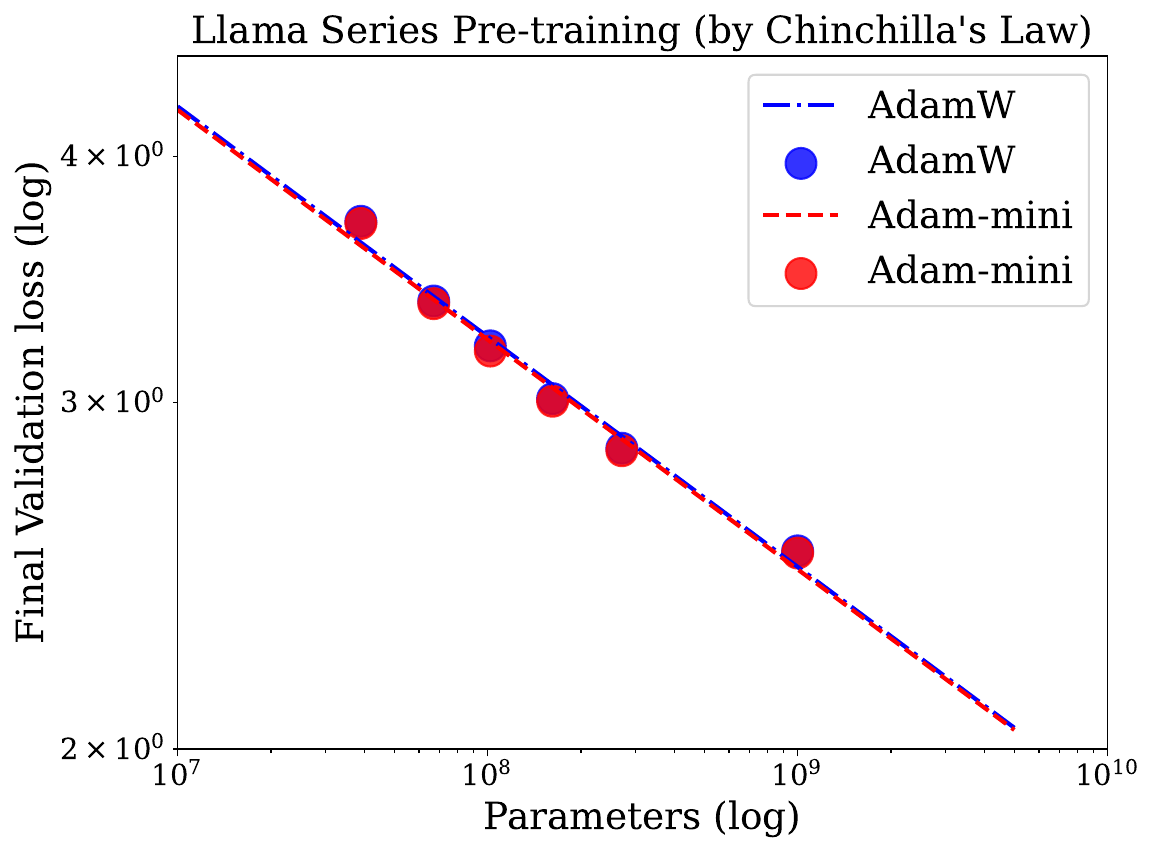}
\end{minipage}
\caption{\small Scaling-law experiments on Llama 2 from 39M to 1B parameters. \AdamMini{} tracks \AdamW{} across compute and model scales while using $50\%$ less memory footprint.}
\label{fig:scaling-law}
\end{figure}

\paragraph{Empirical results.}
We pre-train Llama 2 architectures from 39M to 1B parameters on C4 using Chinchilla-style token budgets \citep{hoffmann2022training}. Figure \ref{fig:scaling-law} shows that \AdamMini{} closely tracks \AdamW{} across both compute and model scales, while using half the optimizer-state memory.

\paragraph{Reproducibility, open-source, and diversity.} The effectiveness of Adam-mini is independently reproduced by a recent benchmark by Stanford  researchers \citep{wen2025fantastic}  with the conclusion that Adam-mini {\it ``closely tracks the performance of AdamW''} and {\it ``sometimes even performs better than AdamW''}. 
We open sourced Adam-mini on \href{https://github.com/zyushun/Adam-mini}{\BLUE{[GitHub]}} and have received 400+ stars, along with 3000+ monthly downloads   via \href{https://pypistats.org/packages/adam-mini}{\BLUE{[PyPI]}}.

Adam-mini is also used by diverse researchers. For instance, particle physicists \citet{Buss2025CaloHadronicAD} want to train large diffusion models, yet find that AdamW is  {\it ``memory-intensive''}, and all experiments are conducted with Adam-mini.  Researchers are also using the codebase of Adam-mini to conduct other research, such as architecture design \citep{chen2026simplegpt}. All these show that Adam-mini enables scientists from diverse fields to carry out research under limited GPU resources.

\paragraph{NorMuon: Adam-mini + Muon.} The Adam-mini idea can also be used to improve Muon. First of all, we restate the update form of Muon here:  
{\small
\begin{eqnarray}
\label{eq_muon_flatten_2}
    \operatorname{vec}(W_{k+1})
 &\overset{}{ = } & \operatorname{vec}(W_k)
  - \eta_k \left( I_m \otimes \left( M_k^\top M_k \right)^{-\frac{1}{2}} \right)_{mn \times mn}
    \operatorname{vec}(M_k), \nonumber \\
 & =&  \operatorname{vec}(W_k)
  - \eta_k  \begin{bmatrix}
M_k^\top M_k & \cdots     & 0 \\
\vdots     & \ddots & \vdots \\
0              & \cdots & M_k^\top M_k
\end{bmatrix}^{-\frac{1}{2}}_{mn \times mn} \operatorname{vec}(M_k).
\end{eqnarray}
}
\yushunrevise{As shown in \eqref{eq_muon_flatten_2} and discussed previously in \eqref{eq_muon_flatten}, Muon is a block-diagonal preconditioner that aligns with the Hessian structure, but it applies the same preconditioner \(M_k^\top M_k\) to every neuron within the same layer (i.e., every row of \(M_k\)). 
This makes it unable to handle neuron-wise heterogeneity \citep{zhang2024transformers,dewulf2026aurora}.}

NorMuon \citep{li2025normuon} addresses this limitation by augmenting Muon with neuron-wise learning rates. That is, given a weight $W \in \mathbb{R}^{m\times n}$ and momentum $M_k \in \mathbb{R}^{m\times n}$, NorMuon changes the Muon update rule in \eqref{eq_muon_flatten} into the following form. 
{\small
\begin{eqnarray}
\label{eq_normuon_flatten}
    \operatorname{vec}(W_{k+1})
 &\overset{}{ = } &  \operatorname{vec}(W_k)
  - \eta_k  \begin{bmatrix}
\RED{v_{\operatorname{mean},k,1}}M_k^\top M_k & \cdots     & 0 \\
\vdots     & \ddots & \vdots \\
0              & \cdots & {\color{green} v_{\operatorname{mean}, k,m}} M_k^\top M_k
\end{bmatrix}^{-\frac{1}{2}}_{mn\times mn} \operatorname{vec}(M_k),
\end{eqnarray}
}
where $v_{\operatorname{mean},k,j} \in \mathbb{R}$ for $ j \in [m]$ are the newly designed neuron-wise learning rates, and they are designed by block-wise mean as in Adam-mini in  \eqref{eq_adam_mini_vmean}. This design does not bring additional memory and computational overhead.
NorMuon is now the leading method in the NanoGPT speedrun. In particular, all three fastest methods use NorMuon \citep{jordan2026modded}. 

\paragraph{Row-wise normalized methods.} Besides NorMuon, there is a recent surge of new row-wise normalized methods that are motivated by Adam-mini (e.g., RMNP  \citep{deng2026rmnp}, Nora \citep{yuan2026nora}, and Sron \citep{wen2025sron}). These methods assign different learning rates to different rows in $W$, which matches the block structure of Hessian. The detailed form of these learning rates varies from Adam-mini, and some designs are shown to match Muon.  

\yushunrevise{Finally, the row-wise normalized design in Adam-mini can also be combined with other optimizers to save memory. For instance, such ideas are used in Apollo and Apollo-mini \citep{zhu2025apollo}.}

\section{Conclusion}
\label{sec_conclusion}

This thesis establishes a systematic investigation into the theoretical grounding and principled foundation of NN optimizers. We begin by investigating fundamental concerns about Adam's convergence guarantees, progress to uncover Hessian-structure-based explanations for Adam-SGD gaps, then dive deeper to reveal the origin behind the special Hessian structure, and ultimately translate these insights into a new optimizer design called Adam-mini. This thesis helps shrink the gaps between optimization theory and large-scale practice, offering both rigorous guarantees and actionable principles for safer, more reliable, more efficient, and better-understood training of neural networks.

\paragraph{Future directions.} Finally, we point out some future directions. 
\begin{itemize}
    \item First, our Adam theory proves the existence of a divergence-convergence critical boundary, but its precise shape and  uniqueness remain to be characterized.
    \item Second, it is intriguing to extend rigorous Hessian analysis to modern architectures and their key components, such as attention and SwiGLU-based MoEs. Despite the complexity of these architectures,  the Hessian structure can be predicted following computational-graph-based analysis in Figure \ref{fig_computation_graph} in Section \ref{sec_hessian}. That is, the Hessian structure can be predicted 
    by asking ``whether there exists a connected path between any two distant links in the
NN computation graph.''
    \item  Third, these Hessian structures may provide useful design principles for more effective second-order optimizers.
\end{itemize}

\section*{Major References}

This PhD dissertation is composed of the following references. 

\begin{itemize}
    \item[1] \BLUE{\bf Adam Can Converge Without Any Modification On Update Rules.}  \\
    {\bf Yushun Zhang}, Congliang Chen, Naichen Shi, Ruoyu Sun,  Zhi-Quan Luo. 
    
    Advances in Neural Information Processing Systems (NeurIPS), 2022 \RED{\bf (Spotlight)}.
    \item[2]  \BLUE{\bf Adam Converges Without Any Modification On Update Rules.}  \\
    {\bf Yushun Zhang}, Bingran Li, Congliang Chen, Zhi-Quan Luo, Ruoyu Sun. 
    
   (This is the journal extended version of paper 1.)

   Under Review.

    \item[3]  \BLUE{\bf Why Transformers Need Adam: A Hessian Perspective.}  \\
    {\bf Yushun Zhang}, Congliang Chen, Tian Ding, Ziniu Li, Ruoyu Sun, Zhi-Quan Luo. 
   
     Advances in Neural Information Processing Systems (NeurIPS), 2024.
     \item[4] \BLUE{\bf Towards Quantifying the Hessian Structure of Neural Networks.}

     Zhaorui Dong$^*$, {\bf Yushun Zhang$^*$}, Jianfeng Yao, Ruoyu Sun
     
    (*: Equal contribution. Alphabetically ordered.)

    IEEE Transactions on Information Theory, 2026.

     \item[5] \BLUE{\bf Adam-mini: Use Fewer Learning Rates To Gain More.}
     
     {\bf Yushun Zhang$^*$}, Congliang Chen$^*$, Ziniu Li, Tian Ding, Chenwei Wu, Diederik P. Kingma, Yinyu Ye, Zhi-Quan Luo, Ruoyu Sun.

    (*: Equal contribution.)
    
     International Conference on Learning Representations (ICLR), 2025.
     
\end{itemize}

\clearpage

\section*{Acknowledgment}

First and foremost, I would like to express my deepest gratitude to my supervisor, Prof. Zhi-Quan Luo, for his continuous guidance and support.  When I first came to Prof. Luo's group six years ago, I knew nothing about research.  Through hundreds of hours of weekly group meetings and one-to-one meetings,  he taught me how to think critically, find valuable research problems, ask fundamental questions, and solve a problem in a principled way. 
Despite his busy schedule, Prof. Luo hosted three-hour group meetings every week, where he always asked insightful and fundamental questions that pushed me to think more deeply about the problem. He encouraged me to actively participate in the group meeting, learn to ask good questions, and explain things clearly to other members from diverse backgrounds. During my presentations, he forced me to be responsible for every word on my slides, and to anticipate potential questions from the audience. All of these significantly improve my presentation skills and logical thinking. I have never experienced another group meeting as valuable as this one.  He often reminds me that “Life is short, do something useful,” which has strongly influenced my research attitude and view of life. His mentorship has shaped not only my academic work but also my growth as a researcher.
I hope to one day become a researcher and advisor like him.

I would like to sincerely thank my thesis committee members, Prof. Ruoyu Sun, Prof. Jim Dai, Prof. Yinyu Ye, Prof. Anthony So, and Prof. Madeleine Udell, for their valuable time and support. Special thanks to Prof. Ruoyu Sun for his generous support and profound influence on me. Throughout the years, he offered me advice ranging from career choices, high-level research directions, unique views on diverse research problems, to the finest technical details.   I will never forget the countless long and in-depth discussions with Prof. Sun on every detail of my research topics.  He has also spent tremendous amount of time revising my paper writing from word to word, slides from page to page, and once spent an entire week teaching me how to do rebuttal.  These discussions with Prof. Sun are the best memories in my PhD life.   He also generously arranged meetings at conferences, invited me to talks, and created opportunities for me to interact with researchers from diverse backgrounds.    Prof. Sun has taught me many more things than I could have expected from a professor.

I would like to thank all the professors who taught me during my PhD studies. In particular, Prof. Ruoyu Sun's summer course “Optimization for Deep Learning” in 2021 led me through the grand gate of research and fundamentally shaped my academic path. I am grateful for other optimization theory courses taught by Prof. Stark Draper, Prof. Shuzhong Zhang, Prof. Andre Milzarek, and Prof. Yinyu Ye. Many technical tools in my research came from courses in matrix analysis and numerical linear algebra by Prof. Tsung-Hui Chang and Prof. Hongyuan Zha; real analysis by Prof. Weiming Ni; measure theory by Prof. Yutian Li; probability theory and random matrices by Prof. Jeff Yao; mathematical analysis by Prof. Xuefeng Wang; topology by Prof. Yong Hou; and high-performance computing by Prof. Zizhong Chen. 

I also thank my undergraduate professors at SUSTech for the courses that built my mathematical foundation: Prof. Xianke Zhang, whose linear algebra course attracted me to the Department of Mathematics; Prof. Bingsheng He, for the basic courses of mathematical analysis, linear algebra, and optimization; Prof. Anyue Chen and Prof. Yong Wang, for the basic course in probability and differential equations; Prof. Jiang Yang and Liuqing Yang for the course in numerical analysis and complex analysis, which have been repeatedly used in my research;  Prof. Jingheng Cai, for his course in Bayesian statistics and for the lunch at Sun Yat-sen University; and Prof. Guoliang Tian, who taught me statistics and guiding my initial research attempts.

I would like to express special thanks to Prof. Yinyu Ye for many one-to-one discussions in his temporary office during his visits to our campus. His sharpness, vision, and passion for research have deeply influenced me. I also sincerely thank Prof. Mingyi Hong, who hosted me for dinner every time I traveled to the US. I am grateful for the great care of Prof. Hong and his group members during my visit to Minnesota, and for the delightful lunch at his house. 
I am also grateful to Prof. Zizhuo Wang, Prof. Xiao Li, and Prof. Benyou Wang for offering me various opportunities to meet diverse researchers. I also thank Prof. Yang Yu and Prof. Jian Li for inviting me to give talks at Nanjing and Tsinghua University.  
I am deeply thankful to Prof. Stephen Boyd for the inspring long discussions on my research during his visit to our campus.
Special thanks go to  Prof. Madeleine Udell for multiple in-depth discussions and for offering me an opportunity to work with her in the future.

I am fortunate to have worked with many outstanding friends and collaborators. I would like to first thank my best friend Ziniu Li, who accompanied and encouraged me during difficult times and continuously supported my growth both academically and personally. His critical thinking, extraordinary level of concentration, and optimistic attitude toward life have greatly influenced me. I also deeply thank Congliang Chen for being an excellent and crucial co-author for years. We have co-authored three long papers together, and I learned a lot from his sharp mind and in-depth thinking. I also sincerely thank Jiawei Zhang, who advised me on my first research topic on the landscape theory and taught me how to conduct rigorous theoretical research.  I am grateful to Naichen Shi for encouraging me to explore the convergence of Adam in 2021 summer. I am also grateful to Bohan Wang for the insightful discussions on the technical errors in my proof draft. Bohan's sharp intuition on theory influenced me a lot. I am also grateful to Dmitry Rybin for his sharp theoretical insights into the project of finite-horizon optimization.  
I really enjoyed our discussions on matrix polynomials by the whiteboard on the six floor of the Daoyuan Building.
Special thanks to Zhaorui Dong for leading me into the fantastic world of random matrix theory. 
I also thank Bingran Li, Senmiao Wang, and Yupeng Chen for the meticulous and reliable collaborations on theory and algorithm designs.

Besides my collaborators, I would also like to thank all the friends I met during my PhD study. First of all, I am thankful to our team, including  Jiancong Xiao, Yingru Li, Hao Liang, Tianjian Zhang, Liangqi Liu, Shuyi Ren, Qingyan Meng, Kai Li, Wenqiang Pu, Ye Liu, Boyuan Wang, Zeyu Qin, Xiaopeng Li, Yeqing Qiu, Juntao Wang, Yi Feng, and Runhan Yang, and many others. All of whom have helped me in many ways during my studies. Special thanks to Kai Li, my wonderful roommate, for making daily life during my PhD both enjoyable and memorable. I am also grateful to many friends outside the group. Special thanks go to my most trusted friends Xueyao Zhang and Ke Xue for exchanging ideas about life and research, and encouraging me during difficult times.  I am deeply thankful to Tian Ding and Dawei Li for the insightful discussions on optimization landscapes.  Tian also organized wonderful seminars on information theory and GPU architecture.  Special thanks to Peijun Xiao and Zhisheng Xiao for hosting dinner parties during my visits to the US.  I am grateful to Guiyu Hong and Kang Gong for insightful discussions on probability theory.  I am also grateful to  Yingru Li, Shuaiyu Jiang, and Jiawei Shan for organizing multiple seminars on high-dimensional statistics, in which I learned helpful tools for my research. I am thankful to Kaiyue Wen for the inspiring discussions on pre-training optimizers during his visit to our campus. Special thanks to Peiying Hu for the wonderful piano lectures. I deeply thank my SDS classmates Jing Zhang, Xiuwen Wang, and Junwen Qiu for their companionship in the early years. 
I also thank Jianhao Ma, Kaifeng Lyu, Jiaye Teng, Bohan Wang, Yiping Lu, and Jiaxiang Li for inviting or hosting my talks at various conferences. Finally, I am deeply thankful to all the group members in Prof. Sun's group, including Zhihang Lin, Ziwei Zhu, Zhuohan Wang, and all others, for the valuable discussions and fun parties.

I would like to thank my friends from industry. I am deeply grateful to my mentor during my internship at Moonshot AI, Jianlin Su, for pointing out the valuable research direction of MuP and all the in-depth discussions. I thank Yunpeng Tai for developing the AtomGPT codebase, which exponentially eased my research. I am grateful to Zhuorui Ye for our excellent collaboration on the MuP project, and to Tong Gao, Zhengying Liu, and other colleagues for the daily discussions that illuminate my life. I am particularly grateful to Zhengyan Zhang from DeepSeek for acknowledging my research, conducting multiple in-depth discussions, and offering me the opportunity for an internship.
I am grateful to Lechao Xiao from Google DeepMind for his valuable advice on career choices. Special thanks to Dr. Diederik Kingma for inviting me to present my work at the Google Brain and for the collaboration on Adam-mini. I am truly grateful for Dr. Kingma's insightful technical suggestions and detailed revisions on paper writing.   I also sincerely thank Dr. Kingma and Prof. Jimmy Ba for mentioning two of my works in the ICLR 2025 Test of Time Award speech. 

I am thankful to the teachers and friends I have met since childhood. In particular, I am sincerely grateful to all my teachers and classmates from class seven, Shenzhen Foreign Languages School, Branch (SZFLSB)--Those were the best three years of my life, and SZFLSB will always be my second home. Special thanks to Mrs. Junying Ma, Mrs. Zhenghua Li, Mrs. Qin Fang, Mr. Yixin He, Mr. Buck, Ms. Miffy, and Mrs. Jennifer for showing extraordinary professionalism and deep love for their students.  Special thanks to Tianzhe Yang, Yu Liang, and Yifan Li for continuous companionship since we were teenagers.  I am also grateful to my high-school roommates Liwei Zhang, Junnian Chen, and Mingbin Tan for the joy and light they have brought into my life.
 I also thank all my friends at SUSTech, especially Yiqi Lin, Mingchen Huang, and Meng Zhou, for their continuous support. I am grateful to my French horn teachers, Mr. Yi Lin and Mrs. Hong Han, for their six years of guidance in the music world. Special thanks to Jay Chou, whose music has been my spiritual support for more than twenty years and a constant companion during long days of research in the library.

Finally, I wish to express my deepest gratitude to my parents, Xiaojin Zhao and Meiyu Zhang, my grandparents, Guizhen He and Fashun Zhao, for taking pride in me and unconditional love, understanding, and support. None of this would have been possible without them.

\clearpage

{\singlespacing
\bibliography{database}

@article{da2020general,
  journal = {The Journal of Machine Learning Research},
  publisher = {JMLRORG},
  volume = {21},
  number = {1},
  pages = {5072--5113},
  author = {Da Silva, Andr{\'e} Belotto and Gazeau, Maxime},
  title = {A general system of differential equations to model first-order adaptive algorithms},
  year = {2020},
}

@ARTICLE{Chatterjee2006,
  journal = {Annals of Probability},
  volume = {34},
  number = {6},
  pages = {2061 -- 2076},
  author = {Chatterjee, Sourav},
  title = {A generalization of the lindeberg principle},
  year = {2006},
}

@article{yan2018unified,
  journal = {arXiv preprint arXiv:1808.10396},
  author = {Yan, Yan and Yang, Tianbao and Li, Zhe and Lin, Qihang and Yang, Yi},
  title = {A unified analysis of stochastic momentum methods for deep learning},
  year = {2018},
}

@article{kingma2014adam,
  journal = {arXiv preprint arXiv:1412.6980},
  author = {Kingma, Diederik P and Ba, Jimmy},
  title = {Adam: A method for stochastic optimization},
  year = {2014},
}

@article{zhang2022adam,
  journal = {Advances in neural information processing systems},
  volume = {35},
  pages = {28386--28399},
  author = {Zhang, Yushun and Chen, Congliang and Shi, Naichen and Sun, Ruoyu and Luo, Zhi-Quan},
  title = {Adam can converge without any modification on update rules},
  year = {2022},
}

@inproceedings{luo2018adaptive,
  booktitle = {International Conference on Learning Representations},
  author = {Luo, Liangchen and Xiong, Yuanhao and Liu, Yan and Sun, Xu},
  title = {Adaptive Gradient Methods with Dynamic Bound of Learning Rate},
  year = {2018},
}

@inproceedings{Manzil2018adaptive,
  url = {https://proceedings.neurips.cc/paper/2018/file/90365351ccc7437a1309dc64e4db32a3-Paper.pdf},
  journal = {Advances in neural information processing systems},
  booktitle = {Advances in Neural Information Processing Systems},
  publisher = {Curran Associates, Inc.},
  volume = {31},
  editor = {S. Bengio and H. Wallach and H. Larochelle and K. Grauman and N. Cesa-Bianchi and R. Garnett},
  author = {Zaheer, Manzil and Reddi, Sashank and Sachan, Devendra and Kale, Satyen and Kumar, Sanjiv},
  title = {Adaptive Methods for Nonconvex Optimization},
  year = {2018},
}

@inproceedings{dosovitskiy2021an,
  url = {https://openreview.net/forum?id=YicbFdNTTy},
  booktitle = {International Conference on Learning Representations},
  author = {Alexey Dosovitskiy and Lucas Beyer and Alexander Kolesnikov and Dirk Weissenborn and Xiaohua Zhai and Thomas Unterthiner and Mostafa Dehghani and Matthias Minderer and Georg Heigold and Sylvain Gelly and Jakob Uszkoreit and Neil Houlsby},
  title = {An Image is Worth 16x16 Words: Transformers for Image Recognition at Scale},
  year = {2021},
}

@article{liu2020improved,
  journal = {arXiv preprint arXiv:2007.07989},
  author = {Liu, Yanli and Gao, Yuan and Yin, Wotao},
  title = {An improved analysis of stochastic gradient descent with momentum},
  year = {2020},
}

@misc{ICLR2025TestOfTime,
  note = {Accessed 27 May 2025},
  howpublished = {\url{https://blog.iclr.cc/2025/04/14/announcing-the-test-of-time-award-winners-from-iclr-2015/}},
  month = {apr},
  urldate = {2025-05-27},
  author = {{ICLR}},
  title = {{Announcing the Test of Time Award Winners from ICLR 2015}},
  year = {2025},
}

@article{gotze2015asymptotic,
  journal = {Random Matrices: Theory and Applications},
  publisher = {World Scientific},
  volume = {4},
  number = {02},
  pages = {1550005},
  author = {G{\"o}tze, Friedrich and K{\"o}sters, Holger and Tikhomirov, Alexander},
  title = {Asymptotic spectra of matrix-valued functions of independent random matrices and free probability},
  year = {2015},
}

@inproceedings{vaswani2017attention,
  journal = {Advances in neural information processing systems},
  booktitle = {Advances in neural information processing systems},
  volume = {30},
  pages = {5998--6008},
  author = {Vaswani, Ashish and Shazeer, Noam and Parmar, Niki and Uszkoreit, Jakob and Jones, Llion and Gomez, Aidan N and Kaiser, {\L}ukasz and Polosukhin, Illia},
  title = {Attention is all you need},
  year = {2017},
}

@misc{openai_chatgpt_2022,
  url = {https://chat.openai.com/},
  note = {Accessed: 2025-05-15},
  author = {OpenAI},
  title = {ChatGPT (Nov 30 version) [Large language model]},
  year = {2022},
}

@article{liu2024deepseek,
  journal = {arXiv preprint arXiv:2412.19437},
  author = {Liu, Aixin and Feng, Bei and Xue, Bing and Wang, Bingxuan and Wu, Bochao and Lu, Chengda and Zhao, Chenggang and Deng, Chengqi and Zhang, Chenyu and Ruan, Chong and others},
  title = {Deepseek-v3 technical report},
  year = {2024},
}

@article{Pastur67,
  journal = {Mathematics of the USSR-Sbornik},
  volume = {1},
  number = {4},
  pages = {457},
  author = {V A Mar{\v{c}}enko and  L A Pastur},
  title = {Distribution of eigenvalues for some set of random matrices},
  year = {1967},
}

@ARTICLE{Lindeberg1922,
  journal = {Mathematische Zeitschrift},
  volume = {15},
  number = {1},
  pages = {211 -- 225},
  author = {Lindeberg, J.W.},
  title = {Eine neue Herleitung des Exponentialgesetzes in der Wahrscheinlichkeitsrechnung},
  year = {1922},
}

@article{achiam2023gpt,
  journal = {arXiv preprint arXiv:2303.08774},
  author = {Achiam, Josh and Adler, Steven and Agarwal, Sandhini and Ahmad, Lama and Akkaya, Ilge and Aleman, Florencia Leoni and Almeida, Diogo and Altenschmidt, Janko and Altman, Sam and Anadkat, Shyamal and others},
  title = {Gpt-4 technical report},
  year = {2023},
}

@article{kunstner2024heavy,
  journal = {arXiv preprint arXiv:2402.19449},
  author = {Kunstner, Frederik and Yadav, Robin and Milligan, Alan and Schmidt, Mark and Bietti, Alberto},
  title = {Heavy-Tailed Class Imbalance and Why Adam Outperforms Gradient Descent on Language Models},
  year = {2024},
}

@article{zhang2024does,
  journal = {arXiv preprint arXiv:2410.21676},
  author = {Zhang, Hanlin and Morwani, Depen and Vyas, Nikhil and Wu, Jingfeng and Zou, Difan and Ghai, Udaya and Foster, Dean and Kakade, Sham},
  title = {How Does Critical Batch Size Scale in Pre-training?},
  year = {2024},
}

@misc{ICLR2018BestPaper,
  note = {Lists \emph{On the Convergence of Adam and Beyond} as the Best Paper Award},
  howpublished = {\url{https://iclr.cc/Conferences/2018/ScheduleOverview}},
  month = {apr},
  urldate = {2025-05-27},
  author = {{ICLR}},
  title = {{ICLR 2018 Schedule Overview}},
  year = {2018},
}

@article{sreckovic2025your,
  journal = {arXiv preprint arXiv:2506.12543},
  author = {Sre{\'c}kovi{\'c}, Teodora and Geiping, Jonas and Orvieto, Antonio},
  title = {Is your batch size the problem? Revisiting the Adam-SGD gap in language modeling},
  year = {2025},
}

@article{brown2020language,
  journal = {arXiv preprint arXiv:2005.14165},
  author = {Brown, Tom B and Mann, Benjamin and Ryder, Nick and Subbiah, Melanie and Kaplan, Jared and Dhariwal, Prafulla and Neelakantan, Arvind and Shyam, Pranav and Sastry, Girish and Askell, Amanda and others},
  title = {Language models are few-shot learners},
  year = {2020},
}

@techreport{collobert2004large,
  institution = {Universit{\'e} de Paris VI},
  author = {Collobert, Ronan},
  title = {Large scale machine learning},
  year = {2004},
}

@article{touvron2023llama,
  journal = {arXiv preprint arXiv:2307.09288},
  author = {Touvron, Hugo and Martin, Louis and Stone, Kevin and Albert, Peter and Almahairi, Amjad and Babaei, Yasmine and Bashlykov, Nikolay and Batra, Soumya and Bhargava, Prajjwal and Bhosale, Shruti and others},
  title = {Llama 2: Open foundation and fine-tuned chat models},
  year = {2023},
}

@misc{jordan2024muon,
  url = {https://kellerjordan.github.io/posts/muon/},
  author = {Keller Jordan and Yuchen Jin and Vlado Boza and Jiacheng You and Franz Cesista and Laker Newhouse and Jeremy Bernstein},
  title = {Muon: An optimizer for hidden layers in neural networks},
  year = {2024},
}

@article{forsythe1955best,
  journal = {Proceedings of the American Mathematical Society},
  publisher = {JSTOR},
  volume = {6},
  number = {3},
  pages = {340--345},
  author = {Forsythe, George E and Straus, Ernst G},
  title = {On best conditioned matrices},
  year = {1955},
}

@article{pastur2020random,
  journal = {Pure and Applied Functional Analysis},
  volume = {5},
  number = {6},
  pages = {1395 -- 1424},
  author = {Pastur, Leonid},
  title = {On random matrices arising in deep neural networks: Gaussian case},
  year = {2020},
}

@inproceedings{reddi2019convergence,
  url = {https://openreview.net/forum?id=ryQu7f-RZ},
  booktitle = {International Conference on Learning Representations},
  author = {Sashank J. Reddi and Satyen Kale and Sanjiv Kumar},
  title = {On the Convergence of Adam and Beyond},
  year = {2018},
}

@inproceedings{yu2019linear,
  booktitle = {International Conference on Machine Learning},
  organization = {PMLR},
  pages = {7184--7193},
  author = {Yu, Hao and Jin, Rong and Yang, Sen},
  title = {On the linear speedup analysis of communication efficient momentum SGD for distributed non-convex optimization},
  year = {2019},
}

@article{qu2022optimal,
  journal = {arXiv preprint arXiv:2209.00809},
  author = {Qu, Zhaonan and Gao, Wenzhi and Hinder, Oliver and Ye, Yinyu and Zhou, Zhengyuan},
  title = {Optimal Diagonal Preconditioning: Theory and Practice},
  year = {2022},
}

@article{bai2023qwen,
  journal = {arXiv preprint arXiv:2309.16609},
  author = {Bai, Jinze and Bai, Shuai and Chu, Yunfei and Cui, Zeyu and Dang, Kai and Deng, Xiaodong and Fan, Yang and Ge, Wenbin and Han, Yu and Huang, Fei and others},
  title = {Qwen technical report},
  year = {2023},
}

@article{porian2024resolving,
  journal = {Advances in Neural Information Processing Systems},
  volume = {37},
  pages = {100535--100570},
  author = {Porian, Tomer and Wortsman, Mitchell and Jitsev, Jenia and Schmidt, Ludwig and Carmon, Yair},
  title = {Resolving discrepancies in compute-optimal scaling of language models},
  year = {2024},
}

@inproceedings{shi2020rmsprop,
  booktitle = {International Conference on Learning Representations},
  author = {Shi, Naichen and Li, Dawei and Hong, Mingyi and Sun, Ruoyu},
  title = {RMSprop converges with proper hyper-parameter},
  year = {2020},
}

@article{marek2025small,
  journal = {arXiv preprint arXiv:2507.07101},
  author = {Marek, Martin and Lotfi, Sanae and Somasundaram, Aditya and Wilson, Andrew Gordon and Goldblum, Micah},
  title = {Small batch size training for language models: When vanilla sgd works, and why gradient accumulation is wasteful},
  year = {2025},
}

@article{wortsman2023small,
  journal = {arXiv preprint arXiv:2309.14322},
  author = {Wortsman, Mitchell and Liu, Peter J and Xiao, Lechao and Everett, Katie and Alemi, Alex and Adlam, Ben and Co-Reyes, John D and Gur, Izzeddin and Kumar, Abhishek and Novak, Roman and others},
  title = {Small-scale proxies for large-scale transformer training instabilities},
  year = {2023},
}

@book{tao2012topics,
  publisher = {American Mathematical Soc.},
  volume = {132},
  author = {Tao, Terence},
  title = {Topics in random matrix theory},
  year = {2012},
}

@article{das2024towards,
  journal = {arXiv preprint arXiv:2402.07114},
  author = {Das, Rudrajit and Agarwal, Naman and Sanghavi, Sujay and Dhillon, Inderjit S},
  title = {Towards Quantifying the Preconditioning Effect of Adam},
  year = {2024},
}

@article{hoffmann2022training,
  journal = {arXiv preprint arXiv:2203.15556},
  author = {Hoffmann, Jordan and Borgeaud, Sebastian and Mensch, Arthur and Buchatskaya, Elena and Cai, Trevor and Rutherford, Eliza and Casas, Diego de Las and Hendricks, Lisa Anne and Welbl, Johannes and Clark, Aidan and others},
  title = {Training compute-optimal large language models},
  year = {2022},
}

@article{ormaniec2024does,
  journal = {arXiv preprint arXiv:2410.10986},
  author = {Ormaniec, Weronika and Dangel, Felix and Singh, Sidak Pal},
  title = {What Does It Mean to Be a Transformer? Insights from a Theoretical Hessian Analysis},
  year = {2024},
}

@article{zhang2024transformers,
  journal = {Advances in Neural Information Processing Systems},
  volume = {37},
  pages = {131786--131823},
  author = {Zhang, Yushun and Chen, Congliang and Ding, Tian and Li, Ziniu and Sun, Ruoyu and Luo, Zhi-Quan},
  title = {Why transformers need adam: A hessian perspective},
  year = {2024},
}

@article{sun2021worst,
  journal = {Mathematical Programming},
  publisher = {Springer},
  volume = {185},
  pages = {487--520},
  author = {Sun, Ruoyu and Ye, Yinyu},
  title = {Worst-case complexity of cyclic coordinate descent: O (n\^{} 2) O (n 2) gap with randomized version},
  year = {2021},
}

@inproceedings{rajbhandari2021zero,
  booktitle = {Proceedings of the international conference for high performance computing, networking, storage and analysis},
  pages = {1--14},
  author = {Rajbhandari, Samyam and Ruwase, Olatunji and Rasley, Jeff and Smith, Shaden and He, Yuxiong},
  title = {Zero-infinity: Breaking the gpu memory wall for extreme scale deep learning},
  year = {2021},
}

@inproceedings{rajbhandari2020zero,
  booktitle = {SC20: International Conference for High Performance Computing, Networking, Storage and Analysis},
  organization = {IEEE},
  pages = {1--16},
  author = {Rajbhandari, Samyam and Rasley, Jeff and Ruwase, Olatunji and He, Yuxiong},
  title = {Zero: Memory optimizations toward training trillion parameter models},
  year = {2020},
}

@article{wen2025fantastic,
  title={Fantastic pretraining optimizers and where to find them},
  author={Wen, Kaiyue and Hall, David and Ma, Tengyu and Liang, Percy},
  journal={arXiv preprint arXiv:2509.02046},
  year={2025}
}

@article{adamcitation,
  title={\url{https://scholar.google.com/scholar?oi=bibs&hl=en&cites=16194105527543080940,10561642725708924006,8776215530672338536,17607154055187750797,8498514209881222824,3931560505490345047,16226219632797774343,16111079307796299296,13264970321150126215,11869859376987161805,11430597276676453789,5669591665425275538,1944550801657937622,275207800116052367,630464762792003720,14056964711882014466,3067190135652312858,17573188221268842192,9561591289532092532,5915381407158841087}},
  author={Google Scholar},
  journal={Google Scholar},
  year={2025}
}

@article{chen2016direct,
  title={The direct extension of ADMM for multi-block convex minimization problems is not necessarily convergent},
  author={Chen, Caihua and He, Bingsheng and Ye, Yinyu and Yuan, Xiaoming},
  journal={Mathematical Programming},
  volume={155},
  number={1},
  pages={57--79},
  year={2016},
  publisher={Springer}
}

@article{sun2020efficiency,
  title={On the efficiency of random permutation for ADMM and coordinate descent},
  author={Sun, Ruoyu and Luo, Zhi-Quan and Ye, Yinyu},
  journal={Mathematics of Operations Research},
  volume={45},
  number={1},
  pages={233--271},
  year={2020},
  publisher={INFORMS}
}

@article{bottou2018optimization,
  title={Optimization methods for large-scale machine learning},
  author={Bottou, L{\'e}on and Curtis, Frank E and Nocedal, Jorge},
  journal={SIAM review},
  volume={60},
  number={2},
  pages={223--311},
  year={2018},
  publisher={SIAM}
}

@article{bertsekas1997nonlinear,
  title={Nonlinear programming},
  author={Bertsekas, Dimitri P},
  journal={Journal of the Operational Research Society},
  volume={48},
  number={3},
  pages={334--334},
  year={1997},
  publisher={Taylor \& Francis}
}

@article{luo1991convergence,
  title={On the convergence of the LMS algorithm with adaptive learning rate for linear feedforward networks},
  author={Luo, Zhi-Quan},
  journal={Neural Computation},
  volume={3},
  number={2},
  pages={226--245},
  year={1991},
  publisher={MIT Press One Rogers Street, Cambridge, MA 02142-1209, USA journals-info~...}
}

@article{bertsekas2000gradient,
  title={Gradient convergence in gradient methods with errors},
  author={Bertsekas, Dimitri P and Tsitsiklis, John N},
  journal={SIAM Journal on Optimization},
  volume={10},
  number={3},
  pages={627--642},
  year={2000},
  publisher={SIAM}
}

@article{zhang2026adam,
  title={Adam Converges Without Any Modification On Update Rules},
  author={Zhang, Yushun and Li, Bingran and Chen, Congliang and Luo, Zhi-Quan and Sun, Ruoyu},
  journal={arXiv preprint arXiv:2603.02092},
  year={2026}
}

@article{ma2026preconditioning,
  title={Preconditioning benefits of spectral orthogonalization in muon},
  author={Ma, Jianhao and Huang, Yu and Chi, Yuejie and Chen, Yuxin},
  journal={arXiv preprint arXiv:2601.13474},
  year={2026}
}

@article{dewulf2026aurora,
  title   = {Aurora: A Leverage-Aware Optimizer for Rectangular Matrices},
  author  = {Dewulf, Alec and Pai, Dhruv and Yang, Li and Zhang, Ashley
             and Keigwin, Ben},
  year    = {2026},
  url     = {https://tilderesearch.com/blog/aurora}
}

@article{li2025normuon,
  title={NorMuon: Making Muon more efficient and scalable},
  author={Li, Zichong and Liu, Liming and Liang, Chen and Chen, Weizhu and Zhao, Tuo},
  journal={arXiv preprint arXiv:2510.05491},
  year={2025}
}

@misc{jordan2026modded,
  author = {Keller Jordan},
  title = {Modded-NanoGPT: A Speedrun to Train a 124M Parameter Transformer Efficiently},
  year = {2026},
  howpublished = {\url{https://github.com/KellerJordan/modded-nanogpt}},
  note = {Accessed: 2026-06-23}
}

@article{deng2026rmnp,
  title={RMNP: Row-momentum normalized preconditioning for scalable matrix-based optimization},
  author={Deng, Shenyang and Ouyang, Zhuoli and Pang, Tianyu and Liu, Zihang and Jin, Ruochen and Yu, Shuhua and Yang, Yaoqing},
  journal={arXiv preprint arXiv:2603.20527},
  year={2026}
}

@article{yuan2026nora,
  title={Nora: Normalized Orthogonal Row Alignment for Scalable Matrix Optimizer},
  author={Yuan, Jinghui and Zou, Jiaxuan and Wang, Shuo and Liu, Yong and Nie, Feiping},
  journal={arXiv preprint arXiv:2605.03769},
  year={2026}
}

@article{wen2025sron,
  title={SRON: State-free LLM Training via Row-wise Gradient Normalization},
  author={Wen, Ziqing and Shi, Yanqi and Wang, Jiahuan and Luo, Ping and Qiao, Linbo and Li, Dongsheng and Sun, Tao},
  year={2025}
}

@article{Buss2025CaloHadronicAD,
    title         = {CaloHadronic: a diffusion model for the generation of hadronic showers},
    author        = {Thorsten Buss and Frank Gaede and Gregor Kasieczka and Anatolii Korol and Katja Krüger and Peter McKeown and Martina Mozzanica},
    year          = {2025},
    journal       = {Journal of Instrumentation},
    volume        = {21},
    pages         = {P01042},
    month         = {January},
    note          = {Also available as arXiv:2506.21720},
    doi           = {10.1088/1748-0221/21/01/P01042},
    eprint        = {2506.21720},
    archivePrefix = {arXiv},
    primaryClass  = {physics.ins-det},
    url           = {https://doi.org/10.1088/1748-0221/21/01/P01042}
}

@article{chen2026simplegpt,
    title         = {SimpleGPT: Improving GPT via A Simple Normalization Strategy},
    author        = {Marco Chen and Xianbiao Qi and Yelin He and Jiaquan Ye and Rong Xiao},
    year          = {2026},
    month         = {February},
    note          = {arXiv preprint arXiv:2602.01212},
    eprint        = {2602.01212},
    archivePrefix = {arXiv},
    primaryClass  = {cs.LG},
    url           = {https://arxiv.org/abs/2602.01212}
}

@article{bai2025adaptive,
  title={Adaptive Preconditioners Trigger Loss Spikes in Adam},
  author={Bai, Zhiwei and Zhou, Zhangchen and Zhao, Jiajie and Li, Xiaolong and Li, Zhiyu and Xiong, Feiyu and Yang, Hongkang and Zhang, Yaoyu and Xu, Zhi-Qin John},
  journal={arXiv preprint arXiv:2506.04805},
  year={2025}
}

@article{abreu2025potential,
  title={The potential of second-order optimization for llms: A study with full gauss-newton},
  author={Abreu, Natalie and Vyas, Nikhil and Kakade, Sham and Morwani, Depen},
  journal={arXiv preprint arXiv:2510.09378},
  year={2025}
}

@article{bai2026towards,
  title={Towards understanding Adam convergence on highly degenerate polynomials},
  author={Bai, Zhiwei and Zhao, Jiajie and Zhou, Zhangchen and Xu, Zhi-Qin John and Zhang, Yaoyu},
  journal={arXiv preprint arXiv:2603.09581},
  year={2026}
}

@article{team2026kimi,
  title={Kimi K3: Open Frontier Intelligence},
  author={Team, Kimi and Bai, Tongtong and Bai, Yifan and Bao, Yiping and Cai, Jianfeng and Cai, Xinyuan and Cao, Peizhou and Cao, Yuxuan and Chai, Ziwei and Charles, Y and others},
  journal={arXiv preprint arXiv:2607.24653},
  year={2026}
}

@online{kexuefm-11848,
        title={A Brief Look at K3's MoE and Attention},
        author={Jianlin Su},
        year={2026},
        month={Aug},
        url={\url{https://spaces.ac.cn/archives/11848}},
}

@article{zhu2025apollo,
  title={Apollo: Sgd-like memory, adamw-level performance},
  author={Zhu, Hanqing and Zhang, Zhenyu and Cong, Wenyan and Liu, Xi and Park, Sem and Chandra, Vikas and Long, Bo and Pan, David Z and Wang, Zhangyang and Lee, Jinwon},
  journal={Proceedings of Machine Learning and Systems},
  volume={7},
  year={2025}
}

@misc{held2026delphi,
  author       = {Held, Will},
  title        = {{Scaling Laws That Extrapolate 300{$\times$} past the Fit}},
  howpublished = {Open Athena Blog},
  year         = {2026},
  month        = may,
  url          = {https://openathena.ai/blog/delphi/}
}

@book{zipf1949human,
  author    = {Zipf, George Kingsley},
  title     = {Human Behavior and the Principle of Least Effort: An Introduction to Human Ecology},
  publisher = {Addison-Wesley Press},
  address   = {Cambridge, MA},
  year      = {1949}
}

@article{deepseekai2026v41flash,
  author       = {{DeepSeek-AI} and Xu, Anyi and Li, B. and Lin, Bangcai and Xue, Bing},
  title        = {DeepSeek-V4.1-Flash: Pushing the Limits of {KV} Cache Compression},
  journal      = {CoRR},
  volume       = {abs/2609.19969},
  year         = {2026},
  doi          = {10.48550/arXiv.2609.19969},
  eprinttype   = {arXiv},
  eprint       = {2609.19969},
  url          = {https://arxiv.org/abs/2609.19969}
}
}
\end{document}